\documentclass[journal]{IEEEtran}

\usepackage[utf8]{inputenc}
\usepackage{hyperref}
\usepackage{amsmath,amsfonts}
\usepackage{graphicx}
\usepackage{array}
\usepackage{textcomp}
\usepackage{stfloats}
\usepackage{url}
\usepackage{subcaption}
\usepackage{bm}
\usepackage{cite}
\usepackage{rotating}
\usepackage{multirow}
\usepackage{hhline}
\usepackage{verbatim}
\usepackage{soul}
\usepackage{afterpage}
\usepackage{color}
\usepackage{latexsym}
\usepackage{amssymb}
\usepackage{mathtools}
\usepackage{pdflscape}
\usepackage{enumerate}

\usepackage[inline]{enumitem}
\usepackage{xspace} 
\usepackage[ruled]{algorithm2e}
\usepackage{algorithmicx}
\usepackage{float}
\usepackage{booktabs}
\usepackage{diagbox}
\usepackage{tabu}
\usepackage{colortbl}

\usepackage{pifont} 
\newcommand{\cmark}{\ding{51}}%
\newcommand{\xmark}{\ding{55}}%

\usepackage[dvipsnames]{xcolor}
\usepackage[normalem]{ulem}
\usepackage{makecell}
\usepackage{algorithm2e}
\usepackage{multicol}
\usepackage[capitalise]{cleveref}
\usepackage[percent]{overpic}
\usepackage{tikz}
\usetikzlibrary{arrows.meta,arrows,tikzmark,calc,backgrounds,decorations.pathreplacing,calligraphy}

\usepackage{cuted}

\usepackage[per-mode=fraction]{siunitx}
\usepackage{comment}
\usepackage[normalem]{ulem}
\usepackage{xspace}
\usepackage{xcolor}

\DeclareSIUnit\px{px}
\DeclareSIUnit\fps{fps}

\definecolor{OliveGreen}{RGB}{0,200,25}

\newcommand{\eg}{e.\,g.\, }
\newcommand{\etal}{et\,al.\ }

\newcommand{\armarVI}{\mbox{ARMAR-6}\xspace}

\definecolor{cadmiumyellow}{rgb}{1.0, 0.96, 0.0}
\definecolor{canaryyellow}{rgb}{1.0, 0.94, 0.0}
\definecolor{bananayellow}{rgb}{1.0, 0.88, 0.21}
\definecolor{yellow(process)}{rgb}{1.0, 0.94, 0.0}
\definecolor{applegreen}{rgb}{0.55, 0.71, 0.0}
\definecolor{capri}{rgb}{0.0, 0.75, 1.0}
\definecolor{corn}{rgb}{0.98, 0.93, 0.36}
\definecolor{cornflowerblue}{rgb}{0.39, 0.58, 0.93}
\definecolor{darkblue}{rgb}{0.0, 0.0, 0.55}

\newcommand{\nokpt}{$\textcolor{red}{\times}$}

\newcommand{\drawlabel}[3]{
	\node[align=center, fill=white, text opacity=1, #3] at #2 {#1};
}

\usepackage{mathrsfs}

\newcommand{\image}{\boldsymbol{A}}

\newcommand{\bp}{\boldsymbol{p}}
\newcommand{\bd}{\boldsymbol{d}}
\newcommand{\bw}{\boldsymbol{w}}

\newcommand{\force}{\boldsymbol{f}}
\newcommand{\bx}{\boldsymbol{x}}
\newcommand{\bk}{\boldsymbol{k}}
\newcommand{\bK}{\boldsymbol{K}}
\newcommand{\bbK}{\bar{\boldsymbol{K}}}
\newcommand{\tbK}{\tilde{\boldsymbol{K}}}

\newcommand{\bh}{\boldsymbol{h}}
\newcommand{\bS}{\boldsymbol{S}}

\newcommand{\lf}{\mathcal{F}}
\newcommand{\manifold}[1]{\mathcal{M}_{#1}}
\newcommand{\nth}{{}^\text{th}}
\newcommand{\ptop}{\mathsf{p2p}}
\newcommand{\ptol}{\mathsf{p2l}}
\newcommand{\ptoP}{\mathsf{p2P}}
\newcommand{\ptoc}{\mathsf{p2c}}
\newcommand{\ptoS}{\mathsf{p2S}}

\newcommand{\setP}[1]{\mathcal{P}_{#1}}
\newcommand{\setD}[1]{\mathcal{D}_{\text{#1}}}
\newcommand{\setV}[1]{\mathcal{V}_{\text{#1}}}
\newcommand{\setO}[1]{\mathcal{O}_{\text{#1}}}
\newcommand{\setC}[1]{\mathcal{C}_{\text{#1}}}
\newcommand{\setT}[1]{\mathcal{T}_{\text{#1}}}
\newcommand{\set}[1]{\{#1\}}

\newcommand{\smalleq}[1]{\scalebox{0.9}{#1}}
\newcommand{\scaleeq}[2]{\scalebox{#1}{$#2$}}

\newcommand{\traj}{\bm{\tau}}
\newcommand{\bnu}{\bm{\nu}}
\newcommand{\vari}{\bm{\eta}}
\newcommand{\objecti}{\smalleq{$O_i$}}
\newcommand{\objectm}{\smalleq{$O_m$}}
\newcommand{\objects}{\smalleq{$O_s$}}

\newcommand{\master}{\scaleeq{0.95}{\mathsf{master}}}
\newcommand{\slave}{\scaleeq{0.95}{\mathsf{slave}}}
\newcommand{\task}[2]{\scalebox{0.95}{$\mathsf{#1}$ \raisebox{0.5pt}{\circled{#2}}}}
\newcommand{\taskabbr}[1]{\scalebox{0.95}{$\mathsf{#1}$}}
\newcommand{\scaletask}[3]{
	\scalebox{0.95}{$\mathsf{#1}$} 
	\scalebox{#3}{\raisebox{0.9pt}{\circled{#2}}}\hspace{-2.5pt}
}

\DeclareMathOperator{\Pri}{Pri}
\DeclareMathOperator{\avg}{avg}
\DeclareMathOperator*{\argmax}{arg\,max}
\DeclareMathOperator*{\argmin}{arg\,min}

\DeclareRobustCommand\sampleline[1]{
	\tikz\draw[#1] (0,0) (0,\the\dimexpr\fontdimen22\textfont2\relax) -- (1em,\the\dimexpr\fontdimen22\textfont2\relax);%
}

\DeclareRobustCommand\inlinecirc[1]{
	\raisebox{1pt}{
		\tikz%
		\node[shape=circle, draw, inner sep=1pt, #1] (0,0) {};
	}
	\hspace{-0.7em}
}

\DeclareRobustCommand\inlinearrow[1]{ 
	\hspace{-0.7em}
	\tikz{
		\draw[-{to[length=1.5mm]},
		#1,line width=1.1pt](0,0) (0,\the\dimexpr\fontdimen22\textfont2\relax) -- (1em,\the\dimexpr\fontdimen22\textfont2\relax);
	}
	\hspace{-0.7em}
}

\DeclareRobustCommand\inlinekpt[2]{
	\raisebox{1pt}{
		\tikz%
		\node[shape=circle, draw=#1, inner sep=1.5pt, #2] (0,0) {};
	}
	\hspace{-0.7em}
}

\DeclareRobustCommand\circled[1]{\tikz[baseline=(char.base)]{
		\node[shape=circle,fill,inner sep=2pt,scale=0.3] (char) {\textcolor{white}{\Huge \textbf{#1}}};}}

\usepackage{amsmath}

\newcolumntype{P}[1]{>{\centering\arraybackslash}p{#1}}
\newcolumntype{M}[1]{>{\centering\arraybackslash}m{#1}}

\newlength\imgwidth
\newlength\imgheight
\newlength{\subfight}
\newsavebox{\subfigbox}
\newcommand{\trsp}{\mathsf{T}}

\begin{document}

\title{K-VIL: Keypoints-based Visual Imitation Learning}

\author{Jianfeng~Gao, 
        Zhi~Tao,  
        No\'emie Jaquier, 
        and~Tamim~Asfour
        \thanks{This work has been supported by the German Federal Ministry of Education
and Research (BMBF) under the project OML and by the Carl Zeiss Foundation under the project JuBot.	~(\emph{Corresponding author: Jianfeng Gao})}
	\thanks{The authors are with the Institute for Anthropomatics and Robotics, Karlsruhe Institute of Technology, Karlsruhe, Germany. 
        E-mails: \{jianfeng.gao, noemie.jaquier, asfour\}@kit.edu; zhitao.robotics@gmail.com.
        }
        \thanks{Published version: \url{https://ieeexplore.ieee.org/document/10189175}}
}

\markboth{IEEE Transactions on Robotics, PREPRINT VERSION, July~2023}
{Gao \MakeLowercase{\textit{et al.}}: K-VIL: Keypoints-based Visual Imitation Learning}

\IEEEpubid{10.1109/TRO.2023.3286074~\copyright~2023 IEEE}

\makeatletter
\let\@oldmaketitle\@maketitle
\renewcommand{\@maketitle}{\@oldmaketitle
	\vspace{-13ex}
}
\makeatother
\maketitle

\begin{strip}
\captionsetup[sub]{labelformat=parens}
\begingroup
    \captionsetup{type=figure}
    \begin{subfigure}[t]{0.26\textwidth}
        \centering
        \begin{tikzpicture}
            \node (image) at (0,0) {
                 \includegraphics[width=\textwidth]{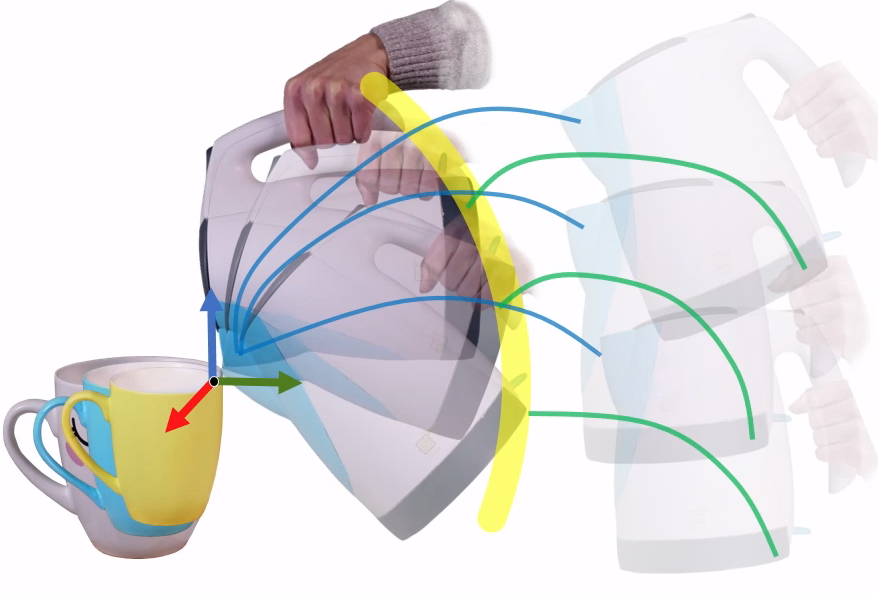}
            };
            \coordinate (k1)                   at (-1.10, -0.30);
            \coordinate (k2_1)                 at ( 0.20,  0.52);
            \coordinate (k2_2)                 at ( 0.38,  0.05);
            \coordinate (k2_3)                 at ( 0.43, -0.60);

            \node[draw, fill=ForestGreen!90, circle, inner sep=0pt,minimum size=5pt] (pk2_1) at (k2_1) {}; 
            \node[draw, fill=ForestGreen!90, circle, inner sep=0pt,minimum size=5pt] (pk2_2) at (k2_2) {}; 
            \node[draw, fill=ForestGreen!90, circle, inner sep=0pt,minimum size=5pt] (pk2_3) at (k2_3) {}; 
            \node[draw, fill=blue!70, circle, inner sep=0pt,minimum size=5pt] (pk1) at (k1) {}; 
            \node at ($(k2_3)           + ( 0.3, -0.3)$) {$\textcolor{ForestGreen}{\bk_2}$};
            \node at ($(k1)             + (-0.3,  0.2)$) {$\textcolor{blue}{\bk_1}$};
            
            \node at (-1.0, -0.80) {\scalebox{1.0}{$\lf$}};
            \node at ( 1.2,  1.20) {\scalebox{0.8}{trajectory variation}};
            \node at (-1.5, -1.60) {\scalebox{0.8}{shape variation}};
            \node at ( 0.5, -1.58) {\scalebox{0.8}{pose variation}};
        \end{tikzpicture}
        \vspace{-1.8em}
        \caption{\small{Human demonstrations}\label{subfig:concept_demo}}
    \end{subfigure}
    \begin{subfigure}[t]{0.24\textwidth}
        \centering
        \begin{tikzpicture}
            \node (image) at (0,0) {
                 \includegraphics[width=\textwidth]{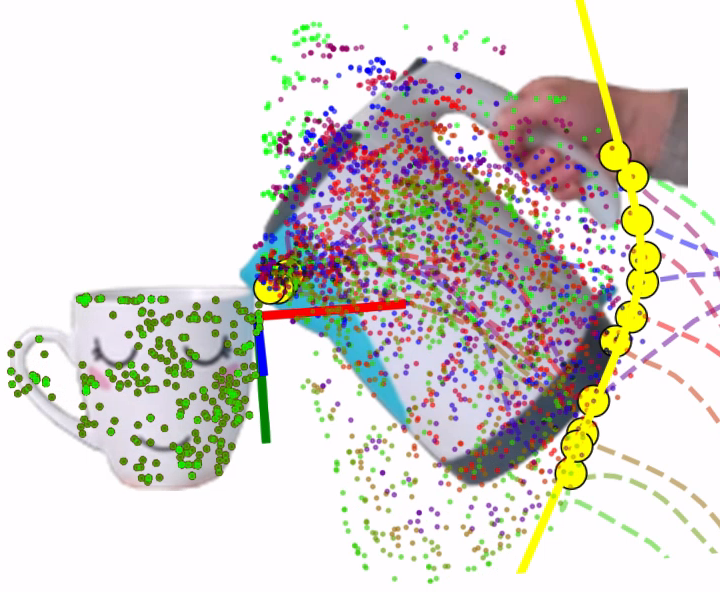}
            };
            \coordinate (k1)            at (-0.60, -0.00);
            \coordinate (k2)            at ( 1.40, -0.52);

            \node at ($(k2)             + ( 0.3, -0.3)$) {$\textcolor{ForestGreen}{\bk_2}$};
            \node at ($(k1)             + (-0.3,  0.2)$) {$\textcolor{blue}{\bk_1}$};
            
            \node at (-0.4, -0.40) {\scalebox{1.0}{$\lf$}};
        \end{tikzpicture}
        \vspace{-1.8em}
        \caption{\small{Extraction of keypoints}\label{subfig:concept_kvil}}
    \end{subfigure}
    \begin{subfigure}[t]{0.26\textwidth}
        \centering
        \begin{tikzpicture}
            \node (image) at (0,0) {
                 \includegraphics[width=\textwidth]{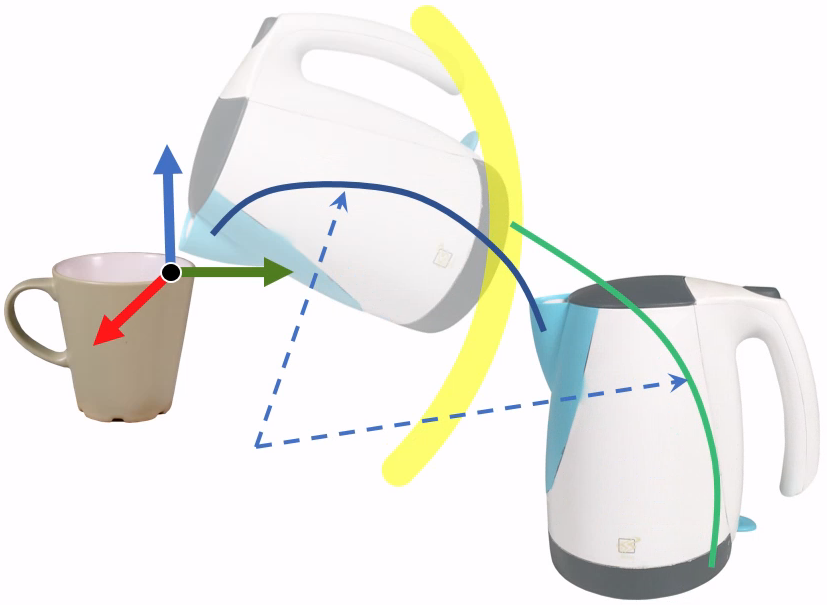}
            };
            \coordinate (k1_g)                 at (-1.20,  0.32);
            \coordinate (k1)                   at ( 0.72, -0.20);
            \coordinate (k2_g)                 at ( 0.55,  0.45);
            \coordinate (k2)                   at ( 1.70, -1.50);

            \node[fill=ForestGreen!90, circle, inner sep=0pt,minimum size=5.5pt] (pk2) at (k2) {}; 
            \node[draw, fill=ForestGreen!100, line width=0.25mm, circle, inner sep=0pt,minimum size=5pt] (pk2_g) at (k2_g) {}; 
            \node[fill=blue!70, circle, inner sep=0pt,minimum size=5.5pt] (pk1) at (k1) {}; 
            \node[draw, fill=blue!70, line width=0.25mm, circle, inner sep=0pt,minimum size=5pt] (pk1_g) at (k1_g) {}; 
            \node at ($(k2_g)             + ( 0.3,  0.3)$) {$\textcolor{ForestGreen}{\bk_2}$};
            \node at ($(k1_g)             + (-0.4,  0.2)$) {$\textcolor{blue}{\bk_1}$};
            
            \node at (-1.2, -0.10) {\scalebox{1.0}{$\lf$}};
            \node at (-1.3,  1.20) {\scalebox{0.8}{\textcolor{blue}{point-to-point}}};
            \node at ( 0.8,  1.20) {\scalebox{0.8}{\textcolor{ForestGreen}{point-to-curve}}};
            \node at (-0.8, -1.20) {\scalebox{0.8}{movement primitives}};
        \end{tikzpicture}
        \vspace{-1.8em}
        \caption{\small{Geometric task representation}\label{subfig:concept_geom}} 
    \end{subfigure}
    \begin{subfigure}[t]{0.22\textwidth}
        \centering
        \begin{tikzpicture}
            \node (image) at (0,0) {
                 \includegraphics[width=\textwidth, trim=90 30 40 20, clip]{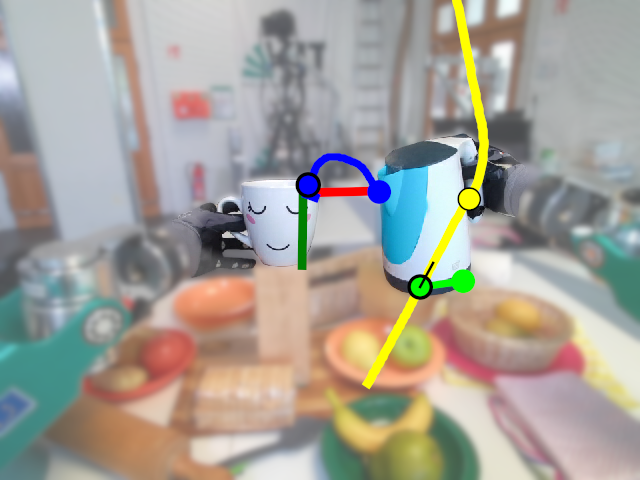}
            };
            \coordinate (k1_g)          at (-0.285, 0.39);
            \coordinate (k2)            at ( 0.92, -0.35);
            \coordinate (k2_g)          at ( 0.59, -0.41);
            \coordinate (k2_m)          at ( 0.97,  0.29);
            
            \node[draw, fill=blue!70, line width=0.25mm, circle, inner sep=0pt,minimum size=6pt] (pk1) at (k1_g) {}; 
            \node[fill=ForestGreen!100, circle, inner sep=0pt,minimum size=6.5pt] (pk2) at (k2) {}; 
            \node[draw, fill=ForestGreen!100, line width=0.25mm, circle, inner sep=0pt,minimum size=6pt] (pk2) at (k2_g) {}; 
            \node[draw, fill=yellow!100, line width=0.25mm, circle, inner sep=0pt,minimum size=6pt] (pk2m) at (k2_m) {}; 
            
            \node at ($(k2)             + ( 0.4,  0.1)$) {$\textcolor{ForestGreen}{\bk_2}$};
            \node at ($(k1_g)           + (-0.3,  0.2)$) {$\textcolor{blue}{\bk_1}$};
            
            \node at (-0.1, 0.10) {\scalebox{1.0}{$\lf$}};
        \end{tikzpicture}
        \vspace{-1.8em}
        \caption{\small{Reproduction}\label{subfig:concept_exe}}
    \end{subfigure}
    \caption{Overview of the K-VIL approach. \subref{subfig:concept_demo} Human demonstration videos of manipulation actions involving categorical objects with shape, pose, and trajectory variations. 
	\subref{subfig:concept_kvil} Sampling of dense candidate points from the object surface.  
	\subref{subfig:concept_geom} Extraction of \emph{sparse keypoints} $\textcolor{blue}{\bk_1}, \textcolor{ForestGreen}{\bk_2}$ subject to certain types of \emph{geometric constraints} (point-to-point and point-to-curve), their associated \emph{local frames} $\lf$ and the \emph{movement primitives} which represent the demonstrated keypoint motions.
	\subref{subfig:concept_exe} Adaptation of the learned generalizable geometric task representation to a new scene, and execution by the robot.
    \label{fig:overview}} 
    \vspace{-1ex}
\endgroup
\end{strip}

\begin{abstract}

Visual imitation learning provides efficient and intuitive solutions for robotic systems to acquire novel manipulation skills. However, simultaneously learning geometric task constraints and control policies from visual inputs alone remains a challenging problem.
In this paper, we propose the \emph{keypoint-based visual imitation learning} (K-VIL) approach that automatically extracts sparse, object-centric, and embodiment-independent task representations from a small number of human demonstration videos.
The task representation is composed of keypoint-based geometric constraints on principal manifolds, their associated local frames, and the  movement primitives that are then needed for the task execution. 
Our approach is capable of extracting such task representations from a single demonstration video, and of incrementally updating them 
when new demonstrations are available.
To reproduce manipulation skills using the learned set of prioritized geometric constraints in novel scenes, we introduce a novel keypoint-based admittance controller.
We evaluate our approach in several real-world applications, showcasing
its ability to deal with cluttered scenes, viewpoint mismatch, new instances of categorical objects, and large object pose and shape variations. Our evaluation demonstrates the efficiency and robustness of our approach in both one-shot and few-shot imitation learning settings.
Videos and source code are available at \href{https://sites.google.com/view/k-vil}{https://sites.google.com/view/k-vil}.
\end{abstract}


\section{Introduction}
Observational learning, i.e., the ability to develop new skills from observed actions and their outcome, is an important learning mechanism in our daily lives \cite{bandura_social_1977, wolpert_principles_2011, burke_neural_2010}. 
For example, by watching a few videos showing people pouring water from a kettle into different teacups (as in \cref{subfig:concept_demo}), we can easily learn 
``what'' a pouring task is and ``how'' to perform it.
From a computational point of view, the spout and bottom of the kettle can be represented by two keypoints $\bk_1, \bk_2$. As shown in Fig.~\ref{subfig:concept_geom}, a pouring task then simply consists in aligning the spout $\bk_1$ with a point above the rim of the cup (point-to-point constraint)
and similarly aligning the bottom $\bk_2$ with a curve that controls the kettle's angle of inclination (point-to-curve constraint).
Such sets of keypoints and keypoint-based geometric constraints can generally be used to represent daily manipulation tasks, i.e., to parameterize the motion of their functional parts relative to some local frames of reference. Moreover, these keypoints and local frames can also be associated with local visual features of object functional parts.

\IEEEpubidadjcol
In this paper, we propose to exploit such task representations to teach manipulation skills to robots from video demonstrations. We additionally aim for generalizable skills, which can be reused in novel scenes (see \cref{subfig:concept_exe} for an example).
In this context, three main challenges arise, namely, 
\begin{enumerate*}[label=(\roman*)]
    \item \label{correspond}the detection and efficient extraction of task-relevant keypoints on objects; 
    \item the definition of generalizable and \mbox{embodiment}-independent task representations; and
    \item the reproduction of the demonstrated task and its adaptation to new scenes.
\end{enumerate*}
To address the first two challenges, we first \emph{densely} sample a set of candidate points from the object mask provided by Mask R-CNN~\cite{he_mask_rcnn} and leverage the correspondence detection of Dense Object Net (DON)~\cite{florencemanuelli2018dense} to track their motions in the demonstration videos and obtain their 3D positions (see Section~\ref{sec:background} for a short background). 
Then, we jointly extract a set of \emph{sparse keypoints} and a set of \emph{keypoint-based geometric constraints} representing the task, as shown in \cref{subfig:concept_kvil} (see Section~\ref{sec:approach}). 
To do so, we exploit principal manifold estimation algorithms (PME)~\cite{meng_principal_2021}, which are intrinsically more data- and time-efficient than approaches based on supervised learning~\cite{gao_kpam_2021,xu_affordance_2021} and reinforcement learning (RL)~\cite{sharma_generalizing_2021,qin_keto_2020}. 
The resulting geometric constraints are expressed relative to local frames defined on target objects and are easily adjustable to pose and shape variations of the target object.
As shown in previous works~\cite{gao_kpam_2021,sharma_learning_2020,sharma_generalizing_2021}, such object-centric task representations facilitate the transfer of manipulation skills between demonstrators and imitators. It also allows our approach to deal with demonstrations provided from different viewpoints.
The representation of the task in the form of keypoints and their constraints also enables the use of simpler control policies~\cite{jin_generalizable_2022}, such as movement primitives, for executing the given task. 
In this paper, we exploit this property to address our third challenge. Namely, we encode the keypoint motions relative to the corresponding local frames as via-point movement primitives (VMPs)~\cite{zhou_learning_2019}, which are flexible in terms of temporal scaling and trajectory adaptation while maintaining the demonstrated motion styles. The learned keypoint motions can then be executed on a robot by leveraging our novel keypoint-based admittance controller (Section~\ref{sec:kpts_control}).
We validate our approach by learning various real-world daily tasks from video demonstrations and reproducing them with a humanoid robot (Section~\ref{sec:eval}). The results show that K-VIL efficiently extracts generalizable manipulation skills, handles viewpoint mismatch, and
deals with large pose and shape variations of categorical objects in cluttered scenes.

Our contributions are threefold:
\begin{enumerate*}[label=(\roman*)]
    \item We introduce the \emph{Keypoint-based Visual Imitation Learning} (K-VIL) approach for automatic and incremental extraction of sparse, object-centric, viewpoint-invariant, and embodiment-independent task representations. 
    K-VIL extracts task representations from a single demonstration video and improves them as new demonstrations are available. The task representations consist of keypoint-based geometric constraints on principal manifolds, their associated local frames, and the movement primitives required to reproduce the task.
    \item We formulate and learn a large variety of geometric constraints, which allow the proposed task representation to be flexible and efficient.
    \item We propose a novel keypoint-based admittance controller that handles a set of prioritized geometric constraints and allows successful reproductions of the learned task in novel scenes.
\end{enumerate*}

\section{Related Work}

Visual Imitation Learning (VIL) is a class of imitation learning (IL) frameworks in which only visual sensory input 
is presented to the imitator. 
The main challenges of VIL are 
\begin{enumerate*}[label=(\roman*)]
    \item the detection of visual correspondences between the demonstrator's and imitator's context, i.e.,  context translation~\cite{Sharma2019ThirdPersonVI,smith_avid_2020};
    \item the fine-grained understanding of scene structures~\cite{sieb2020graph}, along with the design of generalizable task representation; and
    \item the design of sample efficient and scalable control policies.
\end{enumerate*}
This latter challenge is often tackled along with the former ones, as described next.

\subsection{Context Translation}
Context translation has typically been addressed by training context translators in the demonstrator context to predict the observations in the imitator (e.g., the robot) context. 
Pixel-level translators were used in~\cite{smith_avid_2020,liu2018imitation,Dwibedi2018LearningAR,Adversarial2022} to further train RL policies by maximizing the similarity between predicted and received robot observations. 
Despite the performance of such models, their training is computationally expensive and time-consuming.
To improve learning efficiency, Sharma \etal~\cite{Sharma2019ThirdPersonVI} combined a goal-level translator with a task-agnostic control policy, 
which was trained independently and shared among different tasks.
In contrast to these works, K-VIL represents the context via a set of object-centric keypoints and their respective geometric constraints, thus facilitating the context translation between demonstrators and imitators. 
In addition, by leveraging Mask R-CNN and DON models --- which are trained beforehand in a task-agnostic manner and shared among tasks --- \mbox{K-VIL}'s representations can be acquired from a single or few demonstrations.

\subsection{Fine-grained Understanding of Scene Structure}
The above approaches do not scale to categorical objects as they do not explicitly extract the scene structures with respect to objects and their functional parts.
In the literature, the understanding of fine-grained scene structures is mainly achieved through
\begin{enumerate*}[label=\emph{\arabic*})]
    \item \label{descriptor}the viewpoint-invariant representation of fine-grained scene features;
    \item \label{frame}the extraction of a hierarchy of the scene structure; and
    \item \label{const}the definition of task constraints.
\end{enumerate*}

\subsubsection{Viewpoint-invariant representation}
\emph{Dense visual descriptors} such as DON \cite{florencemanuelli2018dense} and Neural Descriptor Fields (NDFs) \cite{simeonov_neural_2021} represent fine-grained scene features by detecting dense correspondences of categorical objects, thus allowing point-based representation of object functional parts. 
However, in~\cite{florence_self-supervised_2020,simeonov_neural_2021,pathakICLR18zeroshot}, access to the robot state space was required in addition to the visual demonstrations, thus violating the purpose of visual imitation.
Yang \etal~\cite{yang2022viptl} proposed a transporter-based representation learning model to extract keypoints from the task-agnostic human and robot play data. Building the similarity function of such a model requires robot execution videos with a similar view setup as the demonstration videos. The same requirement applies to the approaches presented in~\cite{Pari2022TheSE,torabi_imitation_2019,torabi_generative_2019} and prevents robots from learning from human demonstrations taken from a very different viewpoint. 
Sermanet \etal~\cite{sermanet_timecontrastive_2018} proposed Time-Contrastive Networks (TCN) to learn viewpoint-invariant latent representations of the scene. 
This approach requires a large number of demonstration videos and robot play videos to build the correspondence between human and robot arms, which makes the approach embodiment-dependent.
Similar to our paper, Karnan \etal~\cite{Karnan2022VOILAVI} proposed to leverage task-agnostic keypoint detection algorithms for vehicle navigation tasks. 
This approach requires storing the demonstration video and searches the closest demonstration image 
for reward construction. 
This reduces the number of demonstrations compared to the pixel-level context translations of~\cite{smith_avid_2020,liu2018imitation,Dwibedi2018LearningAR,Adversarial2022}. However, by overlooking the different types of geometric constraints that the keypoints are subject to, this approach suffers from averaging problem similar to~\cite{simeonov_neural_2021} (see~\cref{sec:task_constraints} for details).
In contrast, K-VIL uses dense point-based object representation and correspondence detection to align demonstrations recorded from different viewpoints. This significantly reduces the required number of demonstrations. Moreover, K-VIL explicitly extracts viewpoint- and embodiment-independent scene structure and task constraints, thus addressing the average problem and achieving better extrapolation capability in fine-grained manipulation tasks.

\subsubsection{Hierarchy of scene structure.} 
The variance across demonstrations was used to efficiently select appropriate local frames from some candidates in several imitation learning frameworks as a solution to extract hierarchical scene structure~\cite{muhlig_automatic_2009,ureche_task_2015}.
Representing the learned task in such local frames was shown to facilitate the transfer of skills between different embodiments and the design of control policies.
However, in the absence of visual sensory input, the candidates were manually defined at object level in~\cite{muhlig_automatic_2009,ureche_task_2015}. In our work, we instead show that combining dense visual descriptors and a variance-based criterion allows for the efficient extraction of keypoints and local frames at a fine-grained level.

\subsubsection{Task constraints}\label{sec:task_constraints}
Early works on visual servoing~\cite{goos_hierarchical_1999, hespanha_what_1999, gridseth_vita_2016} hand-crafted task constraints as simple geometric constraints (e.g., point-to-point, point-to-line).
To represent more complex constraints, Sieb \etal~\cite{sieb2020graph} proposed visual entity graphs (VEGs) based on DON to disentangle the scene structure into multiple levels, including objects, parts, and points.
A path integral policy was then trained on the similarity loss between the VEGs learned from the demonstrator and the VEGs observed by the robot.
The task constraints are implicitly learned in VEGs, similarly to the neural pose descriptors in~\cite{simeonov_neural_2021}, and are therefore averaged when large shape or pose variations occur in the demonstrations.
To address this issue, Jin \etal~\cite{jin_geometric_2020,jin_generalizable_2022} introduced an explicit representation of geometric constraints
using visual geometric skill kernels and graph neural networks, which generalized better to categorical objects.
However, this approach requires $\sim30$ demonstrations to learn a generalizable representation, since both task correspondences and geometric constraints need to be learned in the graph structure.
In this paper, we exploit the correspondence detection of DON and the variation information to jointly extract explicit, sparse keypoints and endow them with geometric constraints of various types (see Section~\ref{sec:pce}).
This allows us to learn generalizable skills from only a few demonstration videos while alleviating the averaging problem of~\cite{sieb2020graph}.
Moreover, our task representation allows us to replace the RL policy used, e.g., in~\cite{sieb2020graph}, with simpler movement primitives that reproduce the demonstrated keypoint motions and adapt to new goal configurations.

\section{Background}
\label{sec:background}
In this section, we introduce the dense visual correspondence models, the principal manifold estimation algorithm, and the VMPs, which are essential building blocks of K-VIL.

\subsection{Dense Visual Correspondence}
\label{sec:don}
Dense Object Net (DON)~\cite{florencemanuelli2018dense} maps an RGB image \smalleq{$\image \in \mathbb{R}^{W\times H\times C}$} to a dense descriptor image \smalleq{$\image_{\bar{D}} \in\mathbb{R}^{W\times H\times \bar{D}}$}, where \smalleq{$W, H, C$} denote the width, height and the number of channels of the image, and \smalleq{$\bar{D}$} is the dimension of the descriptor space. 
Therefore, each pixel of the input image is represented by a \smalleq{$\bar{D}$}-dimensional descriptor \smalleq{$\bd \in \mathbb{R}^{\bar{D}}$}. 
To train a DON model on an object category, multiple views of posed RGB images of multiple instances of this object category are first collected. The object meshes are then reconstructed using any state-of-the-art scene reconstruction method. 
The reconstructed meshes are then exploited to automatically acquire object masks and retrieve dense correspondence signals, which are used to train the model.
A fully trained DON model maps similar local patches of two images of the categorical objects to patches in the descriptor space with similar descriptors.
In other words, the dense visual correspondence between two pixels is detected if the distance between their descriptors is smaller than a certain threshold.
For example, the spout of the kettle in different image frames is mapped to similar descriptors.

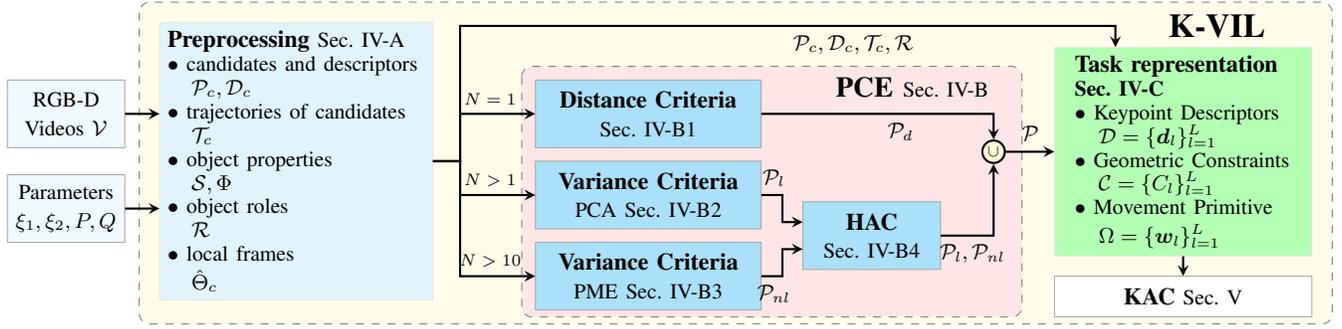
\begin{figure*}[t]
    \centering
    \begin{tikzpicture}[>=stealth, thick,scale=0.9, every node/.style={transform shape}]
        \coordinate (orig)   at (-3,0.7);         
        \coordinate (prepro) at (0.4,0);          
        \coordinate (dist)   at (5.6, 0.7);       
        \coordinate (pca)    at (5.6, -0.5);         
        \coordinate (pme)    at (5.6, -1.7);      
        \coordinate (hac)    at (8.9, -1.1);     
        \coordinate (out)    at (13.5, 0.14);        
        \coordinate (param)  at (-3, -0.7);       
        \coordinate (u_pce)  at (7.6, -1.1);
        \coordinate (u_all)  at (10.7, 0.14);
        \coordinate (kac)    at (13.5, -2.0);      
        
        \node at (14,2) {\Large{{\textbf{K-VIL}}}};
        \node at (9.5, 1.1) {\large{{\textbf{PCE}}} \small{Sec. \ref{sec:pce}}};
        
        \node[draw=gray!70, minimum width=.8cm, minimum height=1cm, anchor=center, text width=1.5cm, align=center, fill=cyan!5, line width=0.05mm] (A) at (orig) {\small{RGB-D \\ Videos $\mathcal{V}$}};
        
        \node[draw=gray!70, minimum width=.8cm, minimum height=1cm, anchor=center, text width=1.5cm, align=center, fill=cyan!5, line width=0.05mm] (P) at (param) {\small{Parameters \\$\xi_1, \xi_2, P, Q$}};
        
        \node[draw=gray!20, minimum width=3.5cm, minimum height=2cm, anchor=center, text width=3.8cm, align=left, fill=cyan!10, line width=0.05mm] (B) at (prepro) {\textbf{Preprocessing} \small{Sec. \ref{sec:preproc}}\\ \small{
            $\bullet$ candidates and descriptors \\$\ \ \ \setP{c}, \mathcal{D}_c$
            \\ $\bullet$ trajectories of candidates \\$\ \ \ \mathcal{T}_c$
            \\ $\bullet$ object properties \\$\ \ \ \mathcal{S}, \Phi$
            \\ $\bullet$ object roles \\$\ \ \ \mathcal{R}$
            \\ $\bullet$ local frames \\$\ \ \ \hat \Theta_c$
            }
        };
        
        \node[draw=gray!70, minimum width=3.0cm, minimum height=1cm, anchor=center, text width=3.1cm, align=center, fill=cyan!30, line width=0.05mm] (C) at (dist) {\textbf{Distance Criteria} \\\small{Sec. \ref{sec:criteria_dist}}};
        
        \node[draw=gray!70, minimum width=3.0cm, minimum height=1cm, anchor=center, text width=3.1cm, align=center, fill=cyan!30, line width=0.05mm] (D) at (pca) {\textbf{Variance Criteria}\\\small{PCA Sec. \ref{sec:stage_1}}};
        
        \node[draw=gray!70, minimum width=3.0cm, minimum height=1cm, anchor=center, text width=3.1cm, align=center, fill=cyan!30, line width=0.05mm] (E) at (pme) {\textbf{Variance Criteria}\\\small{PME Sec. \ref{sec:stage_2}}};
        
        \node[draw=gray!70, minimum width=1.8cm, minimum height=1cm, anchor=center, text width=1.8cm, align=center, fill=cyan!30, line width=0.05mm] (F) at (hac) {\textbf{HAC}\\\small{Sec. \ref{criteria:hac}}};
    
        \node[draw=gray!20, minimum width=3.8cm, minimum height=1cm, anchor=center, text width=3.2cm, align=left, fill=green!30, line width=0.05mm] (G) at (out) {\textbf{Task representation\\ \small{Sec. \ref{sec:kvil_ext}}} \small{
            \\ $\bullet$ Keypoint Descriptors \\$\ \ \ \mathcal{D} = \{\bd_l\}_{l=1}^L$
            \\ $\bullet$ Geometric Constraints \\$\ \ \ \mathcal{C} = \{C_l\}_{l=1}^L$ 
            \\ $\bullet$ Movement Primitive \\$\ \ \ \Omega = \{\bw_l\}_{l=1}^L$
            }
        };
        
        \node[draw=gray!70, minimum width=3.8cm, minimum height=0.6cm, anchor=center, text width=3.2cm, align=center, fill=white!30, line width=0.05mm] (K) at (kac) {\textbf{KAC} \small{Sec. \ref{sec:kpts_control}}};
        
        \node[draw, fill=yellow!20, circle] (U2) at (u_all) {}; 
        
        \draw[->, thick] (A.east) -- node[above]{} ($(B.180) + (0,0.7)$);
        \draw[->, thick] (P.east) -- node[above]{} ($(B.180) + (0,-0.7)$);
        
        \draw[->, thick] (B.east)  -- ($(B.east) + (0.4,0)$) |- (C.west)
            node[pos=0.7, above]{\scriptsize{$N = 1$}};
        \draw[->, thick] (B.east)  -- ($(B.east) + (0.4,0)$) |- (D.west)
            node[pos=0.7, above]{\scriptsize{$N > 1$}};
        \draw[->, thick] (B.east)  -- ($(B.east) + (0.4,0)$) |- (E.west)
            node[pos=0.7, above]{\scriptsize{$N > 10$}};
            
        \draw[->, thick] (C.east)  -| (U2.90) node[pos=0.3, below]{\small{$\setP{d}$}};
        \draw[->, thick] (D.east)  -| ($(F.west) + (-0.3,0.2)$) node[pos=0.3, above]{\small{$\setP{l}$}} -- ($(F.west) + (0,0.2)$);
        \draw[->, thick] (E.east)  -| ($(F.west) + (-0.3,-0.2)$)node[pos=0.3, below]{\small{$\setP{nl}$}} -- ($(F.west) + (0,-0.2)$);
        
        \draw[->, thick] (F.east)  -| (U2.270)node[pos=0.3, below]{\small{$\setP{l}, \setP{nl}$}};
        \draw[->, thick] (U2.east) -- (G.west)node[pos=0.5, above]{\small{$\setP{}$}};
        
        \draw[->, thick] (B.east)  -- ($(B.east) + (0.4,0)$) -- ($(B.east) + (0.4, 2.0)$)  -| ($(G.north) + (-1.0,0)$) node[pos=0.3, below]{\small{$\setP{c}, \mathcal{D}_c, \mathcal{T}_c, \mathcal{R}$}};
        
        \draw[->, thick] (G.south) -- (K.north)node[pos=0.5, above]{};
        
        \node at (10.7, 0.14) {\scaleeq{0.7}{\cup}};
        
        \begin{pgfonlayer}{background}
        \path[fill=yellow!10,rounded corners, draw=black!50, dashed]
            (B.north west)+(-0.3,0.3) rectangle ($(G.south east)+(0.3,-1.0)$); 
        \end{pgfonlayer}
        
        \begin{pgfonlayer}{background}
        \path[fill=red!10,rounded corners, draw=black!50, dashed]
            (C.north west)+(-0.2,0.2) rectangle ($(F.south east)+(1.2,-0.7)$); 
        \end{pgfonlayer}
        
    \end{tikzpicture}
    \caption{Overview of K-VIL's architecture. After preprocessing the demonstration videos, K-VIL jointly extracts a set of sparse keypoints, a set of keypoint-based geometric constraints, and a set of movement primitive parameters that fully represent the task. The robot then leverages the proposed keypoint-based admittance controller to reproduce the task in novel scenes.
    }
    \label{fig:kvil_diagram}
    \vspace{-2ex}
\end{figure*}

In this paper, we aim at retrieving dense correspondences among different instances of the same object category. 
Moreover, once a set of sparse keypoints is extracted, we aim at identifying these keypoints during the task reproduction on new instances of the same object category using their descriptors.
To do so, we construct a correspondence function $\bp = f_c(\image, \bd)$ using the DON model and the camera intrinsic and extrinsic parameters, which can be used to extract the 3D position of a keypoint represented by the descriptor $\bd$.
Similarly to~\cite{florencemanuelli2018dense,Manuelli2020KeypointsIT}, we use 34-layer, stride-8 ResNet as the DON model and set \smalleq{$\bar{D} = 3$}. We also train Mask R-CNN models~\cite{he_mask_rcnn} with the automatically generated object mask dataset similar to~\cite{sieb2020graph}.
We refer the interested readers to~\cite{florencemanuelli2018dense,Manuelli2020KeypointsIT} for additional details of DON models.
For the case where the human hand is involved in the demonstrated tasks, we treat it as a special object and utilize a hand keypoint detection algorithm, e.g., MediaPipe~\cite{lugaresi_mediapipe_2019}, which provides more robust correspondence detection on human hands.

\subsection{Principal Manifold Estimation (PME)}
\label{sec:pme}
Within K-VIL, we are interested not only in extracting a set of keypoints, but also in learning the constraints that they satisfy in order to represent the task. Specifically, we use geometric constraints, which allow us to restrict the keypoint target positions, e.g., as in \cref{subfig:concept_geom}.
In a $3$-D space, a simple geometric constraint can be viewed as a low-dimensional manifold corresponding to a point, a line, a plane, a curve, or a surface.
In this paper, we leverage the principal manifold estimation (PME) algorithm~\cite{meng_principal_2021} to uncover the geometric constraints as low-dimensional embedding from a set of 3D points varying in time (obtained via DON). 
In PME, the principal manifold is defined as a minimum of the functional with a regularity penalty term derived on a Sobolev space.
Specifically, the PME algorithm minimizes the loss
\begin{equation}\label{eq:pme_loss}
    \small \mathcal{L}(f, \pi_d) = \mathbb{E}\left \Vert \bm{x} - f(\pi_d(\bm{x})) \right \Vert^2 + \lambda \Vert \kappa_f \Vert^2,
\end{equation}
where \smalleq{$\pi_d:$} \smalleq{$\mathbb{R}^D\to \mathbb{R}^d$} is the projection index that maps a random $D$-dimensional vector \smalleq{$\bm{x}$} onto a $d$-dimensional principal manifold with $d<D$, \smalleq{$f:$} \smalleq{$\mathbb{R}^d\to \mathbb{R}^D$} is the reconstruction function, \smalleq{$\Vert \kappa_f \Vert^2$} represents the high-dimensional generalization of the total squared curvature of the principal manifold, and $\lambda\in[0, \infty)$ controls the model complexity. 
Therefore, the former term of the loss represents the reconstruction error, while the latter regularizes the model to avoid overfitting.
Note that PME reduces to linear principal component analysis (PCA) when $\lambda\to\infty$. 
The linearity and the dimension $d$ of the principal manifold determine the subspace type. For example, a nonlinear principal manifold of dimension $d=1$ is a principal curve and corresponds to a curve constraint.
We refer the reader to~\cite{meng_principal_2021} for the details of the PME algorithm.

\subsection{Via-point Movement Primitive (VMP)}
\label{sec:vmp}
In addition to extracting the keypoint constraints, we are interested in learning their motions from human demonstration videos. In imitation learning, motions are often represented by movement primitives. Here, we use via-point movement primitives (VMPs)~\cite{zhou_learning_2019}.
A VMP combines a linear elementary trajectory $h_\text{vmp}$ with a nonlinear shape modulation $f_{\text{vmp}}$, so that
\begin{equation*}
    y(x) = h_{\text{vmp}}(x) + f_{\text{vmp}}(x) = g + x(y_0 - g) + \bm{\psi}(x)^\trsp \bm{w},
\end{equation*}
where $x$ is the canonical variable decreasing linearly from 1 to 0, $y$ and $y_0$ are current and start positions, and $g$ is the target position. 
The shape modulation term is defined as a linear regression model based on $N_k$ squared exponential (SE) kernels $\psi_i(x) = \exp(-h_i(x-c_i)^2), i\in[1, N_k]$, where $h_i, c_i$ are pre-defined constants.
Similarly to probabilistic movement primitives (ProMP)~\cite{paraschos2018using}, VMPs assume that the weight parameter $\bm{w}\sim\mathcal{N}(\bm{\mu}_{\bm{w}}, \bm{\Sigma}_{\bm{w}})$ follows a Gaussian distribution, and thus can be learned via maximum likelihood estimation (MLE).
VMPs provide enhanced extrapolation capability compared to ProMP, as they handle via-points (including start and target positions) adaptation to points that lie out of the demonstrated distributions.
In this paper, we leverage VMPs to learn the demonstrated motion styles of each keypoint and to adapt the corresponding trajectories to via-points identified using the dense correspondence function \smalleq{$f_c$} of DON.
In contrast to control policies based on RL (e.g., \cite{sieb2020graph}) or on visual servoing (e.g., \cite{jin_generalizable_2022}), VMP-based control policies endow K-VIL with flexible temporal scaling and reliable via-point adaptation.

\section{Keypoint-based Visual Imitation Learning}
\label{sec:approach}

In this section, we present the proposed K-VIL approach. 
Given $N$ demonstration videos \smalleq{$\setV{} = \set{V_n}_{n=1}^N$} of a task in $D$-dimensional task space, where \smalleq{$D\in\set{2, 3}$}, K-VIL first preprocesses the RGB-D videos and generates the data required for learning the task. 
This includes densely sampled candidate points, their descriptors and trajectories, the spatial properties and roles of the objects, as well as all potential local frames (see Section~\ref{sec:preproc}). 
A sparse set of keypoints and their geometric constraints 
are then estimated via \emph{principal constraint estimation} (PCE). As detailed in Section~\ref{sec:pce}, our proposed PCE first extracts a set of keypoints and their geometric constraints by leveraging PME algorithms. These algorithms rely on observed distances, as well as on the demonstration variability when several demonstrations are provided.
For the cases where the resulting set contains redundant selections of keypoints,
our PCE then leverages Hierarchical Agglomerative Clustering (HAC) to resolve this redundancy and obtain a final set $\setP{}$ of sparse keypoints. 
As explained in Section~\ref{sec:kvil_ext}, $\setP{}$ is then used to extract the task representation consisting of a set of keypoints defined by visual descriptors \smalleq{$\setD{} = \set{\bd_l}_{l=1}^L$}, their associated geometric constraints \smalleq{$\mathcal{C} = \set{C_l}_{l=1}^L$} and the weights of the via-point movement primitives \smalleq{$\Omega = \set{\bw_l}_{l=1}^L$}, which are then exploited to reproduce the keypoint motions. 
The extracted task representation is finally used by the keypoint-based admittance controller (KAC) presented in Section~\ref{sec:kpts_control} to reproduce the demonstrated skill on the robot. 
The proposed K-VIL approach is shown in Fig.~\ref{fig:kvil_diagram}, and its different steps are detailed next. The main notations are listed in \cref{tab:notation1}.

\begin{table*}[ht]
    \centering
    \footnotesize
    \begin{tabu}{ll|ll}
        \toprule
        Notation & \multicolumn{1}{c|}{Meaning} & Notation & \multicolumn{1}{c}{Meaning} \\ \midrule
        $C$               & geometric constraint                                        & $\xi_1,\xi_2$           & lower and upper thresholds of spatial variability      \\
        $d$               & the intrinsic dimension of a principal manifold             & $\lambda$               & the regularization factor of PME                          \\
        $D, \bar{D}$      & the dimension of the task space, descriptor space           & $\kappa_f$              & the curvature of the principal manifold                   \\ \tabucline[0.5pt gray!30 off 0pt]{3-4}
        $g_1, g_2$        & force scaling parameters                                    & $\bd$                   & the descriptor vector of a keypoint / candidate             \\
        $H$               & the total number of candidate points                        & $\force$                & a force vector                                          \\
        $I$               & the number of objects                                       & $\bh_c$                 & the Coriolis and gravitational force in task space    \\
        $L$               & the number of constraints                                   & $\bk, \dot \bk$         & the position and velocity vector of a keypoint        \\
        $N$               & the number of demonstrations                                & $\bp$                   & a position vector of a point                            \\
        $P$               & the number of points sampled on an object                   & $\bm{w}, \bm{\Sigma}_{\bm{w}}$  & a weights vector of the VMP and its covariance  \\
        $Q$               & the number of neighboring points                            & $\bm{\mu}_{\bm{w}}$     & the mean of $\bm{w}$                                  \\ \tabucline[0.5pt gray!30 off 0pt]{1-2}
        $\setC{}$         & the set of constraints                                      & $\bnu$                  & the explained variance                                    \\
        $\setD{}$         & the set of descriptors                                      & $\vari$                 & the spatial variability                                   \\
        $\lf$             & a local frame                                               & $\bK$                   & diagonal stiffness, damping and inertia matrices \\
        $\hat \theta_\lf$ & the configuration of a canonical local frame                & $\bS$                   & the canonical shape                                   \\
        $\hat\Theta_\lf$  & the set of canonical local frame configurations             & $\traj$                 & a trajectory                                            \\ \tabucline[0.5pt gray!30 off 0pt]{3-4}
        $\manifold{}$     & a manifold                                                  & $\sigma(\cdot)$         & the density force on the $d$-dimensional manifold     \\
        $O, \setO{}$      & an object category, the set of objects                      & $\nabla \sigma(\cdot)$  & the density field                                     \\
        $\setP{}$         & the set of keypoints / candidate points                     & $f_c(\cdot, \cdot)$     & the correspondence function                           \\
        $\gamma, \mathcal{R}$ & the role of the object, the set of object roles         & $f(\cdot)$              & the projection index of a principal manifold                \\
        $\mathcal{S}$     & the set of canonical shapes of all objects                  & $\pi_d(\cdot)$          & the reconstruction function of a principal manifold         \\
        $\setV{}$         & the set of demonstration videos                             & $\bm{\psi}(\cdot)$      & the squared exponential (SE) kernels in VMP           \\
        $\varphi, \Phi$   & the spatial scale, the set of spatial scales                & $\psi_i(\cdot)$         & the squared exponential kernel                        \\
        \bottomrule
    \end{tabu}
    \caption{Summary of K-VIL's notations.}
    \label{tab:notation1}
    \vspace{-2ex}
\end{table*}

\subsection{Preprocessing}
\label{sec:preproc}
As previously mentioned, K-VIL first preprocesses the RGB-D videos $\setV{}$ provided as demonstrations. This is achieved via the following five steps, also depicted in Fig.~\ref{fig:kvil_diagram}.

\begin{figure}[t]
	\setlength{\subfight}{1.2cm}
	\sbox\subfigbox{%
		\resizebox{\columnwidth}{!}{
			\includegraphics[height=\subfight]{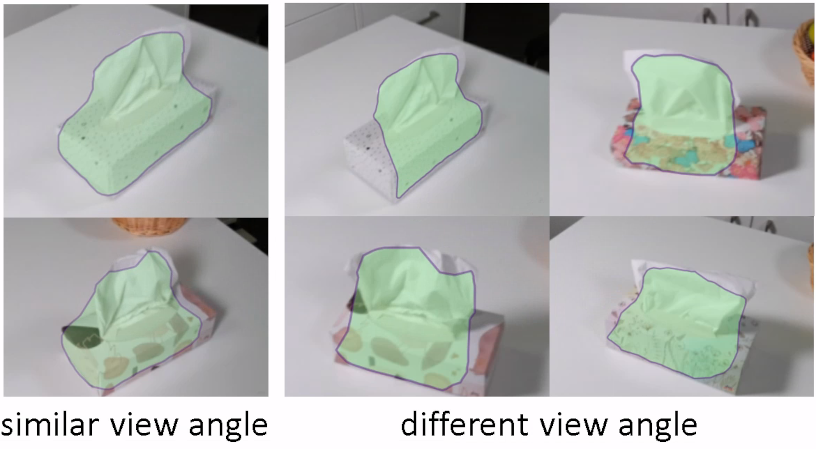}
			\includegraphics[height=\subfight]{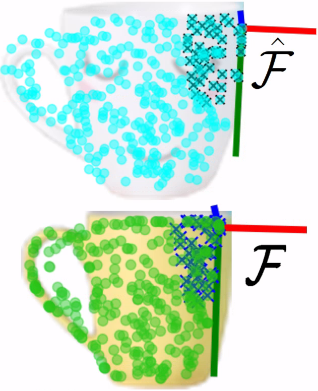}
		}
	}
	\setlength{\subfight}{\ht\subfigbox}
	\centering
	\subcaptionbox{Visible region of a tissue box\label{subfig:intersection}}{
		\includegraphics[height=\subfight]{figure/intersection.png}
	}
	\subcaptionbox{Local frames\label{subfig:local_frame_match}}{
		\includegraphics[height=\subfight]{figure/local_frame_matching.png}
	}
	\caption{\subref{subfig:intersection} Illustration of the visible region of a tissue box. \subref{subfig:local_frame_match} Example of local frame matching on two cups. The canonical shape of the 
	cup category (\emph{top}) is defined on a white cup 
	by the positions of its candidate points ($\textcolor{cyan}{\bullet}$). 
	The correspondence points ($\textcolor{green}{\bullet}$) of the canonical shape are detected on a yellow cup (\emph{bottom}).
	A canonical local frame $\hat{\lf}$ is parametrized by
	$Q=50$ neighboring candidates ($\textcolor{blue}{\times}$). These candidates are then used to find the same local frame $\lf$ on the yellow cup.
	}
	\label{fig:intersection}
    \vspace{-2.0ex}
\end{figure}

\subsubsection{Sampling of candidates}\label{sec:preproc_sample} 
First, we query the list of objects categories \smalleq{$\setO{} = \set{O_i}_{i=1}^I$} involved in the task by feeding the Mask R-CNN model with an image randomly sampled from $\setV{}$. 
From the \emph{visible region} of each object \objecti{}, \smalleq{$P_i$} candidate points are then densely and uniformly sampled and form a set \smalleq{$\setP{i}$}. 
Each candidate point is a potential keypoint or a potential origin of a local frame, and may later be selected as such by K-VIL. Note that we here assume that all relevant points are located in the region on the object surface that is always visible to the imitator (see \cref{subfig:intersection}). 
We denote the set of all candidate points from all objects as \smalleq{$\setP{c} = \bigcup_{i=1}^I \setP{i}$}. 
Their corresponding deep visual feature descriptors are derived from DON~\cite{florencemanuelli2018dense} as \smalleq{$\setD{c} = \set{\bd_h}_{h=1}^{H}$ with $ \bd_h\in\mathbb{R}^3$ and $H=\vert \setP{c}\vert$} the cardinality of $\setP{c}$.

\subsubsection{Trajectories of candidate points}\label{sec:preproc_traj} 
We extract the task space trajectory of all candidate points from the videos using the DON-based correspondence function $f_c(\cdot, \cdot)$ (see \cref{sec:don}), which finds the correspondence pixel of the candidates and maps them to 3D coordinates in the camera local frame.
The obtained trajectories are then smoothed and normalized in time with $T$ timesteps. We obtain a set \smalleq{$\setT{} = \set{\traj_h}_{h=1}^H$} of trajectories of all candidates, where \smalleq{$\traj_h \in \mathbb{R}^{N \times T\times D}$} denotes the trajectory of the $h\nth$ candidate point.
These trajectories are used in the remaining preprocessing steps and in \cref{sec:pce} to extract keypoints and geometric constraints.

\subsubsection{Object properties}\label{sec:preproc_obj}
We define the \emph{canonical shape} \smalleq{$\bS_i \in \mathbb{R}^{P_i \times D}$} of each object category \objecti{} as the positions of all candidates on the object at the first time step of the first demonstration. This notion of canonical shape is illustrated for a cup in Fig.~\ref{subfig:local_frame_match}.
Moreover, we define the \emph{spatial scale} $\varphi_i\in\mathbb{R}$ as the maximum distance between each pair of candidates on the canonical shape, which will be used in \cref{sec:stage_1} to determine object-independent thresholds.

\begin{figure*}[tp]
	\setlength{\subfight}{1.1cm}
	\sbox\subfigbox{%
		\resizebox{\textwidth}{!}{
			\includegraphics[height=\subfight]{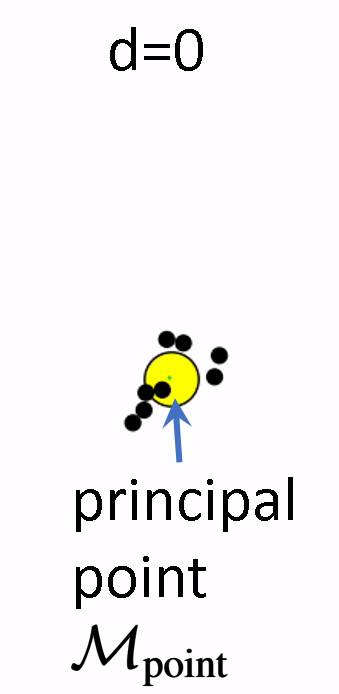}
			\includegraphics[height=\subfight]{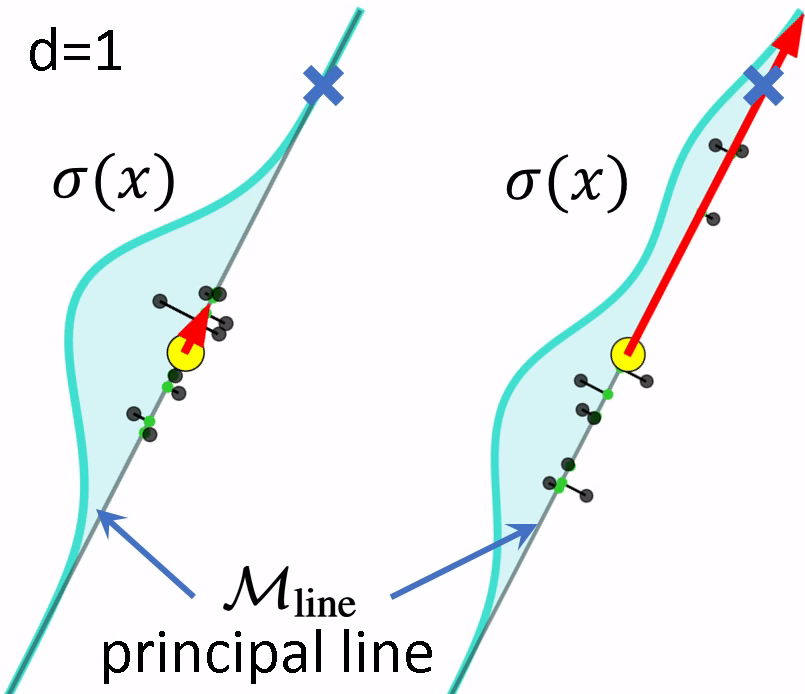}
			\includegraphics[height=\subfight]{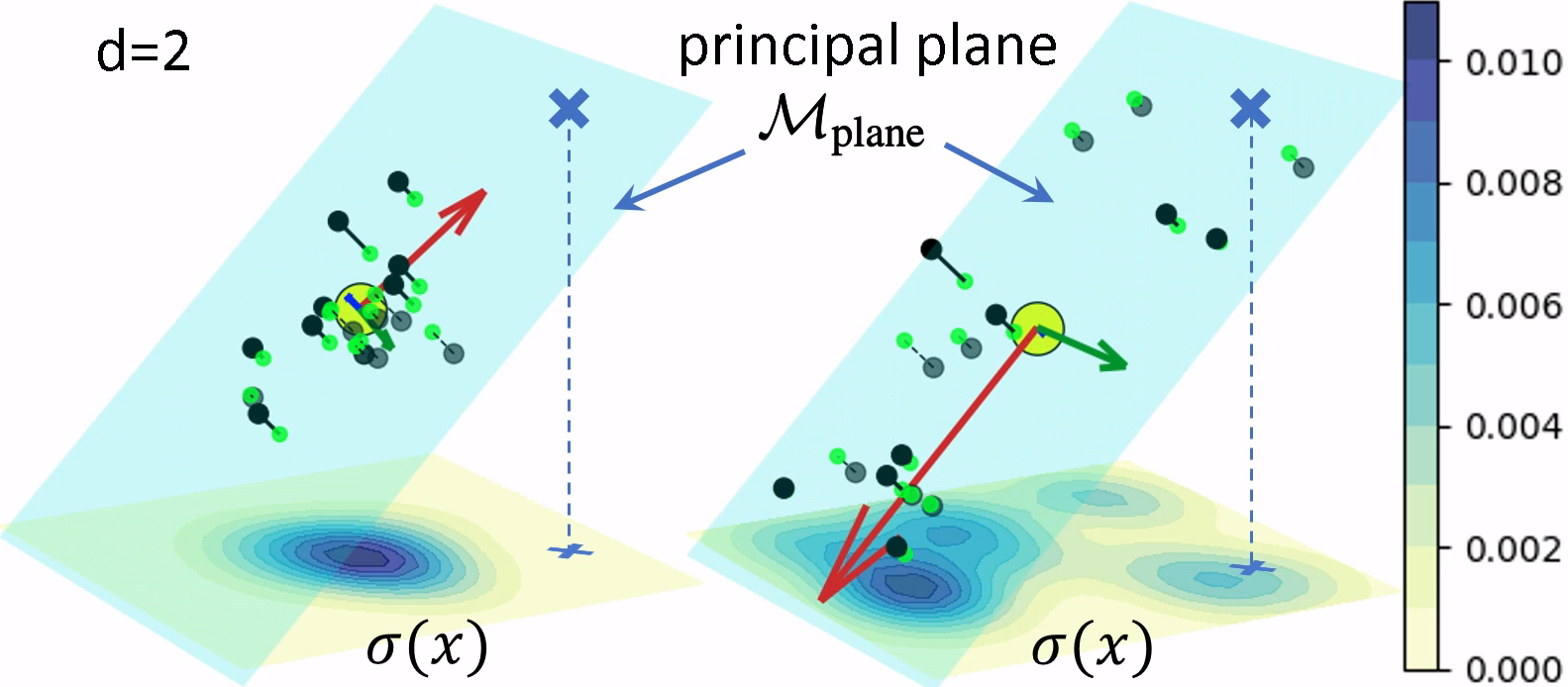}
			\includegraphics[height=\subfight]{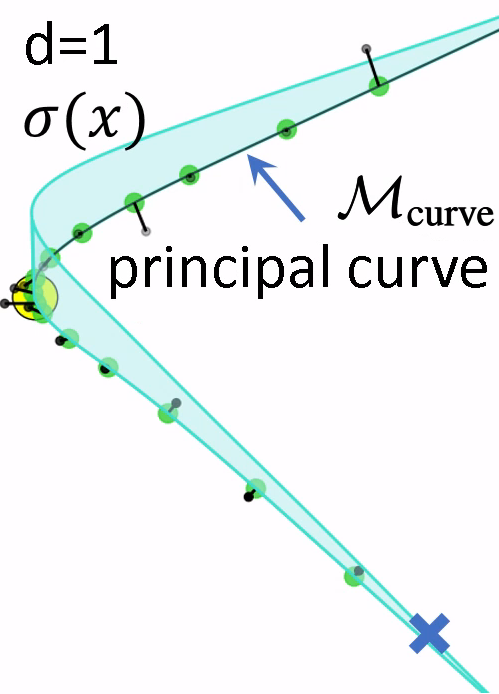}
			\includegraphics[height=\subfight]{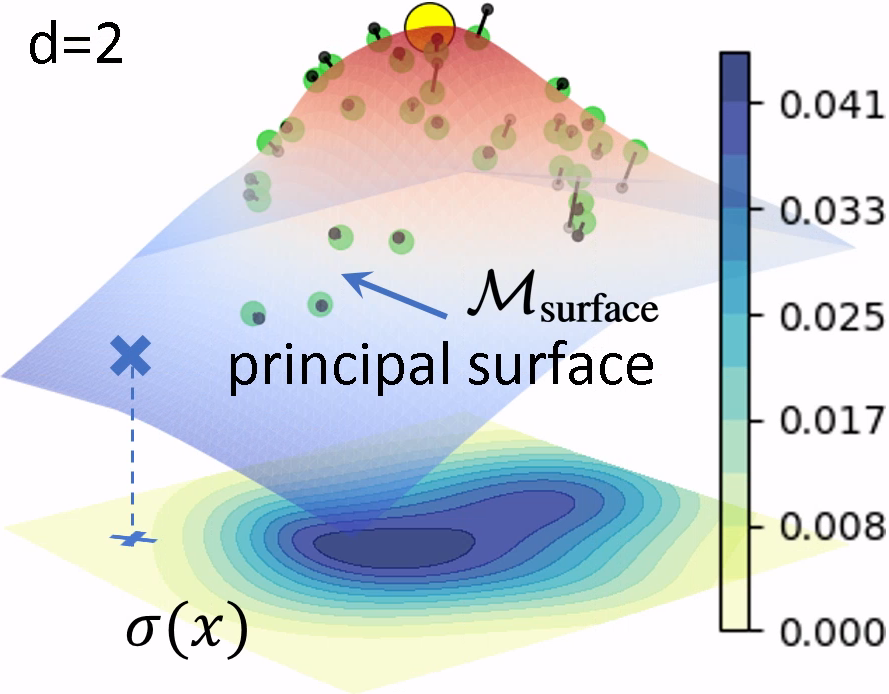}
		}
	}
	\setlength{\subfight}{\ht\subfigbox}
		
	\centering
	
	\subcaptionbox{$\mathsf{p2p}$\label{subfig:pm_p2p}}{
		\includegraphics[height=\subfight]{figure/pm/subfigures_new/p2p.png}
	}
	\subcaptionbox{$\mathsf{p2l}$\label{subfig:pm_p2l}}{
		\includegraphics[height=\subfight]{figure/pm/subfigures_new/p2l.png}
	}
	\subcaptionbox{$\mathsf{p2P}$\label{subfig:pm_p2P}}{
		\includegraphics[height=\subfight]{figure/pm/subfigures_new/p2P.png}
	}
	\subcaptionbox{$\mathsf{p2c}$\label{subfig:pm_p2c}}{
		\includegraphics[height=\subfight]{figure/pm/subfigures_new/p2c.png}
	}
	\subcaptionbox{$\mathsf{p2S}$\label{subfig:pm_p2S}}{
		\includegraphics[height=\subfight]{figure/pm/subfigures_new/p2S.png}
	}
	
	\caption{Five types of geometric constraints. 
	The constraints are obtained from candidate points ($\textcolor{black}{\bullet}$) from \smalleq{$N$} demonstrations. The density function $\sigma(\bx), \bx\in \mathbb R^{d}$ is estimated from the projections ($\textcolor{ForestGreen}{\bullet}$) of the candidate positions on the $d$-dimensional principal manifold.
	We also depict the mean \smalleq{$\bp_m$} ($\textcolor{yellow}{\bullet}$), the spatial variability (\inlinearrow{ForestGreen}, \inlinearrow{red}) on the principal manifold along the principal components, the stress vector $\boldsymbol{s}$ (\sampleline{black}), and examples of extrapolation of the keypoints ($\textcolor{blue}{\boldsymbol{\times}}$) on the manifolds.
	}
	\label{fig:pca_pme}
	\vspace{-2ex}
\end{figure*}

\subsubsection{Object roles}\label{sec:preproc_role}
Geometric constraints do not suffice to entirely represent a task. For example, one of the constraints of a pouring task is the kettle-cup alignment, which could be achieved by moving the cup toward a static kettle. Instead, pouring requires a motion of the kettle. K-VIL addresses this issue by considering the role of the objects for the task at hand.
Namely, we detect object motion saliency similarly to~\cite{muhlig_automatic_2009} to determine the role of the objects \smalleq{$\mathcal{R} = \{\gamma_i \}_{i=1}^{I}$}, where \smalleq{$\gamma_i \in \{\mathsf{master}, \mathsf{slave}\}$}.
The \master{} \objectm{} is the object with the lowest average variance of candidates' trajectory, while other objects
are $\mathsf{slaves}$ \objects{}.
K-VIL accounts for the objects' roles by constructing local frames only on the \master{} and extracting keypoints only on the $\mathsf{slaves}$. 
Therefore, we split the set \smalleq{$\setP{c}$} of all candidates by the object roles to \smalleq{$\setP{m}$} and \smalleq{$\setP{s}$}, denoting the set of candidates on \master{} and \slave{} objects respectively. Similarly, \smalleq{$\setD{c}$} is splitted to \smalleq{$\setD{m}$} and \smalleq{$\setD{s}$}.

\subsubsection{Local frame detection}\label{sec:preproc_frame}

As previously mentioned, K-VIL aims at representing the demonstrated task from an object-centric perspective. This is achieved by defining local frames on the \master{} object. To do so, we initially define one canonical local frame \smalleq{$\hat{\lf}_j$} equal to identity for each candidate point $j\in\setP{m}$ of the \master{} canonical shape (see Fig.~\ref{subfig:local_frame_match}-\emph{top}). 
Each local frame \smalleq{$\hat{\lf}_j$} is assigned the \smalleq{$Q$} closest candidates to $j$, whose positions in \smalleq{$\hat{\lf}_j$} are denoted as the reference values \smalleq{$\bp^*_q$}. 
In other words, \smalleq{$\hat{\lf}_j$} is parameterized by \smalleq{$\hat{\vartheta}_{\hat{\lf}_j} = \{\{\bd_{q}\}_{q=1}^Q, \{\bp^*_q\}_{q=1}^Q\}$}, 
where \smalleq{$\bd_q$} are the descriptors of the \smalleq{$Q$} neighboring candidates. 
These neighboring candidates are then used to detect the same local frame on another instance of the same object category at a different time $t$ (see Fig.~\ref{subfig:local_frame_match}-\emph{bottom}). 
Namely, the local frame \smalleq{$\lf_j$} is detected by minimizing the mean squared displacement of the observed coordinates \smalleq{$\{\bp_q(t)\}_{q=1}^Q$} of the neighboring candidates at time $t$ with their reference values \smalleq{$\{\bp^*_q\}_{q=1}^Q$}, i.e.,
\begin{equation*}
    \textstyle \lf_j(t) = \argmin_{\lf} \sum_{q=1}^Q \left \Vert \bp^*_q - \bp_q(t) \right \Vert^2.
\end{equation*} 
The set of local frames on the $\mathsf{master}$ object is denoted as \smalleq{$\hat \Theta_\lf = \{\hat \vartheta_{\lf_j}: j\in\setP{m}\}$}.

In summary, the preprocessed data for all objects contain the sets \smalleq{$\setP{m}$} and \smalleq{$\setP{s}$} of the candidate points on \master{} and \slave{} objects, respectively, their corresponding descriptors \smalleq{$\setD{m}, \setD{s}$} and trajectories \smalleq{$\setT{}$}, the set \smalleq{$\mathcal{S} = \{\bS_i\}_{i=1}^I$} of the object canonical shapes, the set \smalleq{$\Phi=\{\varphi_i\}_{i=1}^I$} of the object spatial scales, the set \smalleq{$\mathcal{R}$} of the object roles, and the set \smalleq{$\hat \Theta_\lf$} of all local frames.

\subsection{Principal Constraints Estimation (PCE)}
\label{sec:pce}

Given the preprocessed data, our goal is to jointly extract a set \smalleq{$\setP{}$} of keypoints and a set \smalleq{$\mathcal{C} = \{C_l\}_{l=1}^L$} of geometric constraints. 
As shown in \cref{fig:pca_pme}, we consider five basic types of geometric constraints for keypoints in a 3D Cartesian space, namely point-to-point ($\ptop$), point-to-line ($\ptol$), point-to-plane ($\ptoP$), point-to-curve ($\ptoc$) and point-to-surface ($\ptoS$).
The $\ptop$, $\ptol$, and $\ptoP$ constraints are linear and can therefore be estimated by analyzing the variance of the keypoint positions in multiple demonstrations using \emph{PCA}~\cite{halko_finding_2011}.
In contrast, $\ptoc$ and $\ptoS$ are nonlinear geometric constraints, which we estimate with the iterative \emph{PME} (see \cref{sec:pme}). 
Note that more complex constraints such as $\mathsf{colinear, coplanar, parallel}$ and $\mathsf{perpendicular}$ result from combinations of our five basic types of constraints.
To ensure that the constraints are reliably estimated, the criterion of the proposed PCE is adapted to the number of demonstrations. 
Specifically, the single-demonstration case (i.e., one-shot IL) is considered a special case as it does not provide sufficient information to learn generalizable skills. Therefore, we propose heuristically-designed distance-based criteria (\cref{sec:criteria_dist}).
In contrast, when several demonstrations are available (i.e., few-shot IL), the keypoints and geometric constraints are learned based on the variability of the demonstrations (\cref{sec:stage_1,sec:stage_2}).
Moreover, nonlinear constraints are considered only if enough demonstrations are available.

\begin{figure*}[t]
	\centering
	\setlength{\subfight}{1.4cm}
	\sbox\subfigbox{%
		\resizebox{\textwidth}{!}{
			\includegraphics[height=\subfight]{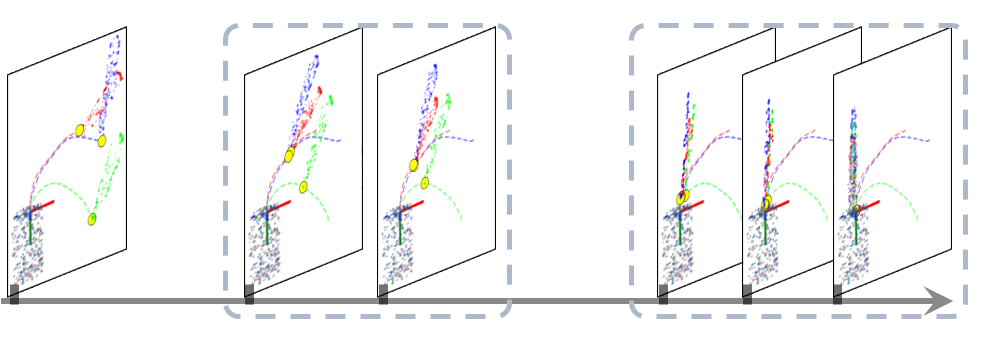}
			\includegraphics[height=\subfight]{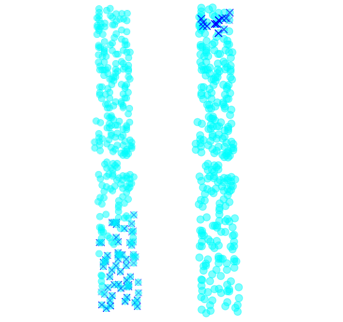}
			\includegraphics[height=\subfight]{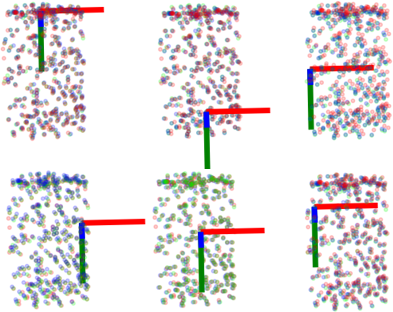}
			\includegraphics[height=\subfight]{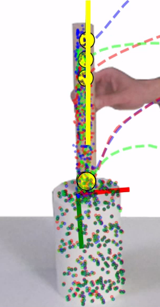}
		}
	}
	\setlength{\subfight}{\ht\subfigbox}
	\centering
	\subcaptionbox{Time clustering\label{subfig:cluster_time}}{
	    \begin{tikzpicture}
    	    \node (image) at (0,0) {
                 \includegraphics[height=\subfight]{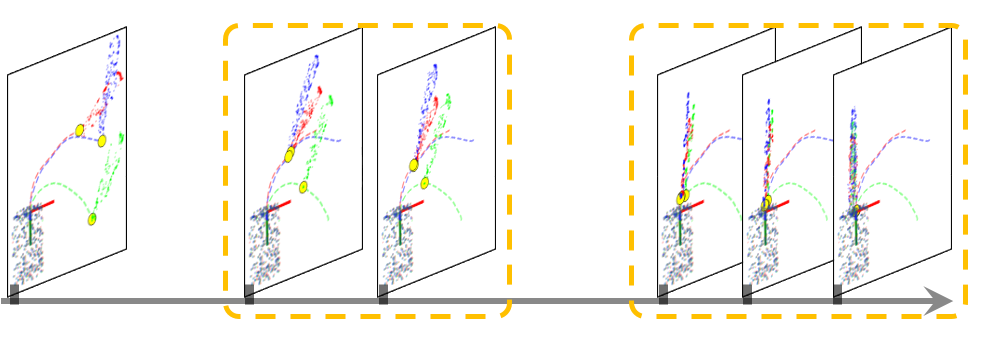}
            };
    	    
    	    \node at (-4.05, -1.35) {\scalebox{0.8}{$0$}};
    	    \node at (-1.95, -1.35) {\scalebox{0.8}{$t-\Delta t_1$}};
    	    \node at (-0.90, -1.35) {\scalebox{0.8}{$t$}};
    	    
    	    \node at ( 1.50, -1.35) {\scalebox{0.8}{$T-\Delta t_2$}};
    	    \node at ( 2.95, -1.35) {\scalebox{0.8}{$T$}};
    	\end{tikzpicture}
	}
	\hspace{-20pt}
	\subcaptionbox{Position clustering\label{subfig:cluster_shape}}{
	    \begin{tikzpicture}
    	    \node (image) at (0,0) {
                 \includegraphics[height=\subfight]{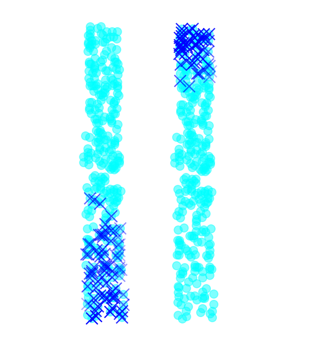}
            };
            
            \draw [pen colour={black!50}, decorate, decoration = {calligraphic brace, raise=5pt, amplitude=5pt}] (-1.4, 1.3) -- (-1.4,-1.2) node[pos=0.5,black]{};
            
            \node at ( -0.4, -1.38) {\scalebox{0.8}{$\bk_1: \ptop$}};
            \node at (  1.0,  1.20) {\scalebox{0.8}{$\bk_2: \ptol$}};
            
            \node[draw, fill=yellow!90, circle, inner sep=0pt,minimum size=5pt, opacity=0.7] (pk2_1) at (-0.38, -1.10) {}; 
            \node[draw, fill=yellow!90, circle, inner sep=0pt,minimum size=5pt, opacity=0.7] (pk2_1) at ( 0.42,  1.15) {}; 
	    \end{tikzpicture}
	}
	\hspace{-10pt}
	\subcaptionbox{Equivalent local frames $\lf$\label{subfig:cluster_lf}}{
	    \begin{tikzpicture}
    	    \node (image) at (0,0) {
                 \includegraphics[height=\subfight]{figure/clustering/cluster_localframe.png}
            };
            \node at ( -0.8, 1.2) {\scalebox{0.8}{$\lf^*$}};
	    \end{tikzpicture}
	}
	\subcaptionbox{Insertion\label{subfig:cluster_result}}{
	    \begin{tikzpicture}
    	    \node (image) at (0,0) {
                 \includegraphics[height=\subfight]{figure/clustering/cluster_result.png}
            };
            \node at ( 0.6, -0.4) {\scalebox{0.8}{$\lf^*$}};
            \node at (-0.3, -0.1) {\scalebox{0.8}{$\bk_1$}};
            \node at (-0.3,  1.0) {\scalebox{0.8}{$\bk_2$}};
	    \end{tikzpicture}
	}
	\caption{Hierarchical agglomerative clustering for inserting a stick into a paper roll (see also \cref{sec:eval} for the task description). 
	\subref{subfig:cluster_time} The selected candidates are first clustered in time (\sampleline{dashed, orange}) to identify the adjacent timesteps. 
	\subref{subfig:cluster_shape} The candidates (\textcolor{blue!80}{$\times$}) in each time cluster (e.g., here in the last time cluster \smalleq{$[T-\Delta t_2, T]$}) are then clustered based on their positions in the canonical shape (\inlinekpt{cyan!50}{fill=cyan!50}) of the stick for each constraint (here $\ptop$ and $\ptol$). For each position cluster, the keypoint (\inlinekpt{black!80}{fill=yellow,opacity=70}) with the lowest variability is finally selected. 
	\subref{subfig:cluster_lf} Since the paper roll has no shape variation, i.e., all canonical local frames are equivalent, the closest frame $\lf^*$ to the selected keypoints is selected.
	\subref{subfig:cluster_result} Final task representation (see also \cref{fig:insertion}).
	}
	\label{fig:cluster}
	\vspace{-1.8ex}
\end{figure*}

\subsubsection{Distance criteria for a single demonstration}\label{sec:criteria_dist}
A single demonstration (\smalleq{$N=1$}) does not provide examples of variations in the demonstrated task, and thus prevents the learning of generalizable skills.
Therefore, in this case, we learn a set of constraints that fully determines the pose of the objects. To do so, we assume that the objects are rigid and extract $3$ keypoints for each \slave{} object in order to fully constrain their position in a $3$-D space. Note that K-VIL can also be applied to $2$-D cases, where $2$ keypoints are sufficient to determine the pose of an object.
We map the trajectories of all candidates on the \slave{} objects into each of the canonical local frames \smalleq{$\lf_{j}(t)$} on the \master{} object at time step $t$, where \smalleq{$j\in\setP{m}$}. 
Therefore, all demonstrations obtained from arbitrary viewpoints are aligned in a common viewpoint defined by the local frame \smalleq{$\lf_{j}(t)$} (see \cref{sec:viewpoint_mismatch}).
Then, \smalleq{$\tilde\traj_j^k(t) \in \mathbb R^{N\times D}$} with \smalleq{$k \in \setP{s}$} represents the positions of the $k\nth$ candidate point in all demonstrations viewed from the $j\nth$ common viewpoint at time step $t$, we use this variable to denote the \emph{candidate positions} in the remaining of the paper.
We observed that, for a variety of daily manipulation tasks, the closest point $k_1$ on the \slave{} object to the \master{} object is often crucial to respect contact or avoid a collision, whereas the furthest point $k_2$, in combination with $k_1$, controls the pose of the object. Motivated by these heuristics, we propose
the following procedure for each \slave{} object.
First, we choose
the local frame \smalleq{$\lf^*(t)$} from the canonical local frames 
as the closest on average to all candidates on the $\mathsf{slave}$ objects.
The two keypoints $k_1, k_2$ on the \slave{} object then correspond to the closest and farthest candidates from the selected local frame \smalleq{$\lf^*(t)$}. 
For a 3-D task space, we select an additional keypoint $k_3$ as the farthest candidate from both $k_1$ and $k_2$.
To fully determine the pose of the \slave{} object, the three keypoints are subject to linear $\ptop$ constraints. 
For each \slave{} object, we finally obtain a set \smalleq{$\setP{d} = \{k_l\}_{l=1}^D$} of keypoints and the corresponding geometric constraints \smalleq{$\mathcal{C} = \{C_l\}_{l=1}^D$}, where \smalleq{$C_l = \{\mathcal{M}_\text{point}(k_l), t, \theta_{\lf^*(t)}, \mathsf{p2p}\}$}
defines a $\ptop$ constraint on a $0$-dimensional principal manifold $\mathcal{M}_\text{point}$ on point $k_l$ at time $t$ in the local frame given by $\vartheta_{\lf^*}$.

\subsubsection{Variance criteria for linear constraints}\label{sec:stage_1}
When several demonstrations (\smalleq{$N>1$}) are available, we leverage their variability to estimate linear constraints beyond $\ptop$. 
To do so, we obtain the candidate positions \smalleq{$\tilde\traj_j^k(t)$} in the canonical local frames \smalleq{$\lf_j(t)$} at time $t$ as described in \cref{sec:criteria_dist} and compute the explained variance \smalleq{$\bnu_j^k(t) = \mathbb{V}_{\scriptscriptstyle{\text{PCA}}}[\tilde\traj_j^k(t)]\in\mathbb{R}^D$} of each candidate using PCA. 
The \emph{spatial variability} $\vari_j^k(t)$ is then defined as
\smalleq{$\vari_j^k(t) = \big (\bnu_j^k(t) \big )^{1/2}/{\tilde \varphi_i}$}, 
with $\tilde \varphi_i$ the spatial scale of the \slave{} object \objecti{} to which the $k\nth$ candidate belongs. 
In contrast to the explained variance, spatial variability removes dependencies on the object size. This allows us to empirically define two object-agnostic lower and upper thresholds $\xi_1,\xi_2$ to identify appropriate linear geometric constraints based on the computed spatial variability. 
Namely, the constraints of each candidate $k$ are determined by the following three conditions:
\begin{enumerate*}[label=(\roman*)]
    \item $\eta_{j,1}^k(t) < \xi_1$ implies a low spatial variability in the first components, as well as across all other dimensions since all the components of the spatial variability are ranked in decreasing order, i.e., $\eta_{j,e}^k>\eta_{j,e+1}^k$, where $e = \set{1, 2}$. This also means that the position of the candidate $k$ remains close to a fixed point \smalleq{$\mathcal{M}_{\text{point}}$} across all demonstrations. Therefore, $k$ is subject to a $\ptop$ constraint;
    \item $\eta_{j,2}^k(t) < \xi_1$ and $\eta_{j,1}^k(t) > \xi_2$ imply that $k$ is constrained on a line \smalleq{$\mathcal{M}_{\text{line}}$} along the first component, i.e., $k$ is subject to $\ptol$ constraint;
    \item Similarly, $\eta_{j,3}^k(t) < \xi_1$ and $\eta_{j,2}^k(t) > \xi_2$ indicate that $k$ is constrained on a plane \smalleq{$\mathcal{M}_{\text{plane}}$} going through the first two components, i.e., $k$ is subject to a $\ptoP$ constraint.
\end{enumerate*}
Any spatial variability \smalleq{$\vari_j^k(t)$} satisfying the above conditions indicates the joint selection of the $k\nth$ candidate, the $j\nth$ local frame and $t\nth$ time step.
All candidates selected as such form a set $\mathcal{P}_l$ of keypoints subject to linear geometric constraints. 
Note that, due to the fact that two distinct points define a line and three non-collinear points define a plane, we learn $\ptol$ constraints when $N > 2$ and of $\ptoP$ constraints when $N > 3$.

\subsubsection{Variance criteria for nonlinear constraints}\label{sec:stage_2}
The linear constraints may not suffice to represent a given task accurately despite them being easily estimated from a few demonstrations. 
For instance, the pouring task of \cref{subfig:concept_geom} requires a point-to-curve constraint.
Therefore, we additionally estimate nonlinear ($\ptoc$ and $\ptoS$) constraints with PME. In the following, a set \smalleq{$\setP{s}$} of all candidate points on $\mathsf{slave}$ objects that do not satisfy any linear constraints are considered as potential candidates for nonlinear constraints.
In our case, we replace the random $D$-dimensional vector $\bm{x}$ in \eqref{eq:pme_loss} with the candidate point $\tilde \traj_j^k(t)$ on the demonstrated trajectory $\tilde\traj_j^k$ at time step $t$, so that the PME loss in \cref{sec:pme} becomes
\begin{equation*}
    \small \mathcal{L}(f, \pi_d) = \mathbb{E}\left \Vert \tilde \traj_j^k(t) - f(\pi_d(\tilde \traj_j^k(t))) \right \Vert^2 + \lambda \Vert \kappa_f \Vert^2, \quad k\in \setP{s}.
\end{equation*}
After obtaining \smalleq{$\pi_d$} from PME, we compute the projections of candidates onto the manifold, i.e., \smalleq{$\hat \traj_j^k(t) = \pi_d(\tilde{\traj}_j^k(t))$}, where \smalleq{$\hat \traj_j^k(t)\in\mathbb{R}^{N\times d}$}.
Then, analogously to \cref{sec:stage_1}, we define the explained variance $\nu^k_{j,\scriptscriptstyle{\parallel}}(t)$ in the tangential direction of the principal manifold 
as the variance of the projections, i.e., \smalleq{$\nu^k_{j,\scriptscriptstyle{\parallel}}(t) = \mathbb{V}[\lVert \hat \traj_j^k(t) \rVert] \in \mathbb{R}$}. 
The explained variance $\nu^k_{j,\scriptscriptstyle{\perp}}(t)$ in the orthogonal direction 
corresponds to the variance of the length of the stress vectors \smalleq{$\boldsymbol{s} = \tilde{\traj}_j^k(t) - f(\hat \traj_j^k(t))$}, i.e., \smalleq{$\nu^k_{j,\scriptscriptstyle{\perp}}(t) = \mathbb{V}[\lVert \boldsymbol{s} \rVert ] $}. 
Similar to the linear case, the spatial variability is defined as \smalleq{$\eta_{j,z}^k(t) = \sqrt{\nu_{j,z}^k(t)}/{\tilde \varphi_i}, z\in \{\perp, {{\parallel}}\}$}.
The set $\mathcal{P}_{nl}$ of keypoints subject to nonlinear geometric constraints is then selected as 
\begin{equation*}
    \mathcal{P}_{nl} = \left \{ k \mid \nu^k_{j,\scriptscriptstyle{\perp}}(t) < \xi_1, \nu^k_{j,\scriptscriptstyle{\parallel}}(t) > \xi_2,  k \in \setP{s} \right \}.
\end{equation*}
The type of the geometric constraints is determined by the intrinsic dimension $d$ of the learned principal manifold, i.e.,  $d=1$ and $d=2$ indicate a $\ptoc$ and a $\ptoS$ constraint, respectively. 
Notice that, in order to guarantee their reliable estimation, nonlinear constraints are considered only when enough demonstrations ($N>10$) are available. 

\subsubsection{Hierarchical Agglomerative Clustering (HAC)}
\label{criteria:hac}
As explained in \cref{sec:stage_1}, each selected candidate point $k$ in the resulting sets of linear and nonlinear constraints $\mathcal{P}_{l}$ and $\mathcal{P}_{nl}$ corresponds to jointly selected time step $t$ and local frame $j$. 
Redundancy may occur due to adjacent timesteps, neighboring keypoints, or equivalent local frames.
To resolve this redundancy, we first cluster the keypoints in time using Hierarchical Agglomerative Clustering (HAC) to identify adjacent timesteps.
\cref{fig:cluster} shows an example of HAC for an insertion task. 
We then use HAC again to cluster the keypoints within each time cluster based on their positions in the canonical shape of the $\mathsf{slave}$ object, thus identifying neighboring keypoints. 
The redundancy is finally resolved by keeping only the keypoint with the lowest variability to represent each position cluster. This keypoint is selected for its robustness against sensor and correspondence detection noise.
If a selected keypoint at a selected time step is subject to multiple constraints represented in different local frames, we select the closest local frame to the keypoint on average. 
In summary, the proposed PCE retrieves a sparse set of $L$ keypoints as the union \smalleq{$\setP{} = \setP{d} \cup \setP{l} \cup \setP{nl}$} and their associated (non)linear constraints \smalleq{$\mathcal{C} = \{C_l\}_{l=1}^L$}, which are exploited to represent the task as explained next.

\subsection{Extraction of K-VIL's complete task representation}
\label{sec:kvil_ext}
\begin{figure}[t]
	\centering
	\resizebox{0.95\columnwidth}{!}{
        \begin{tikzpicture}
            \node (image) at (0,0) {
                 \includegraphics[width=\columnwidth]{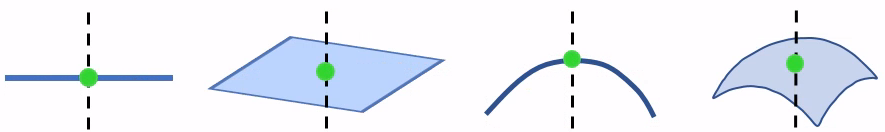}
            };
            \node at (-4, 0.5) {$\manifold{\text{line}}$};
            \node at (-2, 0.5) {$\manifold{\text{plane}}$};
            \node at (0.5, 0.5) {$\manifold{\text{curve}}$};
            \node at (2.5, 0.5) {$\manifold{\text{surface}}$};
        \end{tikzpicture}
    }

	\caption{Orthogonal direction (\sampleline{dashed,black}) to the principal manifolds.}
	\label{fig:ortho_dir}
	\vspace{-5ex}
\end{figure}

\begin{figure}[t]
	\centering
    \begin{tikzpicture}
        \node (image) at (0,0) {
             \includegraphics[width=0.48\columnwidth]{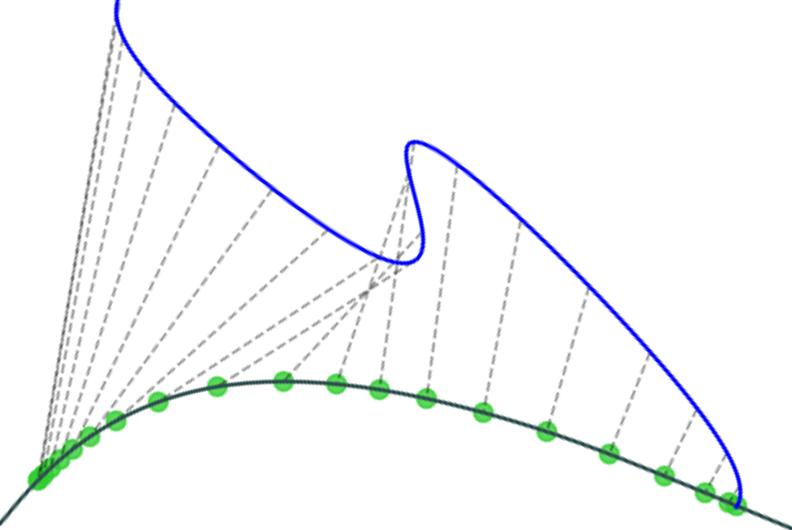}
        };
        \node (image) at (0.48\columnwidth,0) {
            \includegraphics[width=0.5\columnwidth]{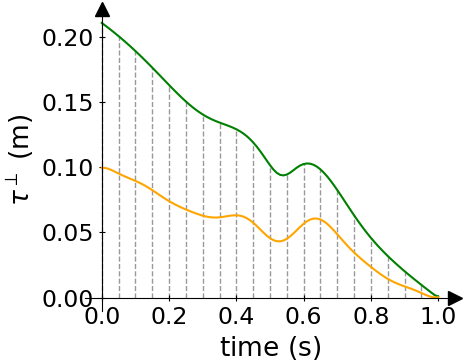}
        };
        \node at (0, 1) {$\textcolor{blue}{\tilde\traj_l}$};
        \node at (0, -1) {$\manifold{\text{curve}}$};
        \node at (-0.9, 1.3) {$t=0$};
        
        \node at (0.48\columnwidth, 1) {$\textcolor{ForestGreen}{\tilde\traj_{l,\perp}}$};
        \node at (0.48\columnwidth, 0) {$\textcolor{orange}{\traj_{l,\perp}}$};
        
        \draw [-{Stealth},red,opacity=0.6,densely dashed] (-1.15, 0.2) to [out=30,in=150] ($(0.48\columnwidth, 0)+(-0.42, 0)$);
        \draw [-{Stealth},red,opacity=0.6,densely dashed] (-0, -0.2) to [out=-15,in=-165] ($(0.48\columnwidth, 0)+(0.68, -0.7)$);
    \end{tikzpicture}

	\caption{Projected and reproduced trajectories using a VMP for a $\ptoc$ constraint.
	The demonstrated trajectory $\textcolor{blue}{\tilde\traj_l}$ (\sampleline{blue}) is projected at each time step in the orthogonal direction (\sampleline{dashed,black}) of the principal manifold $\manifold{\text{curve}}$. 
	The projected trajectory $\textcolor{ForestGreen}{\tilde\traj_{l,\perp}}$ (\sampleline{ForestGreen}) is used to train movement primitives, which is then used to reproduce trajectories, \eg, $\textcolor{orange}{\traj_{l,\perp}}$ (\sampleline{orange}) with a new start position at \SI{0.1}{\meter} and goal position at \SI{0}{\meter}. The arrows (\inlinearrow{red,opacity=0.7, densely dashed}) mark the corresponding projected trajectory between the 3D and 2D plots at two timesteps.}
	\label{fig:traj_projection}
	\vspace{-3ex}
\end{figure}

\begin{figure}[t]
	\centering
	\begin{subfigure}{0.48\columnwidth}
	    \begin{tikzpicture}
            \node (image) at (0,0) {
                 \includegraphics[width=\textwidth]{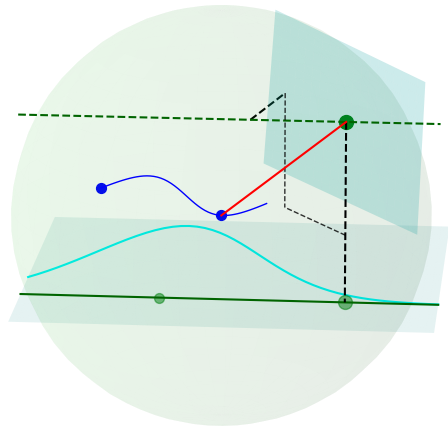}
            };
            \coordinate (k1)                 at (0, 0);
            \coordinate (k2)                 at (1.15, 0.85);
            \coordinate (k2_star)            at (1.15, 0.3);
            \coordinate (k2_proj)            at (1.14, -0.84);
            \coordinate (f2_approach)        at (1.15, -0.24);
            \coordinate (f2_sigma)           at (0.2,  0.88);
            \coordinate (f2_sigma_proj)      at (0.55,  1.13);
            \coordinate (f2)                 at (0.55,  0.05);

            \draw[-{Stealth[length=3mm, width=2mm]}, thick, LimeGreen] (k2) -- (f2_approach);
            \draw[thick, red] (k1) -- (k2);
            \draw[-{Stealth[length=3mm, width=2mm]}, thick, Purple] (k2) -- (f2_sigma);
            \draw[-{Stealth[length=3mm, width=2mm]}, thick, Purple] (k2) -- (f2_sigma_proj);
            \draw[-{Stealth[length=3mm, width=2mm]}, thick, ForestGreen] (k2) -- (f2);
            \draw[-{Stealth[length=3mm, width=2mm]}, thick, Purple] (k2_proj) -- ($(f2_sigma) - (k2) + (k2_proj)$);
            
            \node[draw, fill=ForestGreen!100, circle, inner sep=0pt,minimum size=4pt] (pk2) at (k2) {}; 
            \node[draw, fill=blue!100, circle, inner sep=0pt,minimum size=4pt] (pk1) at (k1) {}; 
            \node[draw, fill=ForestGreen!100, circle, inner sep=0pt,minimum size=4pt] (pk2_star) at (k2_star) {}; 
            \node[draw, fill=ForestGreen!100, circle, inner sep=0pt,minimum size=4pt] (pk2_proj) at (k2_proj) {}; 
            \node at ($(k2)             + ( 0.3,  0.2)$) {$\textcolor{ForestGreen}{\bk_2}$};
            \node at ($(k1)             + ( 0.0, -0.3)$) {$\textcolor{blue}{\bk_1}$};
            \node at ($(k2_star)        + ( 0.3,  0.0)$) {$\textcolor{ForestGreen}{\bk_2^*}$};
            \node at ($(k2_proj)        + ( 0.0, -0.3)$) {$\textcolor{ForestGreen}{\bk_2'}$};
            \node at ($(f2_approach)    + ( 0.3,  0.2)$) {$\textcolor{LimeGreen}{\force_a}$};
            \node at ($(f2_sigma)       + (-0.1, -0.2)$) {$\textcolor{Purple}{\force_\sigma}$};
            \node at ($(f2_sigma) - (k2) + (k2_proj) + (0.2, -0.3)$) {$\textcolor{Purple}{\force_\sigma}$};
            \node at ($(f2_sigma_proj)  + (-0.3,  0.1)$) {$\textcolor{Purple}{\force'_\sigma}$};
            \node at ($(f2)             + ( 0.1, -0.2)$) {$\textcolor{ForestGreen}{\force_2}$};
            
            \node at (-1.5, -0.2) {\scaleeq{0.85}{\sigma(\bm{x})}};
            \node at (-1.5, -1) {\scaleeq{0.85}{\manifold{\text{line}}}};
            \node at (-1.5, 1.2) {\scaleeq{0.85}{\manifold{\text{line}}'}};
            \node [rotate=-25] at (0.85, 1.5) {\scaleeq{0.8}{T_x\manifold{\text{s}}}};
            \node at (0, -1.7) {\scaleeq{0.85}{\manifold{\text{s}}}};
        \end{tikzpicture}
        \vspace{-4ex}
	\end{subfigure}
	\begin{subfigure}{0.48\columnwidth}
	    \begin{tikzpicture}
            \node (image) at (0,0) {
                \includegraphics[width=\textwidth]{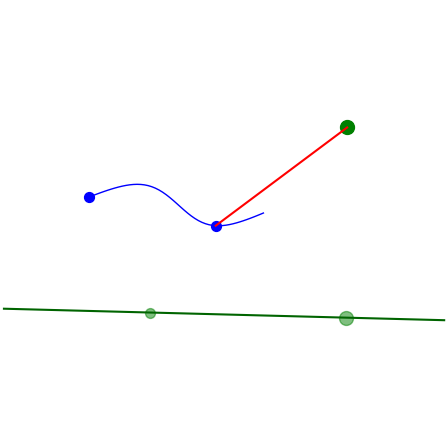}
            };
            \coordinate (k1)                 at (-0.1,  -0.1);
            \coordinate (k1_star)            at (-0.3,  -0.03);
            \coordinate (k2)                 at ( 1.17,  0.82);
            \coordinate (f2)                 at ( 0.55,  0.05);
            \coordinate (f1)                 at (-0.8,   0.1);
            \coordinate (k1_goal)            at (-1.25,   0.15);
            \coordinate (k2_proj)            at (1.13, -0.84);
            \coordinate (k2_goal)            at (-0.1, -0.95);
            \coordinate (pm)                 at (-0.7, -0.93);
            
            \draw[-{Stealth[length=3mm, width=2mm]}, thick, ForestGreen] (k2) -- (f2);
            \draw[-{Stealth[length=3mm, width=2mm]}, thick, blue] (k1) -- (f1);
            \draw[thick, red] (k1) -- (k2);
            \draw[thick, red] (k1_goal) -- (k2_goal);
            \draw[thick, red, dashed] (k1_goal) -- (pm);
            
            \node[draw, fill=black!100, circle, inner sep=0pt,minimum size=4pt] (pk2) at ($(k2)!0.5!(k1)$) {}; 
            \node[draw, fill=ForestGreen!100, circle, inner sep=0pt,minimum size=4pt] (pk2) at (k2) {}; 
            \node[draw, fill=ForestGreen!100, circle, inner sep=0pt,minimum size=4pt] (pk2_goal) at (k2_goal) {}; 
            \node[draw, fill=ForestGreen!100, circle, inner sep=0pt,minimum size=4pt] (p_pm) at (pm) {}; 
            \node[draw, fill=blue!100, circle, inner sep=0pt,minimum size=4pt] (pk1) at (k1) {}; 
            \node[draw, fill=blue!100, circle, inner sep=0pt,minimum size=4pt] (pk1_goal) at (k1_goal) {}; 
            \node[draw, fill=blue!100, circle, inner sep=0pt,minimum size=4pt] (pk1_star) at (k1_star) {}; 
            
            \node at ($(k2)!0.5!(k1)    + (-0.1,  0.35)$) {$\textcolor{black}{\bar\bk}$};
            \node at ($(k2)             + ( 0.3,  0.2)$) {$\textcolor{ForestGreen}{\bk_2}$};
            \node at ($(k2_goal)        + ( 0.0, -0.3)$) {$\textcolor{ForestGreen}{\bk_2^g}$};
            \node at ($(pm)             + ( 0.0, -0.3)$) {$\textcolor{ForestGreen}{\bk_2^m}$};
            \node at ($(k1)             + ( 0.0, -0.3)$) {$\textcolor{blue}{\bk_1}$};
            \node at ($(k1_star)        + ( 0.0, 0.3)$)  {$\textcolor{blue}{\bk_1^*}$};
            \node at ($(k1_goal)        + (-0.05, 0.4)$)  {$\textcolor{blue}{\bk_1^g (\bk_1^m)}$};
            \node at ($(f2)             + ( 0.1, -0.2)$) {$\textcolor{ForestGreen}{\force_2}$};
            \node at ($(f1)             + ( 0.1, -0.3)$) {$\textcolor{blue}{\force_1}$};

        \end{tikzpicture}
        \vspace{-4ex}
	\end{subfigure}
	\caption{
	Illustration of the attraction and density forces when the keypoints $\textcolor{blue}{\bk_1}$ and $\textcolor{ForestGreen}{\bk_2}$ are subject to $\ptop$ and $\ptol$ constraints, respectively.
	\textbf{Left:} The approach force $\force_a$ of $\bk_2$ is computed by the virtual spring-damper system between the attractor $\bk_2^*$ and $\bk_2$. 
	The density force $\force_\sigma$ is then projected onto the tangent space of the sphere at $\bk_2$.
	The control force of $\bk_2$ is the combination of the attraction and the projected density force.
	\textbf{Right:} $\bk_1$ is controlled by the attraction force $\force_1$ following the attractor $\bk_1^*$ and the VMP (\sampleline{blue}) to reach the target $\bk_1^g$, which coincides with the demonstrated target $\bk_1^m$. 
	Note that the target $\bk_2^g$ of $\bk_2$ does not coincide with $\bk_2^m$ due to object shape variation, i.e., the distance (\sampleline{red}) between $\bk_1$ and $\bk_2$ during reproduction is longer than for the demonstration (\sampleline{dashed,red}).
	}
	\label{fig:kac_line}
        \vspace{-1.8ex}
\end{figure}

While the keypoints and associated constraints estimated in Section~\ref{sec:pce} allow us to understand the demonstrated task, a control policy is additionally required for reproducing the task.
Here, we propose to model the observed keypoints trajectories as VMPs~\cite{zhou_learning_2019}. For our purposes, we train the VMPs from an object-centric perspective and according to the constraints estimated via PCE. 
Specifically, for each keypoint subject to a $\ptop$ constraint, a VMP is trained on its observed trajectory \smalleq{$\tilde{\traj}_l$} retrieved in the corresponding local frame $j$ from time step $1$ to the extracted time step $t$, 
i.e., \smalleq{$\tilde{\traj}_l = (\traj_j^k(1) \ldots \traj_j^k(t))^\trsp$}, 
where $k$ and $l$ indicates that the $k\nth$ candidate point in the dense set $\setP{s}$ is selected as the $l\nth$ keypoint in the sparse set $\setP{}$.
Note that, for the case of intrinsic dimension $d>0$ (i.e., $\ptol,\ptoP,\ptoc,\ptoS$ constraints), the constraint is fulfilled if and only if the corresponding keypoint is placed at the time step $t$ on the principal manifold that defines the constraint. 
Although the location of the keypoint on the manifold does not affect the fulfillment of the constraints, it may influence the similarity between the demonstrated object poses and those obtained in the reproduction.
Therefore, we propose to decompose the control of such keypoint by considering the orthogonal and the tangential direction with respect to the corresponding principal manifold independently (see \cref{fig:ortho_dir}).
The keypoint motion along the orthogonal direction represents the demonstrated style of approaching the principal manifold and guarantees the fulfillment of the constraints, 
while the motion along the tangential direction realizes the extrapolation of the keypoint target position and controls the similarity of the object pose between the demonstrations and the reproduction.
Due to potential large shape variations in the objects used when reproducing the task, the final keypoints' target positions on the principal manifold may not align with the demonstrated targets.
Therefore, we only train the VMP on the keypoint trajectories projected onto the orthogonal direction, i.e., $\tilde\traj_{l,\perp}$.
An example of the projected and reproduced trajectories obtained using the learned VMP in the case of a $\ptoc$ constraint is shown in \cref{fig:traj_projection}.
At each time step during reproduction, we uncover the 3D target position of a keypoint by adding an offset generated by the VMP in the orthogonal direction to the orthogonal projection of the current keypoint onto the principal manifold (see $\bk_2^*$ in \cref{fig:kac_line} and \cref{fig:kac_plane}).
Therefore, by setting the VMP goal to $0$, the keypoints fulfill the corresponding geometric constraints at the end of their trajectory.
The motion of the keypoints along the orthogonal and tangential directions is controlled via the keypoint-based admittance controller presented in the next section.
In summary, K-VIL's final task representation is composed of a set of keypoints represented by their descriptors \smalleq{$\mathcal{D}=\{\bd_l\}_{l=1}^L$}, their associated geometric constraints \smalleq{$\mathcal{C} = \{C_l\}_{l=1}^L$}, and their associated movement primitives encoded via the set of weights $\Omega = \{\bw_l\}_{l=1}^L$.

\section{Keypoint-based Admittance Controller}
\label{sec:kpts_control}
After learning the representation of a given task from demonstrations, we aim at reproducing this task with a robot. This means that the robot should be able to interact with the objects such that their keypoints follow the learned constrained trajectories. This requires filling the gap between K-VIL's task representation and real-time robot controllers. To this end, we propose a Keypoint-based Admittance Controller (KAC), which
\begin{enumerate*}[label=(\roman*)]
    \item \label{kac:ch1}handles variable numbers of extracted keypoints for different tasks;
    \item \label{kac:ch2}enables the extrapolation of keypoint target positions on their learned principal manifolds;
    \item \label{kac:ch3}resolves potential interference between different types of geometric constraints.
\end{enumerate*}
Note that \ref{kac:ch2} and \ref{kac:ch3} are required to handle large object shape variations in the task reproduction. 

Specifically, a KAC associates each keypoint in $\setP{}$ with a virtual spring-damper system, whose attractor is computed via the corresponding VMPs (see Section~\ref{kac:attract}). 
As detailed in \cref{kac:adm}, the sum of the attraction forces of the spring-damper systems of all keypoints is then used as the task-space force command for the robot. This allows the KAC to handle a varying number of keypoints for different tasks. 
Regarding \ref{kac:ch2}, the extrapolation of keypoint target positions subject to a $\ptop$ constraint is not allowed. For non-$\ptop$ constraints, this is achieved by decomposing the control in orthogonal and tangential directions of the learned principal manifolds (see \cref{sec:kvil_ext}).
As a result, the keypoints approach the principal manifolds using the motion profiles learned from the projected trajectories in orthogonal directions. 
The control force generated by the virtual spring-damper system of each keypoint remains orthogonal to the principal manifold at each time step, and the keypoints reach the corresponding geometric constraints when the execution of the VMP finishes.
While this leads to the extrapolation of keypoint target positions on the principal manifold, it does not account for the distance between the demonstrated and extrapolated targets. Therefore, we propose to balance extrapolation and regulation by estimating the density function of the demonstrated targets on the principal manifolds, as described in \cref{kac:density}.
This density is then used to compute an additional force, i.e., the density force, that drives each keypoint toward the demonstrated targets.
Finally, we address the interference issue \ref{kac:ch3} by assigning different priorities to different types of geometric constraints in \cref{kac:priority}.
The different steps of KAC are detailed next.

\subsection{Attraction force}\label{kac:attract} 
Given K-VIL's task representation and a new image frame \smalleq{$\image$} for the task reproduction, we can identify the keypoints representing the task. Namely, their positions \smalleq{$\bk_l \in \mathbb{R}^D$}, with \scaleeq{0.9}{l\in [1, L]}, represented in the root frame \smalleq{$\lf_{r}$} of the robot are obtained using the visual descriptors \smalleq{$\bd_l$} and the DON-based correspondence function \smalleq{$f_c(\image, \bd_l)$} (see \cref{sec:don}). 
The attractor \smalleq{$\bk_l^*$} of the virtual spring-damper system at each time step
is computed for each keypoint by the corresponding VMPs projected onto \smalleq{$\lf_{r}$}. 
The attraction force generated by the virtual spring-damper system 
is then computed as 
\begin{equation*}
\force_{a,l} = \bar \bK_p (\bk_l^* - \bk_l) + \bar \bK_d (\dot\bk_l^* - \dot\bk_l),    
\end{equation*} 
where \smalleq{$\bar \bK_p, \bar \bK_d$} are diagonal stiffness and damping matrices, respectively, and $\dot\bk_k$ and $\dot\bk_l^*$ are the velocity of $\bk_l$ and $\bk_l^*$, respectively. 
As explained in \cref{sec:kvil_ext}, in the case of non-$\ptop$ constraints, the VMPs are trained on trajectories projected in directions that are orthogonal to the principal manifold.
Therefore, the learned VMPs and the attraction forces enable the reproduction of the demonstrated motion patterns in the orthogonal direction. 
This implies that the final positions of the keypoints can be extrapolated anywhere on the principal manifolds, e.g., to satisfy object shape variations and other geometric constraints. 
However, without considering the demonstrated targets on the principal manifold, we may lose important information about successful task execution or a specific style of execution.
We obtain such information using kernel density estimation and provide additional density forces driving the keypoints toward the demonstrated targets.

\begin{figure}[t]
	\centering
	\begin{subfigure}{0.48\columnwidth}
	    \begin{tikzpicture}
            \node (image) at (0,0) {
                 \includegraphics[width=\textwidth]{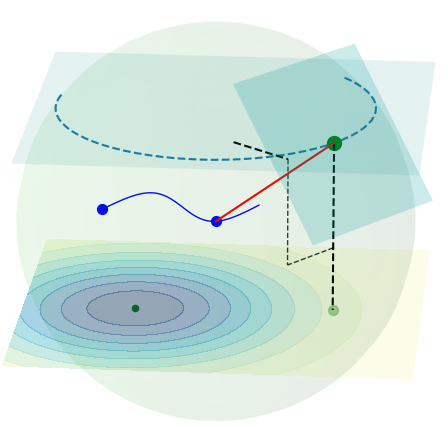}
            };
            \coordinate (k1)                 at (-0.05, -0.05);
            \coordinate (k2)                 at (1.03,  0.65);
            \coordinate (k2_star)            at (1.03,  0.3);
            \coordinate (k2_proj)            at (1.02, -0.87);
            \coordinate (f2_approach)        at (1.03, -0.24);
            \coordinate (f2_sigma)           at (0.06,  0.67);
            \coordinate (f2_sigma_proj)      at (0.56,  0.51);
            \coordinate (f2)                 at (0.6,  -0.49);
            
            \draw[dashed, orange, thick] (0.05, 0.33) -- (1.8, 0.95);  
            
            \draw[-{Stealth[length=3mm, width=2mm]}, thick, LimeGreen] (k2) -- (f2_approach);
            \draw[thick, red] (k1) -- (k2);
            \draw[-{Stealth[length=3mm, width=2mm]}, thick, Purple] (k2) -- (f2_sigma);
            \draw[-{Stealth[length=3mm, width=2mm]}, thick, Purple] (k2) -- (f2_sigma_proj);
            \draw[-{Stealth[length=3mm, width=2mm]}, thick, ForestGreen] (k2) -- (f2);
            \draw[-{Stealth[length=3mm, width=2mm]}, thick, Purple] (k2_proj) -- ($(f2_sigma) - (k2) + (k2_proj)$);
            
            \node[draw, fill=ForestGreen!100, circle, inner sep=0pt,minimum size=4pt] (pk2) at (k2) {}; 
            \node[draw, fill=blue!100, circle, inner sep=0pt,minimum size=4pt] (pk1) at (k1) {}; 
            \node[draw, fill=ForestGreen!100, circle, inner sep=0pt,minimum size=4pt] (pk2_star) at (k2_star) {}; 
            \node[draw, fill=ForestGreen!100, circle, inner sep=0pt,minimum size=4pt] (pk2_proj) at (k2_proj) {}; 
            \node at ($(k2)             + ( 0.3,  0.0)$) {$\textcolor{ForestGreen}{\bk_2}$};
            \node at ($(k1)             + (-0.2, -0.2)$) {$\textcolor{blue}{\bk_1}$};
            \node at ($(k2_star)        + ( 0.3, -0.05)$) {$\textcolor{ForestGreen}{\bk_2^*}$};
            \node at ($(k2_proj)        + ( 0.0, -0.3)$) {$\textcolor{ForestGreen}{\bk_2'}$};
            \node at ($(f2_approach)    + ( 0.3,  0.15)$) {$\textcolor{LimeGreen}{\force_a}$};
            \node at ($(f2_sigma)       + (-0.1,  0.2)$) {$\textcolor{Purple}{\force_\sigma}$};
            \node at ($(f2_sigma) - (k2) + (k2_proj) + (0.2, -0.3)$) {$\textcolor{Purple}{\force_\sigma}$};
            \node at ($(f2_sigma_proj)  + ( 0.1,  0.35)$) {$\textcolor{Purple}{\force'_\sigma}$};
            \node at ($(f2)             + (-0.2,  0.15)$) {$\textcolor{ForestGreen}{\force_2}$};
            
            \node at (-1.4, -0.4) {\scaleeq{0.8}{\sigma(\bm{x})}};
            \node at (-1.4, -1.2) {\scaleeq{0.8}{\manifold{\text{plane}}}};
            \node at (-1.4,  0.7) {\scaleeq{0.8}{\manifold{\text{plane}}'}};
            \node [rotate=19] at (0.7, 1.25) {\scaleeq{0.8}{T_x\manifold{\text{s}}}};
            \node at (0, -1.7) {\scaleeq{0.8}{\manifold{\text{s}}}};
        \end{tikzpicture}
        \vspace{-4ex}
	\end{subfigure}
	\begin{subfigure}{0.48\columnwidth}
	    \begin{tikzpicture}
            \node (image) at (0,0) {
                \includegraphics[width=\textwidth]{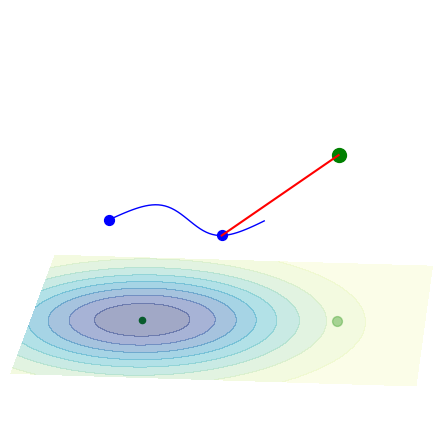}
            };
            \coordinate (k1)                 at (-0.05, -0.24);
            \coordinate (k2)                 at (1.08,  0.52);
            \coordinate (k1_star)            at (-0.25,  -0.16);
            \coordinate (f2)                 at (0.65,  -0.62);
            \coordinate (f1)                 at (-0.8,   0.08);
            \coordinate (k1_goal)            at (-1.08,  -0.05);
            \coordinate (k2_proj)            at (1.13, -0.84);
            \coordinate (k2_goal)            at (-0.1, -1.1);
            \coordinate (pm)                 at (-0.75, -1.0);
            
            \draw[-{Stealth[length=3mm, width=2mm]}, thick, ForestGreen] (k2) -- (f2);
            \draw[-{Stealth[length=3mm, width=2mm]}, thick, blue] (k1) -- (f1);
            \draw[thick, red] (k1) -- (k2);
            \draw[thick, red] (k1_goal) -- (k2_goal);
            \draw[thick, red, dashed] (k1_goal) -- (pm);
            
            \node[draw, fill=black!100, circle, inner sep=0pt,minimum size=4pt] (pk2) at ($(k2)!0.5!(k1)$) {}; \node[draw, fill=ForestGreen!100, circle, inner sep=0pt,minimum size=4pt] (pk2) at (k2) {}; 
            \node[draw, fill=ForestGreen!100, circle, inner sep=0pt,minimum size=4pt] (pk2_goal) at (k2_goal) {}; 
            \node[draw, fill=ForestGreen!100, circle, inner sep=0pt,minimum size=4pt] (p_pm) at (pm) {}; 
            \node[draw, fill=blue!100, circle, inner sep=0pt,minimum size=4pt] (pk1) at (k1) {}; 
            \node[draw, fill=blue!100, circle, inner sep=0pt,minimum size=4pt] (pk1_goal) at (k1_goal) {}; 
            \node[draw, fill=blue!100, circle, inner sep=0pt,minimum size=4pt] (pk1_star) at (k1_star) {}; 
            
            \node at ($(k2)!0.5!(k1)    + (-0.1,  0.35)$) {$\textcolor{black}{\bar\bk}$};
            \node at ($(k2)             + ( 0.3,  0.2)$) {$\textcolor{ForestGreen}{\bk_2}$};
            \node at ($(k2_goal)        + ( 0.0, -0.3)$) {$\textcolor{ForestGreen}{\bk_2^g}$};
            \node at ($(pm)             + ( 0.0, -0.3)$) {$\textcolor{ForestGreen}{\bk_2^m}$};
            \node at ($(k1)             + ( 0.0, -0.3)$) {$\textcolor{blue}{\bk_1}$};
            \node at ($(k1_star)        + ( 0.0, 0.3)$)  {$\textcolor{blue}{\bk_1^*}$};
            \node at ($(k1_goal)        + (-0.05, 0.4)$)  {$\textcolor{blue}{\bk_1^g (\bk_1^m)}$};
            \node at ($(f2)             + ( 0.1, -0.2)$) {$\textcolor{ForestGreen}{\force_2}$};
            \node at ($(f1)             + ( 0.1, -0.3)$) {$\textcolor{blue}{\force_1}$};

        \end{tikzpicture}
        \vspace{-4ex}
	\end{subfigure}
	\caption{
	Illustration of the attraction and density forces when $\textcolor{ForestGreen}{\bk_2}$ is subject to a $\ptoP$ constraint. In contrast to \cref{fig:kac_line}, the density force $\force_\sigma$ is projected onto the intersection line (\sampleline{dashed,orange,line width=0.2mm}) of the shifted principal manifold $\manifold{\text{plane}}'$ and the tangent space. Legend as in \cref{fig:kac_line}.
	}
	\label{fig:kac_plane}
        \vspace{-1.5ex}
\end{figure}

\subsection{Density force}
\label{kac:density} 
Given a non-$\ptop$ constraint,
we project the demonstrated keypoint positions \smalleq{$\tilde{\traj}_l(t)$} at the extracted time step $t$ onto the corresponding $d$-dimensional principal manifold using the learned projection index $\pi_d$, i.e., \smalleq{$\hat\bk_l^m = \pi_d(\tilde{\traj}_l(t)) \in\mathbb{R}^{N\times d}$}.
Since at time step $t$, the $l\nth$ keypoint is supposed to fulfill the geometric constraints, we interpret $\hat \bk_l^m$ as its demonstrated target positions on the manifold. 
We then estimate the density function \smalleq{$\sigma(\bx)$} of the keypoint target positions from \smalleq{$\hat \bk_l^m$} using kernel density estimation~\cite{scikit-learn} with SE kernels.
Examples of estimated density functions for $\ptol$, $\ptoP$, $\ptoc$, and $\ptoS$ constraints are depicted in \cref{subfig:pm_p2l,subfig:pm_p2P,subfig:pm_p2c,subfig:pm_p2S}.
This density function indicates the probability of a keypoint target position on the corresponding principal manifold given the demonstrated target positions. 
In other words, the density function indicates the confidence level of K-VIL when extrapolating the keypoint target positions to new locations on the principal manifold, which may occur due to object shape variations in the reproduction.  
Examples of extrapolated keypoint target positions ($\textcolor{blue}{\boldsymbol{\times}}$) during reproduction are depicted in \cref{subfig:pm_p2l,subfig:pm_p2P,subfig:pm_p2c,subfig:pm_p2S}.
Notice that the target position in \cref{subfig:pm_p2l}-\emph{left} has a lower probability (i.e., a lower extrapolation confidence) than the one in \cref{subfig:pm_p2l}-\emph{right} due to its increased distance with the demonstrated target positions.
This illustrates that the control of such keypoints must not only fulfill the geometric constraints, but also be as close as possible to the demonstrated targets on the constraints.
Therefore, in addition to the attraction force \smalleq{$\force_{a,l}$} that guarantees the fulfillment of the geometric constraint, we define a \emph{density force} \smalleq{$\force_{\sigma, l}$} to drive the keypoints into regions with higher probability. 
To do so, we first define the driving force \smalleq{$\force_{\sigma,1}$} computed from the density field \smalleq{$\nabla \sigma(\bx)$} as
\begin{equation}
    \force_{\sigma,1} = g_1 \cdot f(\nabla \sigma(x)). \label{eq:density_f1}
\end{equation}
For the regions where \smalleq{$\force_{\sigma,1}$} is too small to drive the keypoints, we then define a minimal driving force \smalleq{$\force_{\sigma,2}$} as 
\begin{equation}
    \textstyle \force_{\sigma,2} = g_2 \cdot f(\frac{\bk^m - \bk'_l}{\Vert \bk^m - \bk'_l \Vert_2}), \label{eq:density_f2}
\end{equation}
where \smalleq{$\bk'_l=\pi_d(\bk_l) \in\mathbb{R}^d$} is the projection of the keypoint onto the principal manifold, 
\smalleq{$f(\cdot)$} is the reconstruction function (see \cref{sec:stage_2}), $g_1$ and $g_2$ are the force scaling parameters. 
Note that \smalleq{$\force_{\sigma,2}$} points directly to the mean of the demonstrated targets \smalleq{$\hat{\bk}_l^m$} on the principal manifold, i.e., \smalleq{$\bk^m = \avg(\hat{\bk}_l^m)$}.
Then the density force is the maximum of \smalleq{$\force_{\sigma,1}$} and \smalleq{$\force_{\sigma,2}$}, i.e.,
\begin{equation}
    \label{eq:density_force}
    \textstyle \force_{\sigma, l} = \argmax_{\force}\Vert \force\Vert_2, \quad \force \in \{\force_{\sigma,1}, \force_{\sigma,2}\}
\end{equation}
Examples of such density forces \smalleq{$\force_\sigma$} in the case of $\ptol$ and $\ptoP$ constraints are shown in \cref{fig:kac_line,fig:kac_plane}.

\subsection{Priority}\label{kac:priority} 
In the case of large object shape variations, controlling a $\ptol$ constraint with the same priority as a $\ptop$ constraint may lead to a violation of the latter.
To reduce such interference, we propose to set a higher priority to $\mathsf{p2p}$ constraints compared to the other constraint types.
For the sake of clarity, we use \cref{fig:kac_line,fig:kac_plane} to explain this concept, where \cref{fig:kac_line} shows the case of two constraints, $\ptop$ for $\bk_1$ and $\ptol$ for $\bk_2$, while $\bk_2$ is subject to a $\ptoP$ constraint in \cref{fig:kac_plane}. 
In both cases, we construct a sphere centered at $\bk_1$ with radius \smalleq{$\Vert \bk_2 - \bk_1\Vert$} and define the tangent space of the sphere at $\bk_2$ as the plane formed by all the lines tangent to the sphere at $\bk_2$.
For clarity, \cref{fig:kac_line,fig:kac_plane} also depicts the corresponding principal manifolds \smalleq{$\manifold{\text{line}}, \manifold{\text{plane}}$} shifted in parallel to go through $\bk_2$ as \smalleq{$\manifold{\text{line}}', \manifold{\text{plane}}'$}. 
Assuming solid connections (\sampleline{red}) between $\bk_1$ and $\bk_2$, large density forces \smalleq{$\force_\sigma$} generated for $\bk_2$ will also drag $\bk_1$ along the same direction.
This may lead to the violation of the $\ptop$ constraint of $\bk_1$, and cause collision if $\bk_1$ it is close to the \master{} object. 
To reduce such interference
when the principal manifold is a line \smalleq{$\manifold{\text{line}}$}, we project \smalleq{$\force_{\sigma}$} onto the tangent space \smalleq{$T_x\manifold{\text{s}}$} of the sphere \smalleq{$\manifold{\text{s}}$}, so that the motion of $\bk_1$ remains unaffected by the projected density force \smalleq{$\force'_\sigma$} (see \cref{fig:kac_line}).
Similarly, when a principal manifold is a plane (\cref{fig:kac_plane}), we project \smalleq{$\force_{\sigma}$} onto the intersection between the tangent space and the shifted principal plane \smalleq{$\mathcal M'_{\text{plane}}$}. 
This also holds for the nonlinear constraints ($\mathsf{p2c}$ and $\mathsf{p2S}$), for which a linear approximation is considered at each time step.

In summary, on the one hand, the density force allows the reproduced task to be similar to the demonstrations on the principal manifold.
On the other hand, the priority mechanism reduces the interference of the density force with $\ptop$ constraints, while maintaining the extrapolation capability of K-VIL.
Overall, the decomposition of the control force into attraction force (\cref{kac:attract}) and projected density force (\cref{kac:density,kac:priority}) is key to balancing the similarity of the reproduced task to the demonstration and the extrapolation capability.
In practice, the stiffness and damping gains of the virtual spring-damper systems are empirically tuned for good tracking accuracy and control stability.
Notice that one-shot IL is considered as a special case (see \cref{sec:criteria_dist}). This is due to the fact that the learned task representation is composed of $3$ keypoints subject to $\mathsf{p2p}$ constraints. In this case, no density force is needed, and the constraint priorities are defined as \scaleeq{0.9}{\Pri_1 > \Pri_2 > \Pri_3}, since $\bk_1$ usually represents the contact point of two objects.
Therefore, to ensure a higher control precision of $\bk_1$, we set the stiffness gains of the three keypoints to 
\smalleq{$\bbK_{p,1}$}\smalleq{$=5\bbK_{p,2}$}\smalleq{$=10\bbK_{p,3}$} and the respective damping gains to \smalleq{$\bbK_{d,l} = 2\bbK_{p,l}^{1/2}$}, where \smalleq{$l\in[1, 3]$}, which ensure a critically damped behavior for control stability.

\subsection{Admittance control}\label{kac:adm} 
The goal of the KAC is to compute the control command of the robot arm from the attraction forces \smalleq{$\force_a$} and the projected density forces \smalleq{$\force'_\sigma$} of all keypoints. 
To do so, we first compute the control force of each keypoint as 
\scaleeq{0.9}{\force_l = \force_{a, l} +\force'_{\sigma,l}}
and define a virtual tool-center-point (TCP) \scaleeq{0.9}{\bar \bk = \sum_{l=1}^L \bk_l/L} as the mean of all keypoint positions (see \cref{fig:kac_line}-\emph{right}, \cref{fig:kac_plane}-\emph{right}).
The virtual TCP is driven by a virtual force and torque
\begin{equation*}
    \textstyle \force_{f} = \sum_{l=1}^L \force_l \;\; \text{ and }\;\; \force_{\tau} = \sum_{l=1}^L (\bk_l - \bar{\bk})\times \force_l
\end{equation*}
with $\times$ denoting the vector cross product.
The total control force \scaleeq{0.85}{\force_v = [\force_{f}^\trsp, \force_{\tau}^\trsp]^\trsp} is applied to the robot end-effector (i.e., the humanoid hand) to calculate the virtual acceleration as
\begin{equation*}
    \ddot{\bx}_v = \tbK_p (\bx_0 - \bx_v) - \tbK_d \dot{\bx}_v - \tbK_m \force_v,
\end{equation*} 
where \smalleq{$\bx_0, \bx_v$} are the initial and virtual poses of the robot end-effector, $\dot{\bx}_v$ is its virtual velocity, and
\scaleeq{0.9}{\tbK_m, \tbK_d}, and \scaleeq{0.9}{\tbK_p} are the inertia, damping and stiffness factors, respectively. 
The robot is controlled using a task space inverse dynamics controller, whose task space control force \smalleq{$\force_m$} is calculated as 
\begin{equation*}
    \force_m =  \bK_p(\bx_v - \bx) + \bK_d (\dot{\bx}_v - \dot{\bx}) + \bh_c,
\end{equation*}
where \smalleq{$\bx$, $\dot{\bx}$} are the current end-effector pose and velocity, \scaleeq{0.9}{\bK_d}, and \scaleeq{0.9}{\bK_p} are the damping and stiffness factors of the impedance controller, respectively, and \smalleq{$\bh_c$} represents the Coriolis and gravitational force in the task space.

\begin{figure}[t]
    \centering
    \setlength\tabcolsep{0.5pt}
    \tabulinesep=0pt
    \begin{tabu}{M{18mm}M{18mm}M{29mm}M{18mm}}
        \multicolumn{4}{c}{
            \begin{subfigure}{0.98\columnwidth}
                \centering
                \begin{tikzpicture}
                    \hspace{-22pt}
                    \node[inner sep=0mm] (image) at (0, 0) {
                        \includegraphics[width=\textwidth]{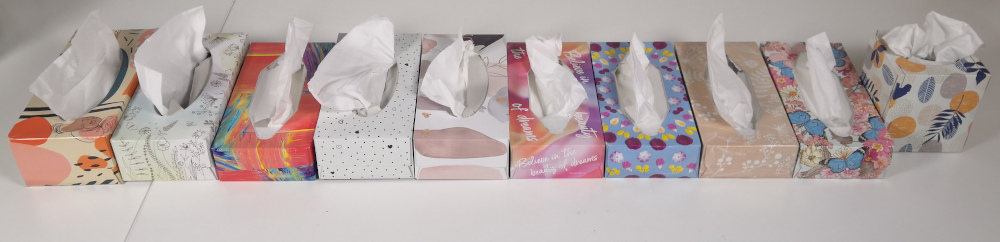}
                    };
                    \drawlabel{(a)}{(4.0, 0.8)}{fill opacity=0.5}
                    \foreach \x in {1,...,10} { \node [below] at (0.83*\x-4.6, -0.55) {\scalebox{0.8}{\x}}; }
                \end{tikzpicture}
                \phantomsubcaption
                \label{subfig:tisse}
                \vspace{-0.9em}
            \end{subfigure}
        } \\
        \multicolumn{3}{c}{
            \begin{subfigure}{65mm}
                \centering
                \begin{tikzpicture}
                    \hspace{-1pt}
                    \node[inner sep=0mm] (image) at (0, 0) {
                        \includegraphics[width=65mm]{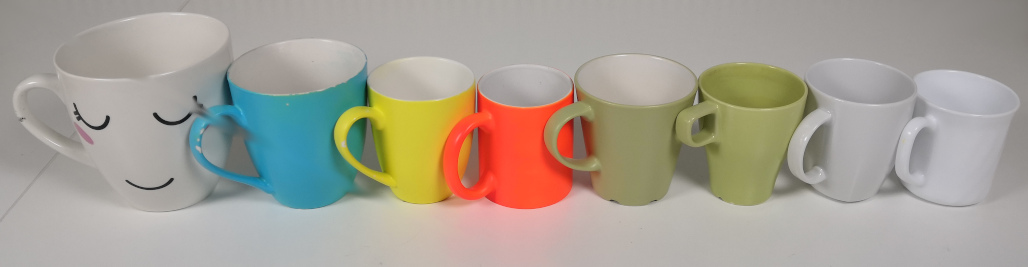}
                    };
                    \drawlabel{(b)}{(2.94, 0.6)}{fill opacity=0.5}
                    \foreach \x in {1,...,8} { \node [below] at (0.72*\x-2.9, -0.45) {\scalebox{0.8}{\x}}; }
                \end{tikzpicture}
                \phantomsubcaption
                \label{subfig:cups}
                \vspace{-1.1em}
            \end{subfigure}
        }
        & 
        \multirow{2}{*}{
            \begin{subfigure}{0.3\columnwidth}
                \centering
                \begin{tikzpicture}
                    \hspace{-8pt}
                    \node[inner sep=0mm] (image) at (0, -0.08) {
                        \includegraphics[height=4.1cm,trim=18 15 5 15, clip]{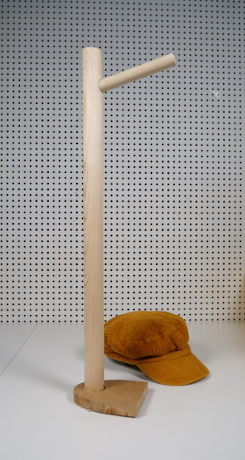}
                    };
                    \drawlabel{(c)}{(0.7, 1.8)}{fill opacity=0.5}
                \end{tikzpicture}
                \phantomsubcaption
                \label{subfig:rack_hat}
            \end{subfigure}
        }
        \\
        \multicolumn{3}{c}{
            \begin{subfigure}{65mm}
                \centering
                \begin{tikzpicture}
                    \hspace{-1pt}
                    \node[inner sep=0mm] (image) at (0, 0) {
                        \includegraphics[width=65mm]{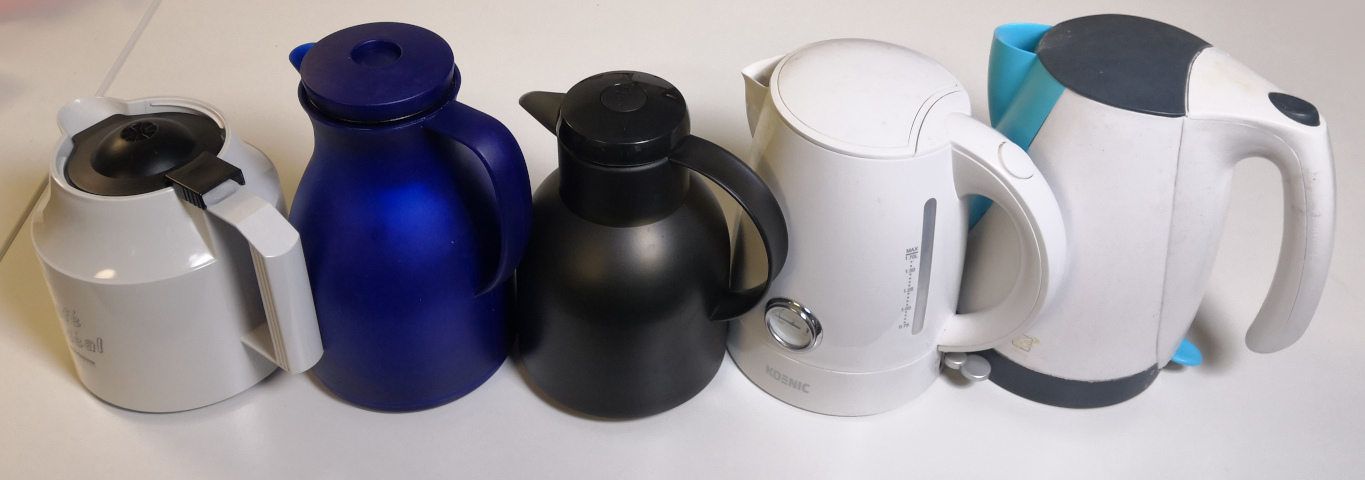}
                    };
                    \drawlabel{(d)}{(2.94, 0.9)}{fill opacity=0.5}
                    \foreach \x in {1,...,5} { \node [below] at (1.1*\x-3.6, -0.76) {\scalebox{0.8}{\x}}; }
                \end{tikzpicture}
                \phantomsubcaption
                \label{subfig:kettles}
                \vspace{-1.1em}
            \end{subfigure}
        }
        \\
        \begin{subfigure}{18mm}
            \centering
            \begin{tikzpicture}
                \hspace{-2pt}
                \node[inner sep=0mm] (image) at (0, 0) {
                    \includegraphics[height=2.5cm, trim=15 5 15 5, clip]{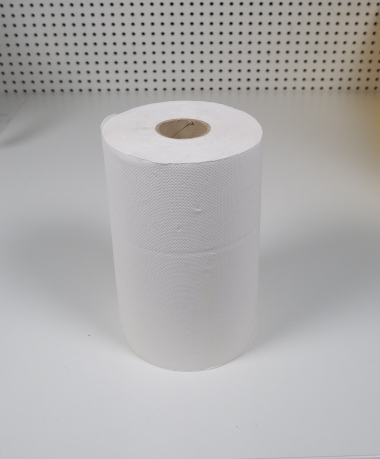}
                };
                \drawlabel{(e)}{(0.56, 1.0)}{fill opacity=0.5}
            \end{tikzpicture}
            \phantomsubcaption
            \label{subfig:paper_roll}
        \end{subfigure}
        & 
        \begin{subfigure}{29mm}
            \centering
            \begin{tikzpicture}
                \hspace{2pt}
                \node[inner sep=0mm] (image) at (0, 0) {
                    \includegraphics[height=2.5cm, trim=15 4 15 10, clip]{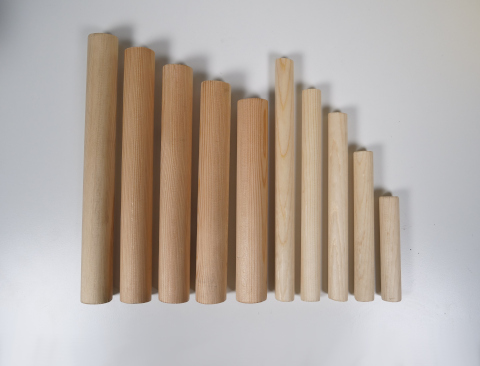}
                };
                \drawlabel{(f)}{(1.25, 1.0)}{fill opacity=0.5}
                \foreach \x in {1,...,5} { \node [below] at (0.28*\x-1.4, -0.8) {\scalebox{0.8}{\x}}; }
                \foreach \x in {6,...,10} { \node [below] at (0.21*\x-1.0, -0.8) {\scalebox{0.8}{\x}}; }
            \end{tikzpicture}
            \phantomsubcaption
            \label{subfig:sticks}
        \end{subfigure}
        & 
        \multicolumn{2}{c}{
            \begin{subfigure}{65mm}
                \vspace{-1.2em}
                \centering
                \begin{tikzpicture}
                    \hspace{2pt}
                    \node[inner sep=0mm] (image) at (0, 0) {
                        \includegraphics[height=2.5cm]{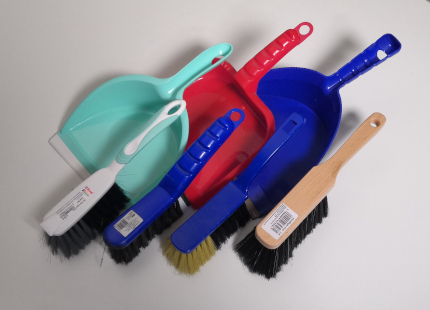}
                    };
                    \drawlabel{(g)}{(1.5, 1.0)}{fill opacity=0.5}
                    \foreach \x in {1,...,3} { \node [below] at (0.8*\x-1.5, 1.2) {\scalebox{0.8}{\x}}; }
                    \foreach \x in {1,...,4} { \node [below] at (0.6*\x-1.8, -0.8) {\scalebox{0.8}{\x}}; }
                \end{tikzpicture}
                \phantomsubcaption
                \label{subfig:dustpan_brush}
            \end{subfigure}
        }
    \end{tabu}
    \vspace{-2.5ex}
    \caption{Objects used in our paper include \subref{subfig:tisse} tissue boxes, \subref{subfig:cups} teacups,  \subref{subfig:rack_hat} a rack and a hat,  \subref{subfig:kettles} kettles, \subref{subfig:paper_roll} a paper roll, \subref{subfig:sticks} sticks, \subref{subfig:dustpan_brush} dustpans and brushes. Note that the rack can be assembled with stick \#6-10 to have multiple shape variations.
    }
    \label{fig:objects}
    \vspace{-2ex}
\end{figure}

\section{Evaluation}
\label{sec:eval}
\begin{figure}[tb]
	\setlength{\subfight}{1.2cm}
	\sbox\subfigbox{%
		\resizebox{0.95\columnwidth}{!}{
			\includegraphics[height=\subfight]{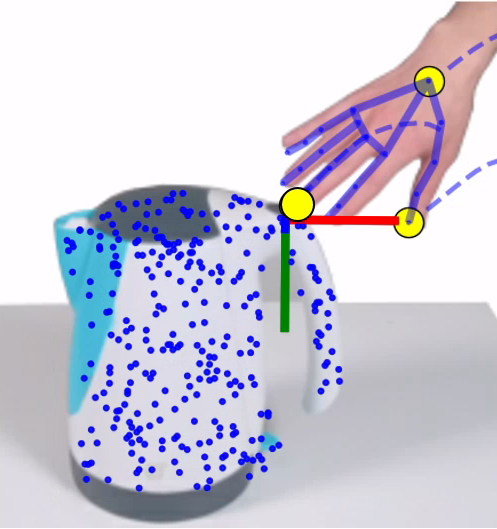}
			\hspace{1pt}
			\includegraphics[height=\subfight]{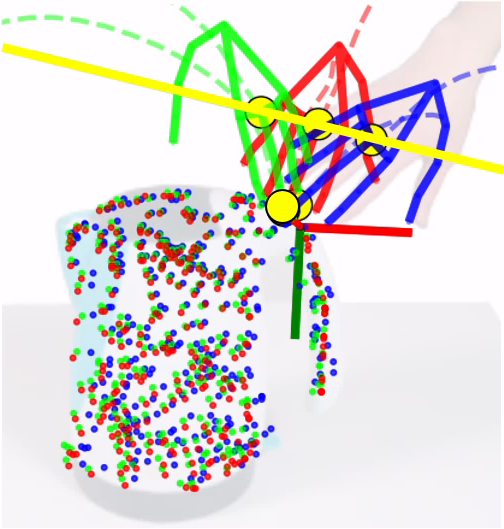}
			\hspace{1pt}
			\includegraphics[height=\subfight]{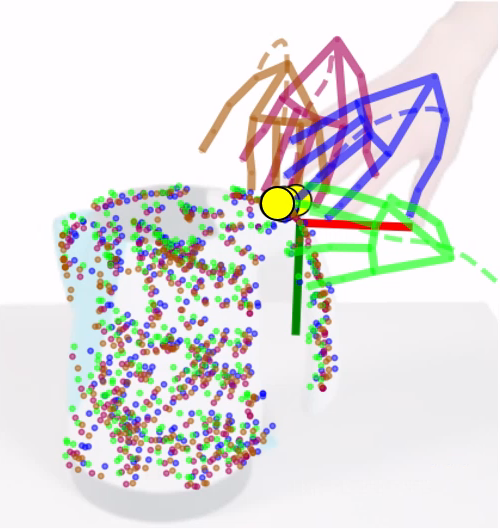}
		}
	}
	\setlength{\subfight}{\ht\subfigbox}
	\centering
        \vspace{-0.4em}
	\subcaptionbox{$\mathsf{p2p}$: $\bk_1, \bk_2, \bk_3$\label{subfig:press_1}}{
	    \begin{tikzpicture}
                \hspace{-2.5pt}
                \node (image) at (0, 0) {
                    \includegraphics[height=\subfight]{figure/real/press_kettle_button/cfg/kvil_demo_1.png}
                };
                \node at ( 0.50, -0.1) {\scalebox{0.9}{$\lf$}};
                \node at (-0.10,  0.5) {\scalebox{0.9}{$\bk_1$}};
                \node at ( 0.50,  1.0) {\scalebox{0.9}{$\bk_2$}};
                \node at ( 1.00,  0.0) {\scalebox{0.9}{$\bk_3$}};
                \node at (-1.00,  1.0) {\circled{1}};
            \end{tikzpicture}
            \vspace{-0.5em}
	}
	\hspace{-0.5em}
	\subcaptionbox{$\mathsf{p2p}$: $\bk_1, \mathsf{p2l}$: $\bk_2$\label{subfig:press_3}}{
	    \begin{tikzpicture}
                \hspace{-2.5pt}
                \node (image) at (0, 0) {
                    \includegraphics[height=\subfight]{figure/real/press_kettle_button/cfg/kvil_demo_3.png}
                };
                \node at ( 0.5, -0.1) {\scalebox{0.9}{$\lf$}};
                \node at (-0.2,  0.5) {\scalebox{0.9}{$\bk_1$}};
                \node at ( 0.1,  1.0) {\scalebox{0.9}{$\bk_2$}};
                \node at (-0.8,  0.6) {\scalebox{0.9}{$\manifold{\text{line}}$}};
                \node at (-1.0,  1.0) {\circled{3}};
            \end{tikzpicture}
            \vspace{-0.5em}
	}
	\hspace{-0.5em}
	\subcaptionbox{$\mathsf{p2p}$: $\bk_1$\label{subfig:press_4}}{
	    \begin{tikzpicture}
                \hspace{-2.5pt}
                \node (image) at (0, 0) {
                    \includegraphics[height=\subfight]{figure/real/press_kettle_button/cfg/kvil_demo_4.png}
                };
                \node at ( 0.5, -0.1) {\scalebox{0.9}{$\lf$}};
                \node at (-0.2,  0.5) {\scalebox{0.9}{$\bk_1$}};
                \node at (-1.0,  1.0) {\circled{4}};
            \end{tikzpicture}
            \vspace{-0.5em}
	}
	\caption{
            Press a button on the kettle to open the lid with variations in hand orientation. 
            The candidate points (colored points) and the skeleton of hands (colored line segments) are overlain on the objects at time step $T$. Different colors denote different trials. 
            The keypoints $\bk_l$ (\inlinekpt{black}{fill=yellow!80, line width=0.4pt}) are extracted from different number \circled{N} of demonstrations in each subfigure.
            The demonstrated trajectories (\sampleline{dashed,thick}), the local frames $\lf$, and the estimated principal manifolds (\raisebox{-1pt}{\sampleline{yellow(process),line width=2pt}}) are also depicted.
            Notice that the demonstrations were provided from different viewpoints, although they are here represented aligned to the local frame $\lf$ for a better illustration of the extracted keypoints and constraints (for viewpoint mismatch, see \cref{fig:clean_table_task}.)
        }
	\label{fig:press_button}
	\vspace{-1.5ex}
\end{figure}

\begin{figure}[t]
	\setlength{\subfight}{1.2cm}
	\sbox\subfigbox{%
		\resizebox{0.95\columnwidth}{!}{
			\includegraphics[height=\subfight]{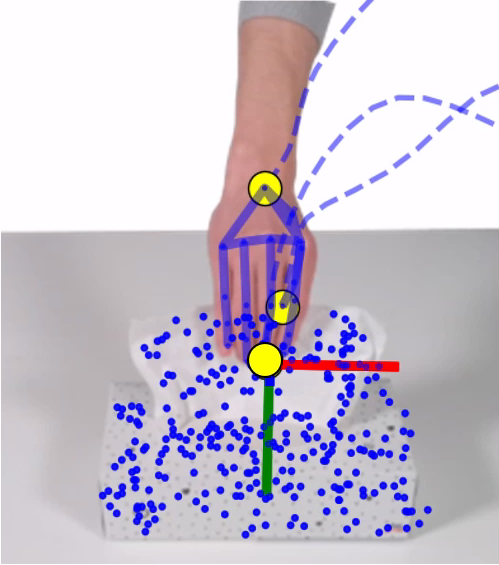}
			\includegraphics[height=\subfight]{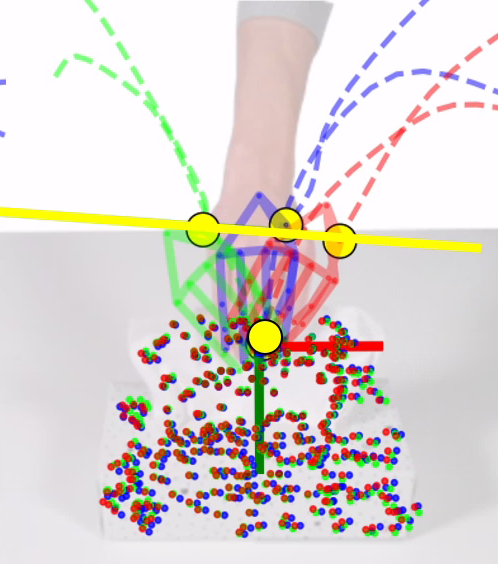}
			\includegraphics[height=\subfight]{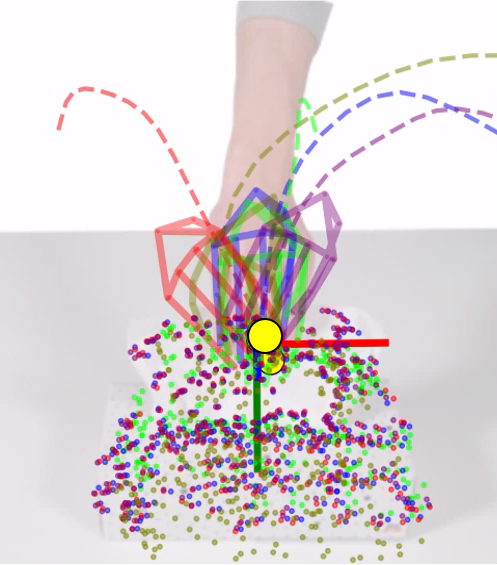}
		}
	}
	\setlength{\subfight}{\ht\subfigbox}
	\centering
	\subcaptionbox{$\mathsf{p2p}$: $\bk_1, \bk_2, \bk_3$\label{subfig:tissue_1}}{
	    \begin{tikzpicture}
                \hspace{-2.5pt}
                \node (image) at (0, 0) {
                    \includegraphics[height=\subfight]{figure/real/fetch_tissue/cfg/kvil_demo_1.png}
                };
                \node at ( 0.90, -0.45) {\scalebox{0.9}{$\lf$}};
                \node at (-0.70, -0.10) {\scalebox{0.9}{$\bk_1$}};
                \draw[-{Stealth[length=3mm, width=2mm]}, line width=0.5mm, white] (-0.55, -0.1) -- (0, -0.35);
                \node at (-0.35,  0.60) {\scalebox{0.9}{$\bk_2$}};
                \node at ( 0.55, -0.05) {\scalebox{0.9}{$\bk_3$}};
                \node at (-1.00,  1.20) {\circled{1}};
            \end{tikzpicture}
            \vspace{-0.5em}
	}
	\hspace{-0.5em}
	\subcaptionbox{$\mathsf{p2p}$: $\bk_1, \mathsf{p2l}$: $\bk_2$\label{subfig:tissue_3}}{
	    \begin{tikzpicture}
                \hspace{-2.5pt}
                \node (image) at (0, 0) {
                    \includegraphics[height=\subfight]{figure/real/fetch_tissue/cfg/kvil_demo_3.png}
                };
                \node at ( 0.9, -0.35) {\scalebox{0.9}{$\lf$}};
                \node at (-0.7, -0.10) {\scalebox{0.9}{$\bk_1$}};
                \draw[-{Stealth[length=3mm, width=2mm]}, line width=0.5mm, white] (-0.55, -0.1) -- (0, -0.26);
                \node at ( 0.8,  0.40) {\scalebox{0.9}{$\bk_2$}};
                \node at (-0.8,  0.50) {\scalebox{0.9}{$\manifold{\text{line}}$}};
                \node at (-1.0,  1.20) {\circled{3}};
            \end{tikzpicture}
            \vspace{-0.5em}
	}
	\hspace{-0.5em}
	\subcaptionbox{$\mathsf{p2p}$: $\bk_1$\label{subfig:tissue_4}}{
	    \begin{tikzpicture}
                \hspace{-2.5pt}
                \node (image) at (0, 0) {
                    \includegraphics[height=\subfight]{figure/real/fetch_tissue/cfg/kvil_demo_4.png}
                };
                \node at ( 0.9, -0.35) {\scalebox{0.9}{$\lf$}};
                \node at (-0.7, -0.10) {\scalebox{0.9}{$\bk_1$}};
                \draw[-{Stealth[length=3mm, width=2mm]}, line width=0.5mm, white] (-0.55, -0.1) -- (0, -0.26);
                \node at (-1.0,  1.20) {\circled{4}};
            \end{tikzpicture}
            \vspace{-0.5em}
	}
	\caption{Fetch tissue with variations in hand orientation. Legend as in \cref{fig:press_button}.}
	\label{fig:fetch_tissue}
	\vspace{-1.8ex}
\end{figure}

\begin{figure}[t]
	\setlength{\subfight}{1.7cm}
	\sbox\subfigbox{%
		\resizebox{\columnwidth}{!}{
			\includegraphics[height=\subfight]{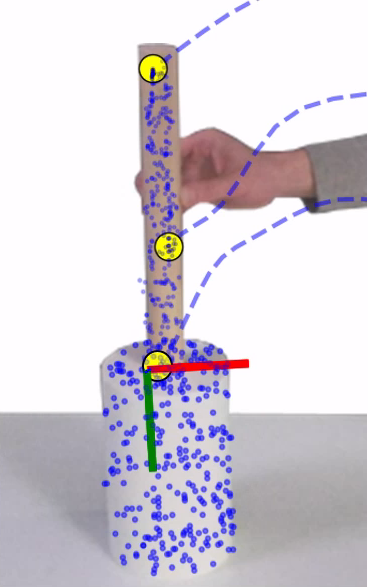}
			\includegraphics[height=\subfight]{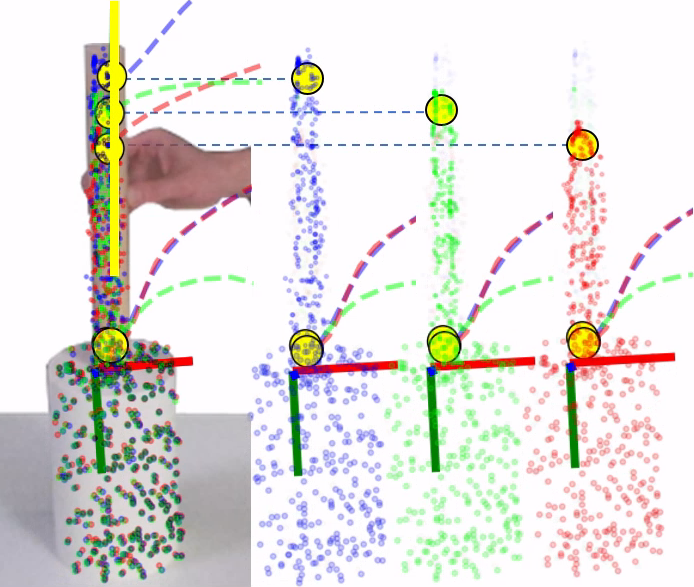}
			\includegraphics[height=\subfight]{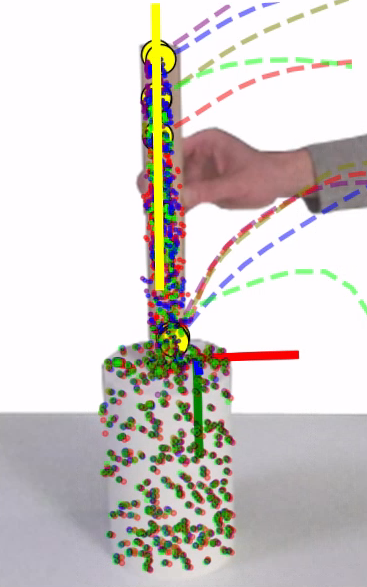}
		}
	}
	\setlength{\subfight}{\ht\subfigbox}
	\centering
        \vspace{-0.3em}
	\subcaptionbox{\scaleeq{0.9}{\mathsf{p2p}$: $\bk_1, \bk_2, \bk_3}\label{subfig:insert_1}}{
	    \begin{tikzpicture}
                \hspace{-2.5pt}
                \node (image) at (0, 0) {
                    \includegraphics[height=\subfight]{figure/real/insertion/cfg/kvil_demo_1.png}
                };
                \node at (0.5, -0.40) {$\lf$};
                \node at (-0.45, -0.15) {$\bk_1$};
                \node at (-0.45, 1.1) {$\bk_2$};
                \node at (-0.45, 0.30) {$\bk_3$};
                \node at (-0.8, 1.5) {\circled{1}};
            \end{tikzpicture}
            \vspace{-0.8em}
	}
    \hspace{-1.5em}
    \subcaptionbox{$\mathsf{p2p}$: $\bk_1, \mathsf{p2l}$: $\bk_2$\label{subfig:insert_3}}{
	    \begin{tikzpicture}
                \hspace{-2.5pt}
                \node (image) at (0, 0) {
                    \includegraphics[height=\subfight]{figure/real/insertion/cfg/kvil_demo_3.png}
                };
                \node at (-0.7, -0.40) {$\lf$};
                \node at (-1.65, -0.15) {$\bk_1$};
                \node at (-1.65, 1.0) {$\bk_2$};
                \node at (-1.8, 1.5) {\circled{3}};
                \node at (-0.8, 1.5) {\scalebox{0.9}{$\manifold{\text{line}}$}};
            \end{tikzpicture}
            \vspace{-0.8em}
	}
	\hspace{-1.5em}
	\subcaptionbox{$\mathsf{p2p}$: $\bk_1, \mathsf{p2l}$: $\bk_2$\label{subfig:insert_5}}{
	    \begin{tikzpicture}
                \hspace{-2.5pt}
                \node (image) at (0, 0) {
                    \includegraphics[height=\subfight]{figure/real/insertion/cfg/kvil_demo_5.png}
                };
                \node at (0.7, -0.40) {$\lf$};
                \node at (-0.45, -0.15) {$\bk_1$};
                \node at (-0.45, 1.0) {$\bk_2$};
                \node at (-0.8, 1.5) {\circled{5}};
                \node at (0.35, 1.5) {\scalebox{0.9}{$\manifold{\text{line}}$}};
            \end{tikzpicture}
            \vspace{-0.8em}
	}

	\caption{Approach the insertion position to insert sticks with 3 length variations into a paper roll. Legend as \cref{fig:press_button}.
	}
	\label{fig:insertion}
        \vspace{-1.5ex}
\end{figure}

We evaluate our approach in five daily tasks involving different types of geometric constraints and various categorical objects (see \cref{fig:objects}). Namely, the considered tasks are
press button (\taskabbr{PB}, \cref{fig:press_button}), 
fetch tissue (\taskabbr{FT}, \cref{fig:fetch_tissue}), insert sticks into a paper roll (\taskabbr{IS}, \cref{fig:insertion}), 
pour water (\taskabbr{PW}, \cref{fig:pouring}), 
hang hat on a rack (\taskabbr{HH}, \cref{fig:hanging}), clean table with a dustpan and a brush (\taskabbr{CT}, \cref{fig:clean_table_task}). 
As a prerequisite for our experiments, we train DON in a self-supervised and task-agnostic manner. 
The training dataset was collected with a handheld Azure Kinect camera moving around objects such as tissue boxes, teacups, a rack, a hat, kettles, a paper roll, sticks, dustpans and brushes (see \cref{fig:objects}). 
The collected data was then post-processed via a 3D reconstruction process using Open3D~\cite{Zhou2018}. 
We used MediaPipe to detect $21$ keypoints on the human hands and treat the hands as a special type of object. 
It is important to emphasize that K-VIL is not limited to DON and MediaPipe, but instead can be used with any correspondence detection model, e.g., NDFs.
In all experiments, we sample \smalleq{$P_i = 300$} (for hands \scaleeq{0.9}{P_i=21}) candidate points on each \slave{} object and use \smalleq{$Q=50$} (for hands \scaleeq{0.9}{Q=10}) neighboring candidates on the \master{} object as references for local frame detection. 
We use 20 kernels for the VMPs. 
The empirical thresholds $\xi_1, \xi_2$, and the controller gains are fine-tuned for each task.
A full list of control parameters is included in the example code.


We first evaluate the ability of K-VIL to extract generalizable task representations given different number of demonstrations (see \cref{sec:one_shot} and \cref{sec:incremental} for a one-shot and a few-shot visual imitation learning setup, respectively, as well as \cref{fig:press_button,fig:fetch_tissue,fig:pouring,fig:hanging,fig:insertion,fig:clean_table_task}). We demonstrate how variations in objects' pose and shape contribute to the efficient extraction of generalizable task representations and evaluate the ability of K-VIL to reproduce the corresponding tasks learned from a different number of demonstrations.
We discuss the problems that arise when only scarce demonstrations are provided in \cref{sec:eval_task_repr_summary}.
We then show how they are resolved by providing more demonstrations and summarize the number of demonstrations required to learn a generalizable representation for each task.
The proposed KAC is finally evaluated in terms of the control accuracy, precision, and success rate in \cref{sec:eval_kac}. For more visualizations of the evaluation results 
on one/few-shot imitation learning, reproduction of skills learned from a different number of demonstrations, and other types of geometric constraints, we refer the interested reader to the accompanying videos and to the \href{https://sites.google.com/view/k-vil}{paper website}.

\subsection{Evaluation Protocols}
\label{sec:eval_protocal}

For each task, we record a few demonstration videos (RGB-D) of a human performing the task using an Azure Kinect mounted on the head of the humanoid robot \armarVI~\cite{Asfour2019}.  
For tasks involving categorical objects, we distinguish between the object instances used for training of the vision models (the DON and Mask R-CNN models), for the demonstrations, and for the reproductions. 
If we only have one instance of a specific object category, this instance is used for the training, demonstrations, and reproductions.
We define a set of \emph{extraction tasks} $\mathcal{T}_E = \set{\taskabbr{PB}, \taskabbr{FT}, \taskabbr{PW}, \taskabbr{HH}, \taskabbr{IS}, \taskabbr{CT}}$ for which we evaluate K-VIL's ability to extract generalizable task representations given $N \in \set{1, 3, 4, 5, 11}$ demonstrations, respectively.
For clarity, we only evaluate the representations of the last time cluster, i.e., the goal configuration of each task when $t=T$. 
We then define a set of \emph{reproduction tasks} $\mathcal{T}_{R} = \set{\task{Task}{N}: \taskabbr{Task} \in \mathcal{T}_E}$, each of which is the reproduction of the \taskabbr{Task} by \armarVI with the task representation extracted from $N$ demonstrations.
In order to evaluate the reproduction and adaptation of the learned task representations, e.g., the geometric constraints, in new cluttered scenes, the scene is perturbed arbitrarily before each trial of execution. Specifically, the involved objects and the robot hands are placed in arbitrarily different locations within the workspace and the view of the camera.
The first image frame captured by the robot is used to parameterize the task with K-VIL's representation.
This includes optimizing the local frames, identifying the keypoints, configuring the geometric constraints, and generating keypoint motion trajectories using the learned VMPs.
We consider a task learned from $N$ demonstrations and from a third-person view to be \emph{generalizable} when it can be successfully reproduced by the robot with categorical objects in new cluttered scenes.
Next, we describe the specifications of each considered task in terms of the collection of demonstrations and successful reproductions by the robot. Table~\ref{table:kac_eval} provides the list of considered tasks.

\textbf{Press Button} (\taskabbr{PB}):
A human demonstrates how to open the lid of a kettle by pressing the corresponding button with the tip of the middle finger of either the left or the right hand. 
The kettle \#5 of \cref{subfig:kettles} is considered in this task.
Both hands look similar and the demonstrator approaches the button with different hand poses (e.g., see \cref{fig:press_button}).
To reproduce the demonstrated human motion by the robot, we design fixed maps between keypoints of human hands and keypoints on the robot hand. 
The reproduction of the \taskabbr{PB} task is considered successful if the robot reaches the button with its fingertip within \SI{5}{\milli\meter} to the target (the button) and if the lid is opened by closing the finger with a small angle.

\begin{figure}[t]
	\setlength{\subfight}{1.1cm}
	\sbox\subfigbox{%
		\resizebox{0.95\columnwidth}{!}{
			\includegraphics[height=\subfight]{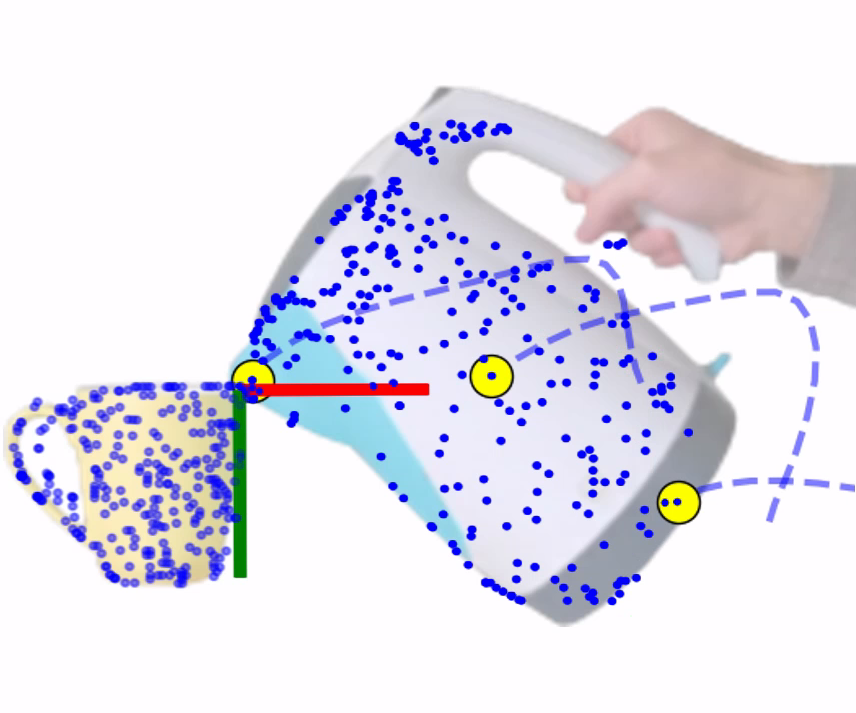}
			\includegraphics[height=\subfight]{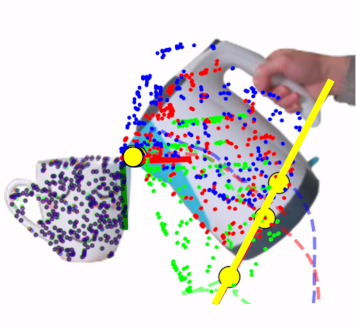}
		}
	}
	\setlength{\subfight}{\ht\subfigbox}
	\centering
        \vspace{-1em}
	\subcaptionbox{$\mathsf{p2p}$: $\bk_1, \bk_2, \bk_3$\label{subfig:pour_1}}{
	    \begin{tikzpicture}
                \hspace{-2.5pt}
                \node (image) at (0, 0) {
                    \includegraphics[height=\subfight]{figure/real/pouring/cfg/kvil_demo_1.png}};
                \node at (-0.6, -0.45) {\scalebox{0.9}{$\lf$}};
                \node at (-1.1,  0.10) {\scalebox{0.9}{$\bk_1$}};
                \node at ( 1.4, -0.90) {\scalebox{0.9}{$\bk_2$}};
                \node at ( 0.3, -0.40) {\scalebox{0.9}{$\bk_3$}};
                \node at (-1.7,  0.95) {\circled{1}};
            \end{tikzpicture}
            \vspace{-1.2em}
	}
	\hfill
	\subcaptionbox{$\mathsf{p2p}$: $\bk_1, \mathsf{p2l}$: $\bk_2$\label{subfig:pour_3}}{
	    \begin{tikzpicture}
                \hspace{-2.5pt}
                \node (image) at (0, 0) {
                    \includegraphics[height=\subfight]{figure/real/pouring/cfg/kvil_demo_3.png}};
                \node at (-0.3, -0.45) {\scalebox{0.9}{$\lf$}};
                \node at (-0.8,  0.20) {\scalebox{0.9}{$\bk_1$}};
                \node at ( 1.3, -0.80) {\scalebox{0.9}{$\bk_2$}};
                \node at ( 0.1, -0.80) {\scalebox{0.9}{$\manifold{\text{line}}$}};
                \node at (-1.9,  0.95) {\circled{3}};
            \end{tikzpicture}
            \vspace{-1.2em}
	}

    \subcaptionbox{$\mathsf{p2p}$: $\bk_1, \mathsf{p2P}$: $\bk_2$\label{subfig:pour_4}}{
	    \begin{tikzpicture}
                \hspace{-2.5pt}
                \node (image) at (0, 0) {
                    \includegraphics[height=\subfight]{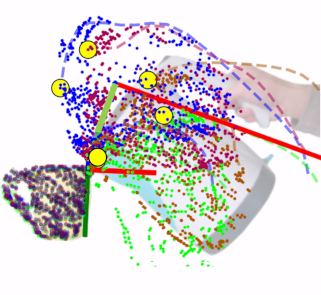}};
                \node at (-0.60, -0.45) {\scalebox{0.9}{$\lf$}};
                \node at (-1.10,  0.00) {\scalebox{0.9}{$\bk_1$}};
                \node at (-1.40,  0.90) {\scalebox{0.9}{$\bk_2$}};
                \node at (-1.70,  1.30) {\circled{4}};
                \node at ( 1.20, -0.20) {\scalebox{0.9}{$\manifold{\text{plane}, x}$}};
                \node at (-1.25,  0.45) {\scalebox{0.9}{$\manifold{\text{plane}, y}$}};
            \end{tikzpicture}
            \vspace{-1em}
	}
	\hfill
	\subcaptionbox{$\mathsf{p2p}$: $\bk_1, \mathsf{p2c}$: $\bk_2$\label{subfig:pour_11}}{
	    \begin{tikzpicture}
                \hspace{-2.5pt}
                \node (image) at (0, 0) {
                    \includegraphics[height=\subfight]{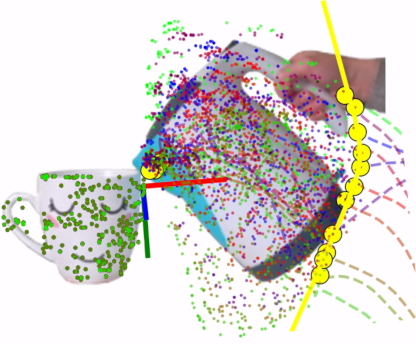}};
                \node at (-0.40, -0.45) {\scalebox{0.9}{$\lf$}};
                \node at (-0.85,  0.15) {\scalebox{0.9}{$\bk_1$}};
                \node at ( 1.50, -0.80) {\scalebox{0.9}{$\bk_2$}};
                \node at (-1.70,  1.30) {\scalebox{0.8}{\circled{11}}};
                \node at ( 0.30, -1.20) {\scalebox{0.9}{$\manifold{\text{curve}}$}};
            \end{tikzpicture}
            \vspace{-1em}
	}
	\caption{Pouring task with variations in the shape of cups and the orientation of the kettle. The principal plane in \subref{subfig:pour_3} is represented by orthogonal vectors \smalleq{$\mathcal{M}_{\text{plane},x}$} (\sampleline{red,line width=0.5mm}) and \smalleq{$\mathcal{M}_{\text{plane},y}$} (\sampleline{green,line width=0.5mm}). Other legends as in \cref{fig:press_button}.
	}
	\label{fig:pouring}
	\vspace{-1.5ex}
\end{figure}

\textbf{Fetch Tissue} (\taskabbr{FT}):
A human demonstrates how to fetch tissue from two tissue boxes (\#3 and \#4 in \cref{subfig:tisse}) with different hand poses (see \cref{fig:fetch_tissue}). The tissue boxes \#1-3 are used to train the vision models, whereas \#4-10 are used for reproduction. 
The reproduction is considered successful if the robot can successfully grasp the tissue and pull it out of the boxes with a predefined pulling action. 
Although the shape variations between the tissue boxes \mbox{\#3-4} in the demonstrations are not obvious, box \#10 introduces large shape variations for reproduction.

\textbf{Insert Stick} (\taskabbr{IS}):
The sticks \#2-4 in \cref{subfig:sticks} are used to train the vision models and to demonstrate the insertion task by a human (see \cref{fig:insertion}). 
Note that we do not insert the stick into the paper roll, as otherwise the keypoints will be occluded (we defer tasks with occlusion to future work).
No pose variations are considered in this task. However, shape variations are introduced via sticks of different lengths and thicknesses.
Moreover, we place the sticks with an initial tilting of $\sim$ \ang{-30} to \ang{80} in the reproduction, thus extrapolating the demonstration range ($\sim$ \ang{0} to \ang{45}).
A successful reproduction 
is obtained by placing the lower tip of the stick right above the hole in the center of the paper roll without collision with the paper roll during execution (see \cref{subfig:paper_roll}).

\textbf{Pour Water} (\taskabbr{PW}):
The vision models for this task are trained with the teacups \#5-8 in \cref{subfig:cups} and the kettles \#4-5 in \cref{subfig:kettles}. A human demonstrates the pouring task several times with the kettle \#5 and teacups \#1 and \#3. The demonstrations incorporate teacup shape variations and kettle pose variations (see \cref{fig:pouring}).
The teacups \#1-4 and all the kettles are used in reproduction.
The reproduction of the \taskabbr{PW} task is successful if the spout of the kettle aligns above the rim of the teacup and the kettle is tilted appropriately.

\begin{figure}[t]
	\setlength{\subfight}{1.1cm}
	\sbox\subfigbox{%
		\resizebox{0.95\columnwidth}{!}{
			\includegraphics[height=\subfight]{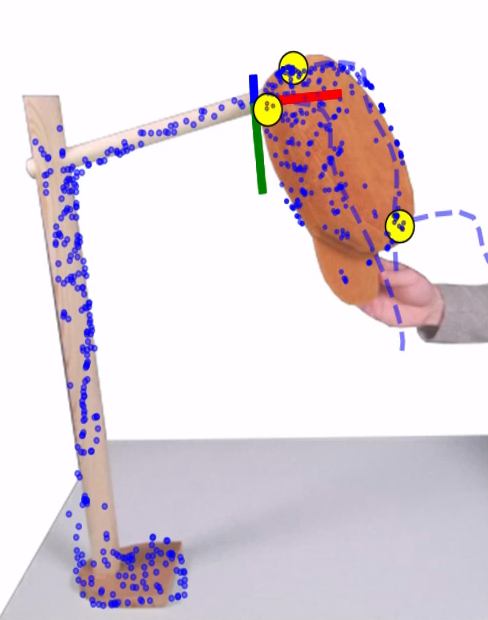}
			\includegraphics[height=\subfight]{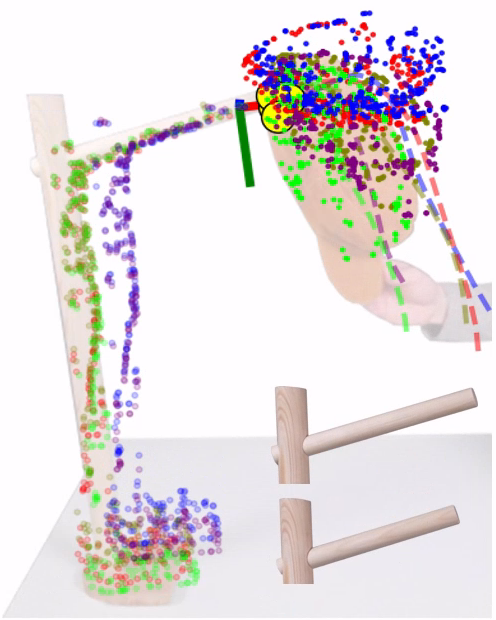}
			\includegraphics[height=\subfight]{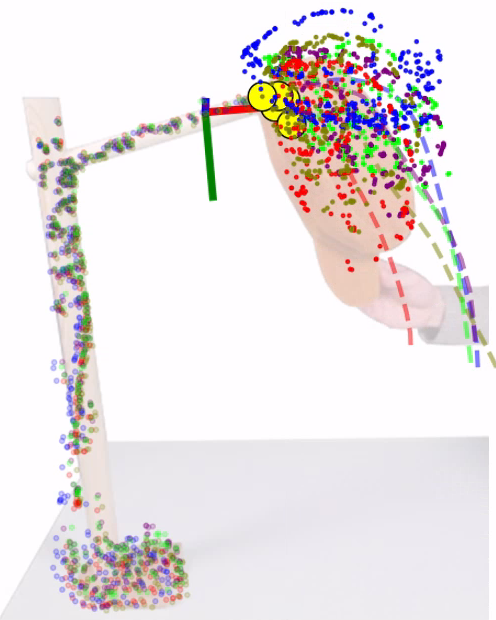}
		}
	}
	\setlength{\subfight}{\ht\subfigbox}
	\centering
	\subcaptionbox{$\mathsf{p2p}$: $\bk_1, \bk_2, \bk_3$\label{subfig:hat_1}}{
	    \begin{tikzpicture}
            \hspace{-2.5pt}
            \node (image) at (0, 0) {
                \includegraphics[height=\subfight]{figure/real/hang_hat/cfg/kvil_demo_1.png}
            };
            \node at (-0.15, 0.55) {\scalebox{0.9}{$\lf$}};
            \node at (-0.15, 1.20) {\scalebox{0.9}{$\bk_1$}};
            \node at ( 1.00, 0.30) {\scalebox{0.9}{$\bk_2$}};
            \node at ( 0.60, 1.30) {\scalebox{0.9}{$\bk_3$}};
            \node at (-1.00, 1.30) {\circled{1}};
        \end{tikzpicture}
            \vspace{-0.8em}
	}
	\hfill
	\subcaptionbox{$\mathsf{p2p}$: $\bk_1$\label{subfig:hat_3}}{
	    \begin{tikzpicture}
            \hspace{-2.5pt}
            \node (image) at (0, 0) {
                \includegraphics[height=\subfight]{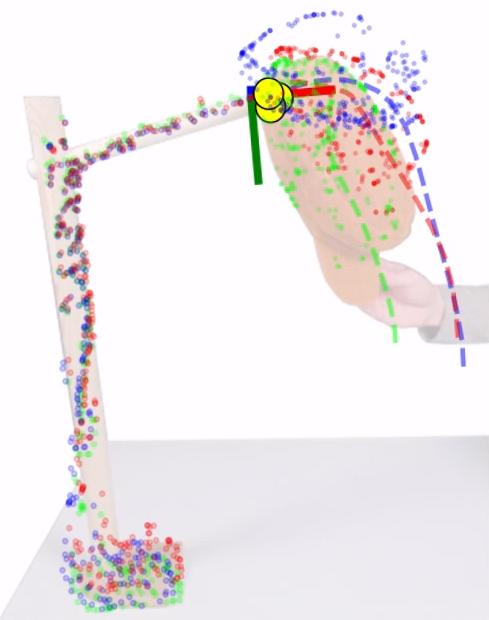}
            };
            \node at (-0.15, 0.55) {\scalebox{0.9}{$\lf$}};
            \node at (-0.15, 1.20) {\scalebox{0.9}{$\bk_1$}};
            \node at (-1.00, 1.30) {\circled{3}};
        \end{tikzpicture}
            \vspace{-0.8em}
	}
	\hfill
	\subcaptionbox{$\mathsf{p2p}$: $\bk_1$\label{subfig:hat_4}}{
	    \begin{tikzpicture}
            \hspace{-2.5pt}
            \node (image) at (0, 0) {
                \includegraphics[height=\subfight]{figure/real/hang_hat/cfg/kvil_demo_4_long_short.png}
            };
            \node at (-0.15,  0.55) {\scalebox{0.9}{$\lf$}};
            \node at (-0.15,  1.20) {\scalebox{0.9}{$\bk_1$}};
            \node at (-1.00,  1.30) {\circled{4}};
            \node at ( 0.80, -0.75) {\#7};
            \node at ( 0.80, -1.30) {\#8};
        \end{tikzpicture}
            \vspace{-0.8em}
	}

    \subcaptionbox{$\mathsf{p2p}$: $\bk_1$\label{subfig:hat_5}}{
	    \begin{tikzpicture}
            \hspace{-2.5pt}
            \node (image) at (0, 0) {
                \includegraphics[height=\subfight]{figure/real/hang_hat/cfg/kvil_demo_5.png}
            };
            \node at (-0.4, 0.55) {\scalebox{0.9}{$\lf$}};
            \node at (-0.3, 1.20) {\scalebox{0.9}{$\bk_1$}};
            \node at (-1.0, 1.30) {\circled{5}};
        \end{tikzpicture}
            \vspace{-0.8em}
	}
    \hfill
	\subcaptionbox{Shape Variations of the Hat\label{subfig:hat_shape}}{
	    \includegraphics[height=\subfight]{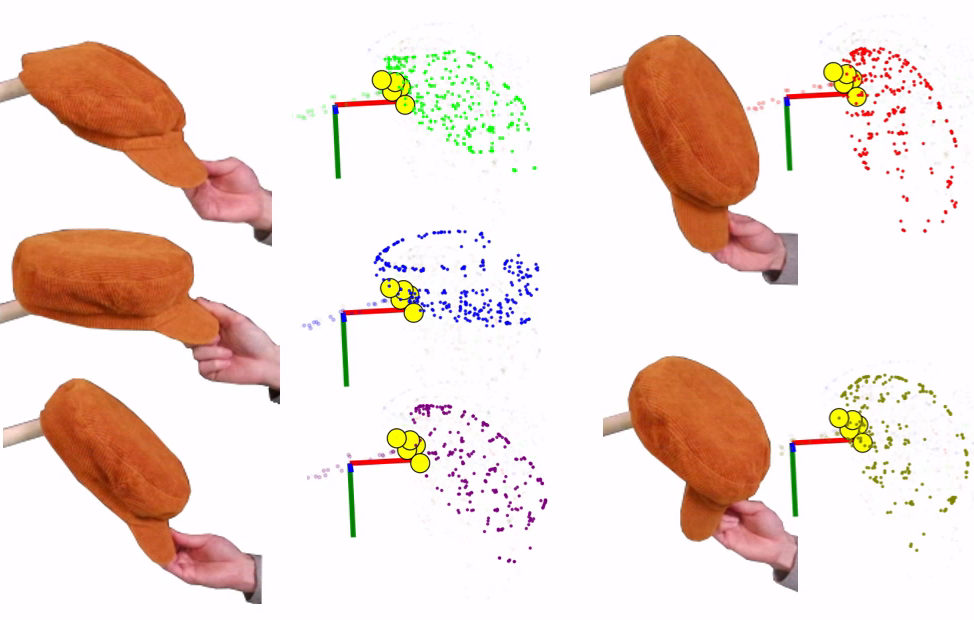}
	}
	\caption{Hang a hat on a rack. Note that there are only slightly deformations of the hat in \subref{subfig:hat_1}-\subref{subfig:hat_4} and relatively more obvious deformations in \subref{subfig:hat_5} (see \subref{subfig:hat_shape}), while pose variations in the hats are considered in all cases. The racks in \subref{subfig:hat_4} have two shape variations. 
	Legend as \cref{fig:press_button}.
	}
	\label{fig:hanging}
	\vspace{-1.5ex}
\end{figure}

\textbf{Hang Hat} (\taskabbr{HH}):
The rack can be assembled with different lengths of sticks (\#6-10 in \cref{subfig:sticks}). The racks assembled with sticks \#7-8 are used for training the vision models and for the demonstrations and \#6-10 are used for the reproduction.
In particular, the stick \#7 is used in \cref{subfig:hat_1,subfig:hat_3,subfig:hat_5} and the sticks \#7-8 are used in \cref{subfig:hat_4}.
Successful reproductions are observed if the rim of the hat is placed on top of the tip of the stick regardless of the stick length and the initial pose of the hat.

\textbf{Clean Table} (\taskabbr{CT}):
The dustpans \#2-3 and the brushes \#3-4 in \cref{subfig:dustpan_brush} are used for training the vision models and for the demonstrations, while the dustpan \#1 and the brushes \#1-2 are used for the reproduction. 
The \taskabbr{CT} is successful if the head of the brush aligns parallel above the edge of the dustpan.

\subsection{Evaluation of K-VIL's Task Representation}
\label{sec:eval_task_repr}

As discussed in \cref{sec:pce}, K-VIL's task representation can be acquired from one or a few demonstration videos based on the distance and variance criteria. 
Therefore, the number of demonstrations and the variations in object poses and shapes play an essential role.
In this evaluation, we are interested in the following questions:
\begin{enumerate*}[label=(\roman*)]
    \item How do task representations learned from a different number of demonstrations affect the performance of task reproduction? In other words, what are the limitations of the task representations learned from scarce demonstrations?
    \item How many demonstrations are required to learn generalizable task representations?
    \item How do the shape and pose variations contribute to the successful extraction of such task representations?
\end{enumerate*}
We evaluate K-VIL in one-shot and few shots imitation learning setups in \cref{sec:one_shot} and \cref{sec:incremental}, respectively. 
We finally answer the above questions in \cref{sec:eval_task_repr_summary}.

\subsubsection{One-shot Imitation Learning}\label{sec:one_shot}
We first apply K-VIL to one-shot imitation learning scenarios, where one demonstration (\smalleq{$N=1$}) is provided for each task. As previously explained, when a single demonstration is provided, K-VIL learns a task representation based on the distance criteria of \cref{sec:criteria_dist}, resulting in a set of $3$ linear $\ptop$ constraints.

\textbf{Task extraction}: The insertion task is first learned from a single demonstration consisting of inserting the stick \#4 of \cref{subfig:sticks} in a paper roll. In this task, the \master{} object is the paper roll, and the \slave{} object is the stick.
As shown in \cref{subfig:insert_1}, K-VIL extracts $3$ keypoints subject to $\ptop$ constraints on the stick. Note that the local frame $\lf$ and the keypoint $\bk_1$ are located near the contact point, and $\bk_2$ is the farthest point on the stick from the paper roll (i.e., the \master{} object).
Similarly, in \cref{subfig:press_1,subfig:tissue_1,subfig:pour_1,subfig:hat_1}, local frames are constructed on the \master{} objects (i.e., the kettle, the tissue box, the teacup, and the rack, respectively) and $3$ $\ptop$ constraints are extracted to fully constrain the pose of the \slave{} objects (i.e., the hand, the hand, the kettle and the hat, respectively).  

\textbf{Task reproduction}: As described in \cref{kac:priority}, the priorities of the $3$ keypoints are ranked as \scaleeq{0.9}{\Pri_1 > \Pri_2 > \Pri_3}. 
This respects the fact that $\bk_1$ is usually the contact point of two objects, 
and allows the KAC to reproduce the motion of $\bk_1$ more accurately than the motions of $\bk_2$ and $\bk_3$.
The first five columns of \cref{table:insert_priority} show examples of reproduction of the insertion task \taskabbr{IS} learned from a single demonstration (task N: \task{IS}{1}), as well as reproductions obtained by removing the priority from KAC as an ablation study (task O: \task{IS_{np}}{1}). 
For illustration purposes, we display the cases with a short stick (\#10), a long stick (\#6, longer than the sticks used in the demonstration), and an extra long stick (the concatenation of \#8 and \#9). 
The largest length difference is $\sim$\SI{300}{\milli\meter}.
The images in the first row show the ability of K-VIL to correctly adapt the task representations of an insertion task in new scenes. 
This includes constructing the local frame $\lf$ on the \master{} object (the paper roll) and the three $\ptop$ geometric constraints in $\lf$, identifying the keypoints on the \slave{} object (the sticks) and generating the VMP trajectories.
It is important to notice that, without priority in the KAC, the keypoint $\bk_1$ in \task{IS_{np}}{1}-\emph{short} is not able to reach its target as accurately as in \task{IS}{1}-\emph{short}.
Moreover, as opposed to \task{IS_{np}}{1}-\emph{long}, \task{IS}{1}-\emph{long} is successfully executed without collision between the stick and the paper roll, thanks to the priority in KAC.
However, both \task{IS}{1}-\emph{ext. long} and \mbox{\task{IS_{np}}{1}-\emph{ext. long}} result in a collision between the stick and the paper roll during execution due to the extended length of the stick compared to the demonstrations.
The collision is more acute without priority in the KAC.
\begin{figure}[t]
    \centering
    \setlength\tabcolsep{0.5pt}
    \tabulinesep=0pt
    \begin{tabu}{M{0.33\columnwidth}M{0.33\columnwidth}M{0.33\columnwidth}}
        \begin{subfigure}{0.33\columnwidth}
            \centering
            \begin{tikzpicture}
                \node[inner sep=0mm] (image) at (0, 0) {
                    \includegraphics[width=\textwidth]{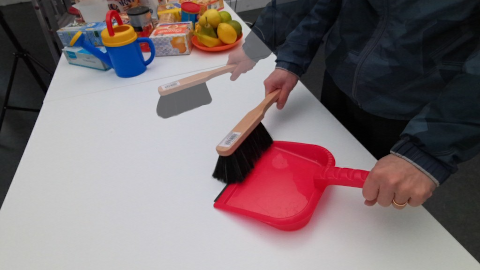}
                };
                \drawlabel{(a)}{(1.15, 0.55)}{fill opacity=0.5}
            \end{tikzpicture}
            \phantomsubcaption
            \label{subfig:ct_demo1}
        \end{subfigure} &
        \begin{subfigure}{0.33\columnwidth}
            \centering
            \begin{tikzpicture}
                \node[inner sep=0mm] (image) at (0, 0) {
                    \includegraphics[width=\textwidth]{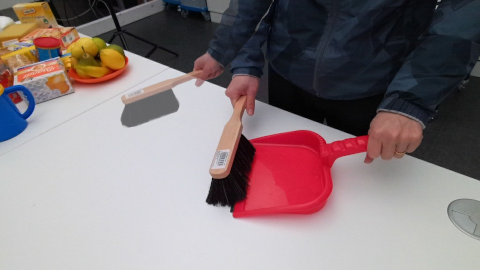}
                };
                \drawlabel{(b)}{(1.15, 0.55)}{fill opacity=0.5}
            \end{tikzpicture}
            \phantomsubcaption
            \label{subfig:ct_demo2}
        \end{subfigure} &
        \begin{subfigure}{0.33\columnwidth}
            \centering
            \begin{tikzpicture}
                \node[inner sep=0mm] (image) at (0, 0) {
                    \includegraphics[width=\textwidth]{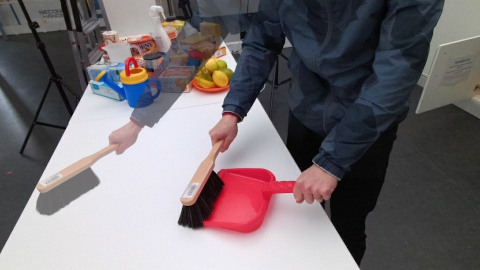}
                };
                \drawlabel{(c)}{(1.15, 0.55)}{fill opacity=0.5}
            \end{tikzpicture}
            \phantomsubcaption
            \label{subfig:ct_demo3}
        \end{subfigure}
        \\
        \multicolumn{2}{c}{
            \begin{subfigure}{0.6\columnwidth}
                \vspace{-13pt}
                \centering
                \begin{tikzpicture}
                    \node[inner sep=0mm] (image) at (0, 0) {
                        \includegraphics[height=2cm]{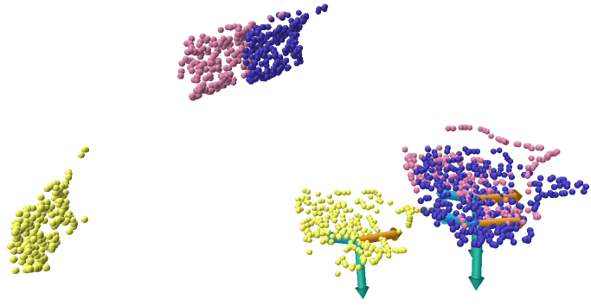}
                    };
                    \node at ( 0.1, -0.85) {\scalebox{0.8}{$\lf_c$}};
                    \node at ( 1.5, -0.75) {\scalebox{0.8}{$\lf_b$}};
                    \node at ( 1.7, -0.45) {\scalebox{0.8}{$\lf_a$}};
                    
                    \node at ( 2.0,  0.2) {\scalebox{0.8}{dustpans}};
                    \node at (-0.8,  0.1) {\scalebox{0.8}{brushes}};
                    \drawlabel{(d)}{(3.35, 0.72)}{fill opacity=0.5};
                \end{tikzpicture}
                \phantomsubcaption
                \label{subfig:ct_mismatch}
            \end{subfigure}
        } &
        \begin{subfigure}{0.33\columnwidth}
            \vspace{-10pt}
            \centering
            \begin{tikzpicture}
                \node[inner sep=0mm] (image) at (0, 0) {
                    \includegraphics[width=\textwidth, trim=0 5 0 10, clip]{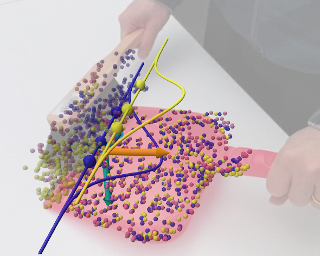}
                };
                \drawlabel{(e)}{(1.15, 0.82)}{fill opacity=0.5}
            \end{tikzpicture}
            \phantomsubcaption
            \label{subfig:ct_constraints}
        \end{subfigure}
        \\
        \begin{subfigure}{0.33\columnwidth}
            \vspace{-10pt}
            \centering
            \begin{tikzpicture}
                \node[inner sep=0mm] (image) at (0, 0) {
                    \includegraphics[width=\textwidth]{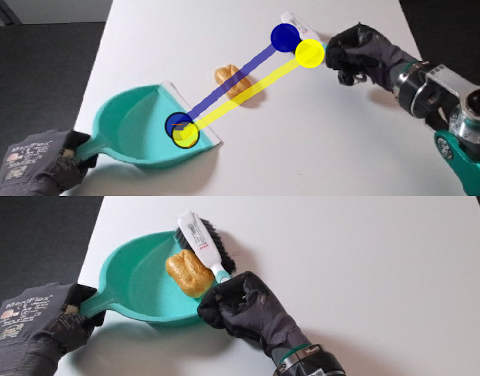}
                };
                \drawlabel{(f)}{(1.15, 0.88)}{fill opacity=0.5}
            \end{tikzpicture}
            \phantomsubcaption
            \label{subfig:ct_deploy}
        \end{subfigure}
        &
        \begin{subfigure}{0.33\columnwidth}
            \vspace{-21.5pt}
            \centering
            \begin{tikzpicture}
                \hspace{-17pt}
                \node[inner sep=0mm] (image) at (0, 0) {
                    \includegraphics[height=2.3cm]{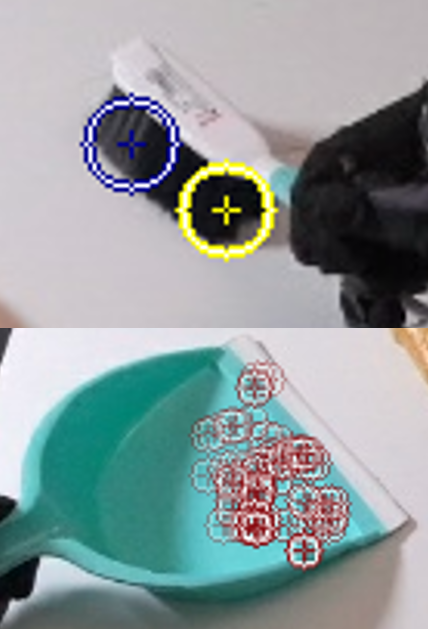}
                };
                \node at (-0.55, 0.35) {\scalebox{0.8}{$\bk_1$}};
                \node at (-0.20, 0.15) {\scalebox{0.8}{$\bk_2$}};
                \drawlabel{(g)}{(0.45, 0.88)}{fill opacity=0.5}
            \end{tikzpicture}
            \phantomsubcaption
            \label{subfig:ct_kpt}
        \end{subfigure}
        &
        \begin{subfigure}{0.33\columnwidth}
            \vspace{-10pt}
            \centering
            \begin{tikzpicture}
                \hspace{-36pt}
                \node[inner sep=0mm] (image) at (0, 0) {
                    \includegraphics[height=2.3cm]{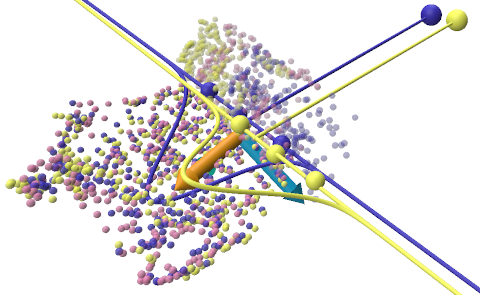}
                };
                \drawlabel{(h)}{(2.05, 0.86)}{fill opacity=0.5}
            \end{tikzpicture}
            \phantomsubcaption
            \label{subfig:ct_deploy_constraint}
        \end{subfigure}
    \end{tabu}
    \vspace{-1.2em}
    \caption{
            K-VIL handles viewpoint mismatch in the three demonstrations \subref{subfig:ct_demo1}-\subref{subfig:ct_demo3} by aligning the corresponding local frames on the master object dustpan in \subref{subfig:ct_mismatch}, which results in an aligned viewpoint in \subref{subfig:ct_constraints}. 
            Two $\ptol$ constraints and their probability density functions on the principal lines are visualized. 
            The robot reproduces the \task{CT}{3} task from a new viewpoint with a novel brush and dustpan \subref{subfig:ct_deploy}, with the keypoints \mbox{(\inlinekpt{blue!80}{fill=blue!80}$\bk_1$, \inlinekpt{yellow!80}{fill=yellow!80}$\bk_2$)} detected on the brush hair \subref{subfig:ct_kpt}.
            The local frame on the dustpan is determined by the $Q=50$ neighboring points as shown in \subref{subfig:ct_kpt}. 
            \subref{subfig:ct_deploy} and \subref{subfig:ct_deploy_constraint} depict the keypoints and their movement primitives in 2D and 3D respectively.
    }
    \label{fig:clean_table_task}
    \vspace{-1.5ex}
\end{figure}

\begin{table*}
    \centering
    \setlength\tabcolsep{0.5pt}
    \setlength\imgwidth{19mm}
    \begin{tabular}{cM{\imgwidth}M{\imgwidth}M{\imgwidth}M{\imgwidth}M{\imgwidth}M{\imgwidth}M{\imgwidth}M{\imgwidth}M{\imgwidth}}
       \toprule
            Tasks 
            & N: \task{IS}{1} 
            & O: \task{IS_{np}}{1} 
            & N: \task{IS}{1} 
            & N: \task{IS}{1} 
            & O: \task{IS_{np}}{1} 
            & P: \task{IS}{3}
            & Q: \task{IS_{np}}{3} 
            & P: \task{IS}{3} 
            & Q: \task{IS_{np}}{3}
            \\
        \midrule
            stick
            & short
            & short
            & long
            & ext. long
            & ext. long
            & short
            & short
            & ext. long
            & ext. long
            \\
        \midrule
            \multirow{2}{*}[-4em]{\rotatebox[origin=c]{90}{TR}}
                & \multicolumn{5}{c|}{3$\times \ptop$}
                & \multicolumn{4}{c}{$\ptop, \ptol$}
            \\
            & \begin{tikzpicture}
                \hspace{-2.5pt}
                \node (image) at (0, 0) {
                    \includegraphics[width=\imgwidth]{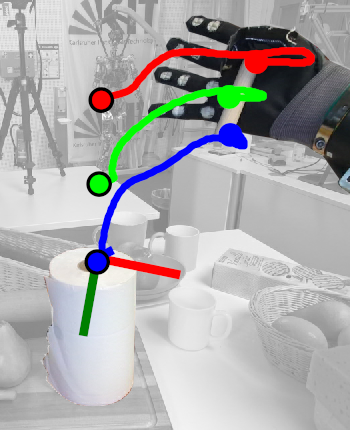}
                };
                \node at (-0.45, -0.60) {$\lf$};
            \end{tikzpicture}
            & \begin{tikzpicture}
                \hspace{-2.5pt}
                \node (image) at (0, 0) {
                    \includegraphics[width=\imgwidth]{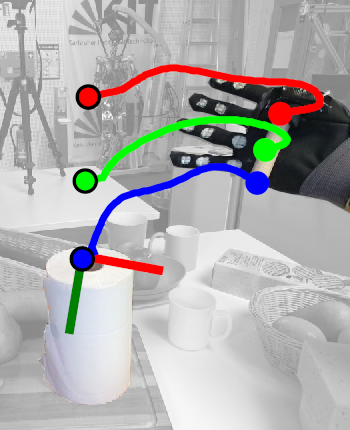}
                };
                \node at (-0.45, -0.60) {$\lf$};
            \end{tikzpicture}
            & \begin{tikzpicture}
                \hspace{-2.5pt}
                \node (image) at (0, 0) {
                    \includegraphics[width=\imgwidth]{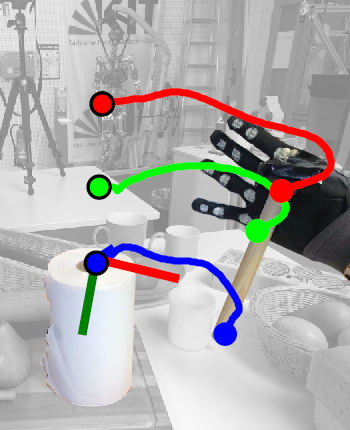}
                };
                \node at (-0.45, -0.60) {$\lf$};
            \end{tikzpicture}
            & \begin{tikzpicture}
                \hspace{-2.5pt}
                \node (image) at (0, 0) {
                    \includegraphics[width=\imgwidth]{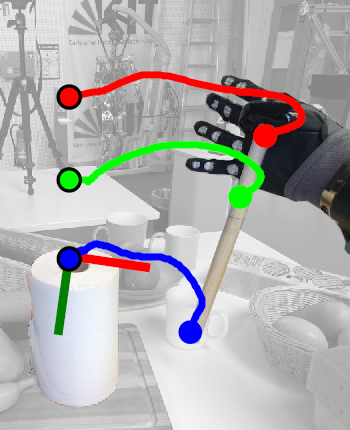}
                };
                \node at (-0.45, -0.60) {$\lf$};
            \end{tikzpicture}
            & \begin{tikzpicture}
                \hspace{-2.5pt}
                \node (image) at (0, 0) {
                    \includegraphics[width=\imgwidth]{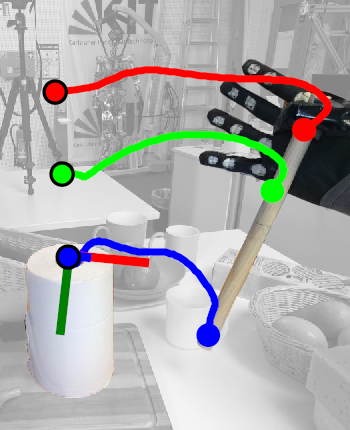}
                };
                \node at (-0.45, -0.60) {$\lf$};
            \end{tikzpicture}
            & \begin{tikzpicture}
                \hspace{-2.5pt}
                \node (image) at (0, 0) {
                    \includegraphics[width=\imgwidth]{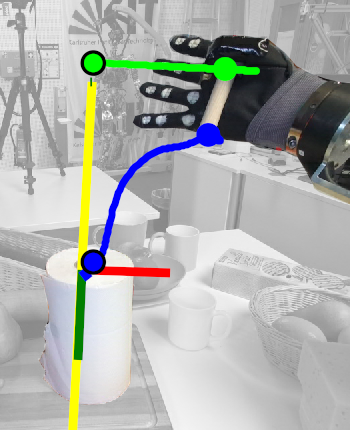}
                };
                \node at (-0.45, -0.60) {$\lf$};
            \end{tikzpicture}
            & \begin{tikzpicture}
                \hspace{-2.5pt}
                \node (image) at (0, 0) {
                    \includegraphics[width=\imgwidth]{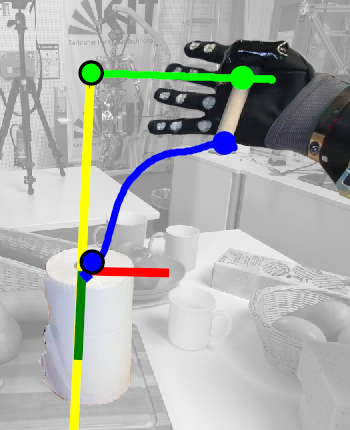}
                };
                \node at (-0.45, -0.60) {$\lf$};
            \end{tikzpicture}
            & \begin{tikzpicture}
                \hspace{-2.5pt}
                \node (image) at (0, 0) {
                    \includegraphics[width=\imgwidth]{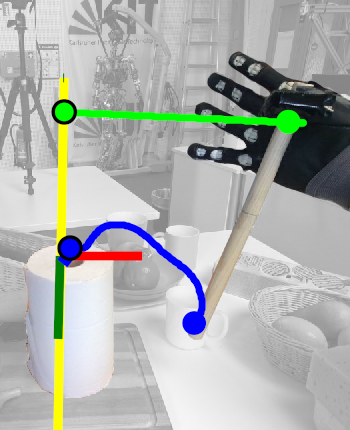}
                };
                \node at (-0.45, -0.60) {$\lf$};
            \end{tikzpicture}
            & \begin{tikzpicture}
                \hspace{-2.5pt}
                \node (image) at (0, 0) {
                    \includegraphics[width=\imgwidth]{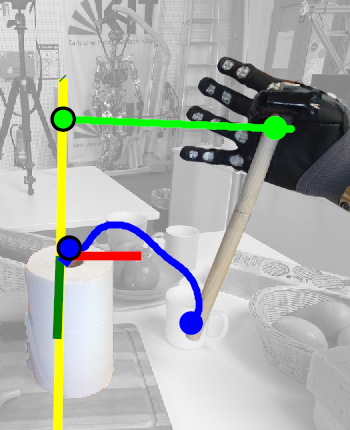}
                };
                \node at (-0.45, -0.60) {$\lf$};
            \end{tikzpicture}
        \\
        \rotatebox[origin=c]{90}{Reproduction}
            & \includegraphics[width=\imgwidth]{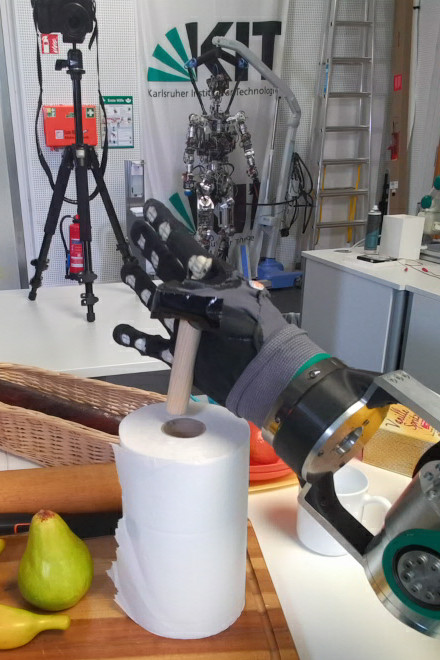}
            & \includegraphics[width=\imgwidth]{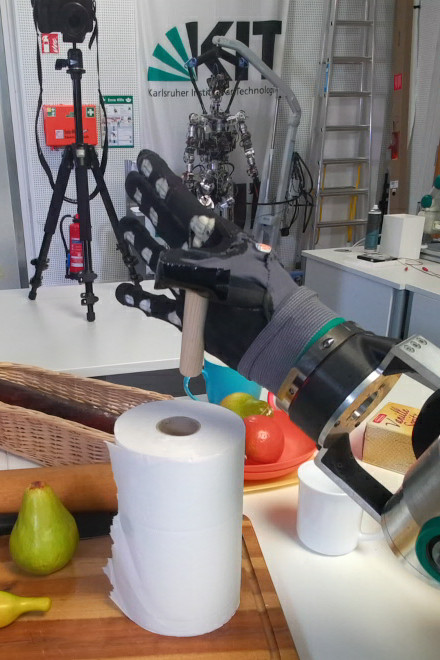}
            & \includegraphics[width=\imgwidth]{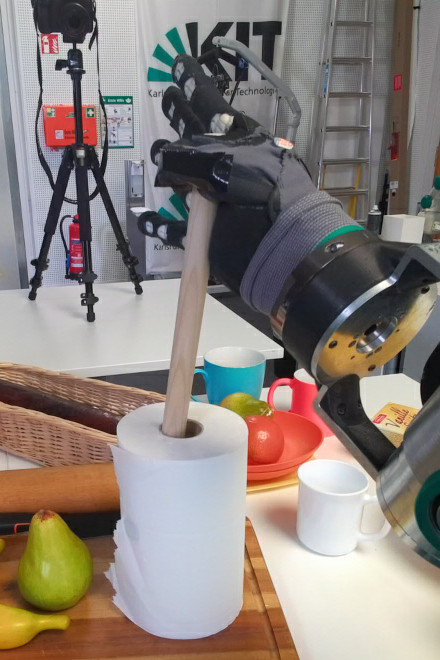}
            & \includegraphics[width=\imgwidth]{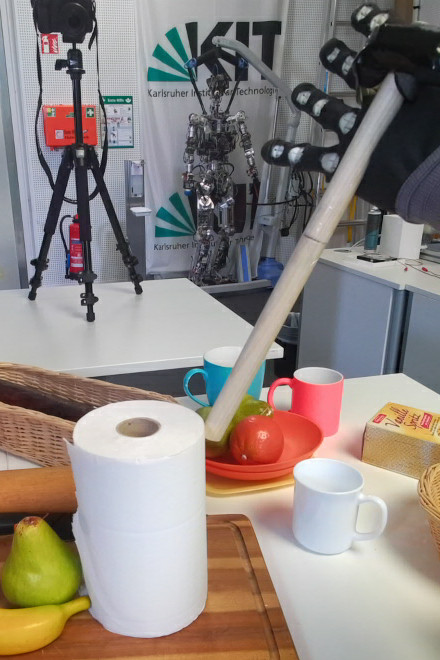}
            & \includegraphics[width=\imgwidth]{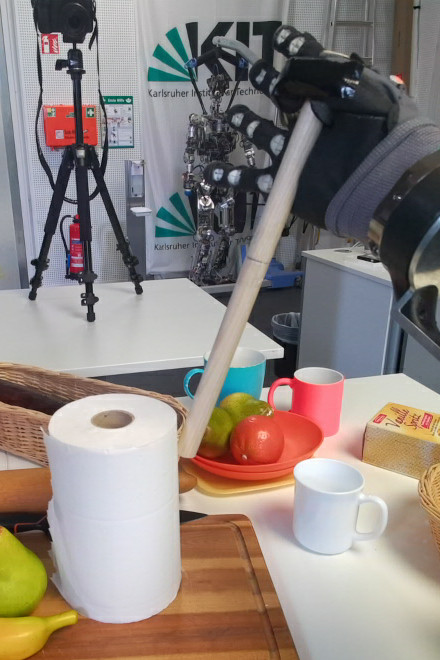}
            & \includegraphics[width=\imgwidth]{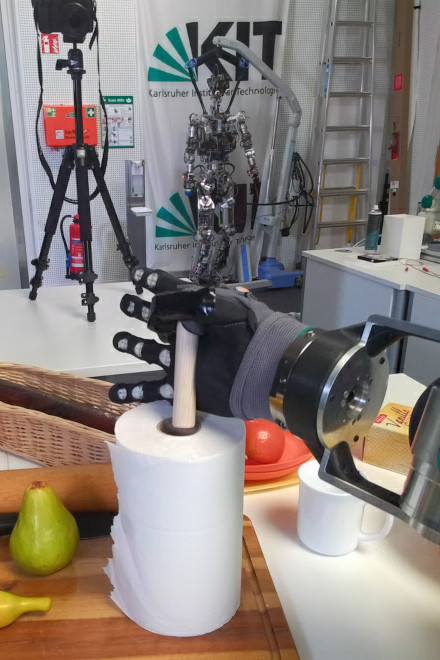}
            & \includegraphics[width=\imgwidth]{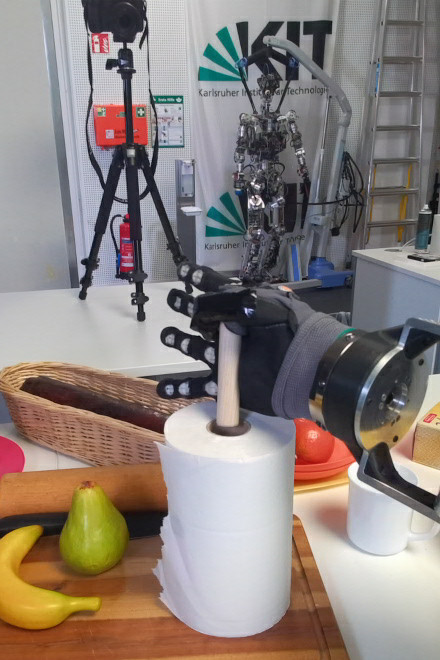}
            & \includegraphics[width=\imgwidth]{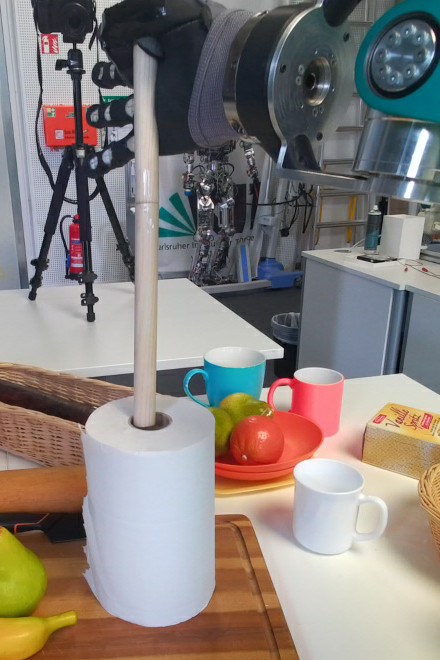} 
            & \includegraphics[width=\imgwidth]{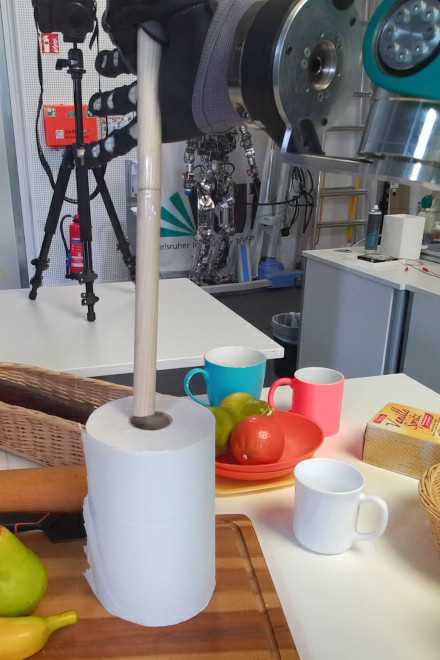} 
        \\
        \bottomrule
    \end{tabular}
    \caption{
    Reproduction of the insertion tasks with/without priorities. 
    Given an image of the scene before execution, K-VIL's task representation (TR) of each task is used to identify the local frame $\lf$, 
    the keypoints (\inlinekpt{blue!80}{fill=blue!80}$\bk_1$, \inlinekpt{red!80}{fill=red!80}$\bk_2$, \inlinekpt{applegreen!80}{fill=applegreen!80}$\bk_3$), 
    their targets positions (\inlinekpt{black}{fill=blue!80, line width=0.6pt}$\bk_1^g$, \inlinekpt{black}{fill=red!80, line width=0.6pt}$\bk_2^g$, \inlinekpt{black}{fill=applegreen!80, line width=0.6pt}$\bk_3^g$), 
    their movement primitives (\sampleline{blue},\sampleline{red},\sampleline{applegreen}), 
    and the line principal manifold (\sampleline{yellow(process),line width=2pt}) in tasks P and Q. 
    The short, long and extremely long sticks correspond to sticks \#10, \#6, and the concatenation of \#8 and \#9.
    Task names and statistics are listed in \cref{table:kac_eval}. The subscript $\text{np}$ indicates that the task was reproduced without priority in KAC. The figures in each column are from one of the 20 trials for each task.
    }
    \label{table:insert_priority}
    \vspace{-1ex}
\end{table*}

Moreover, one-shot VIL may generally fail when the learned geometric constraints are not reachable. This problem is exacerbated when demonstrations are provided from a third-person view. 
For example, consider \cref{subfig:press_1} and \cref{subfig:tissue_1}, that show the task representations learned from a single third-person-view demonstration of the \taskabbr{PB} and \taskabbr{FT} task, respectively. 
The reproductions of such task representations fail (see tasks A: \task{PB}{1} and D: \task{FT}{1} in \cref{table:failure_case_fetch_press}) due to unreachable target keypoint positions. 
This also indicates that task representations learned from one demonstration are not necessarily generalizable enough for motion reproduction. 

Despite a few failures in the execution, K-VIL's task representations are reliably adapted to new scenes. In other words, K-VIL is able to successfully identify the keypoint positions, locate their targets, and generate the corresponding VMPs by learning their representation thanks to the combination of the proposed task representation with dense visual correspondence models. It is worth noting that this is already achieved by learning the corresponding representation from a single demonstration.
Moreover, thanks to the prioritized KAC, \mbox{K-VIL} can handle shape variations in categorical objects via the extrapolation of the keypoint target positions. 
However, it cannot cope with very large shape variations. In other words, providing a single demonstration limits the learning of embodiment-independent generalizable task representations.
Therefore, we then evaluate the performance of K-VIL in the case where several demonstrations are available.

\begin{table*}
    \centering
    \setlength\tabcolsep{0.5pt}
    \setlength\imgwidth{28mm}
    \begin{tabular}{cM{\imgwidth}M{\imgwidth}M{\imgwidth}M{\imgwidth}M{\imgwidth}M{\imgwidth}}
       \toprule
            Tasks 
            & A: \task{PB}{1}
            & B: \task{PB}{3} 
            & C: \task{PB}{4} 
            & D: \task{FT}{1} 
            & E: \task{FT}{3}
            & F: \task{FT}{4}
        \\
        \midrule
            \multirow{2}{*}[-2.5em]{\rotatebox[origin=c]{90}{TR}}
            & 3$\times \ptop$
            & $\ptop, \ptol$
            & $\ptop$
            & 3$\times \ptop$
            & $\ptop, \ptol$
            & $\ptop$
        \\
            & \begin{tikzpicture}
                \hspace{-2.5pt}
                \node (image) at (0, 0) {
                    \includegraphics[width=\imgwidth]{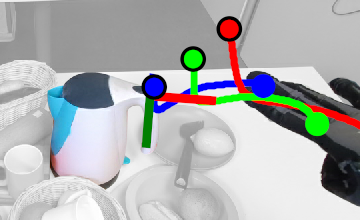} 
                };
                \node at (-0.45, -0.10) {$\lf$};
            \end{tikzpicture} 
            & \begin{tikzpicture}
                \hspace{-2.5pt}
                \node (image) at (0, 0) {
                    \includegraphics[width=\imgwidth]{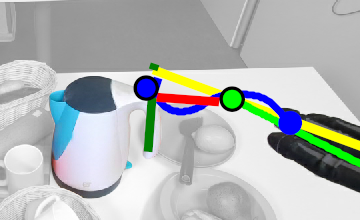}
                };
                \node at (-0.45, -0.10) {$\lf$};
            \end{tikzpicture} 
            & \begin{tikzpicture}
                \hspace{-2.5pt}
                \node (image) at (0, 0) {
                    \includegraphics[width=\imgwidth]{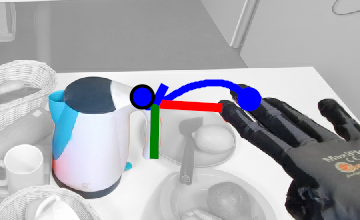}
                };
                \node at (-0.45, -0.10) {$\lf$};
            \end{tikzpicture}  
            & \begin{tikzpicture}
                \hspace{-2.5pt}
                \node (image) at (0, 0) {
                    \includegraphics[width=\imgwidth]{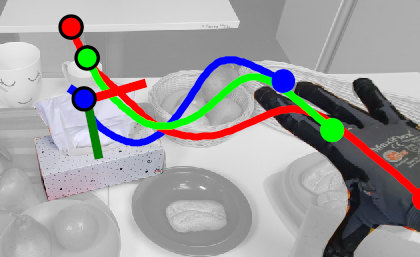}
                };
                \node at (-1.05, -0.0) {$\lf$};
            \end{tikzpicture}  
            & \begin{tikzpicture}
                \hspace{-2.5pt}
                \node (image) at (0, 0) {
                    \includegraphics[width=\imgwidth]{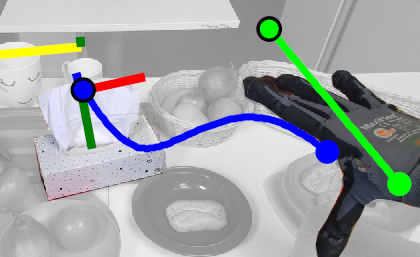}
                };
                \node at (-1.05, -0.0) {$\lf$};
                \draw[canaryyellow,line width=0.55mm] (-1.05, 0.535) -- (1.0, 0.7);
                \node at (0.34, 0.65) {\inlinekpt{black}{fill=green!80, line width=0.6pt}};
            \end{tikzpicture} 
            & \begin{tikzpicture}
                \hspace{-2.5pt}
                \node (image) at (0, 0) {
                    \includegraphics[width=\imgwidth]{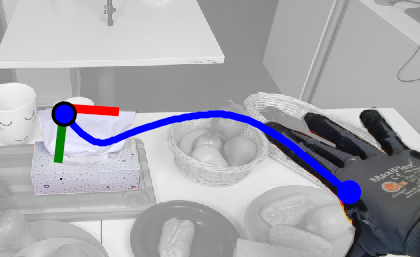}
                };
                \node at (-1.2, -0.10) {$\lf$};
            \end{tikzpicture} 
        \\
        \rotatebox[origin=c]{90}{Reproduction} 
            & \includegraphics[width=\imgwidth]{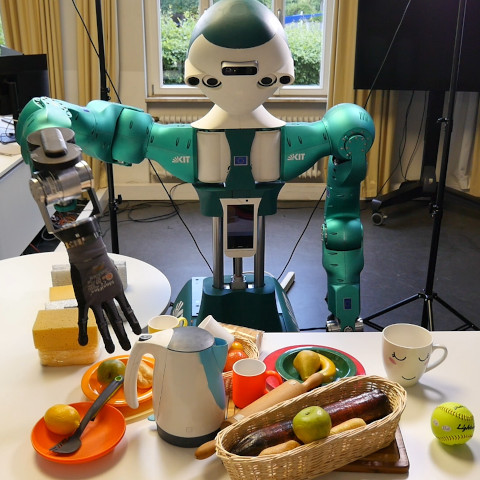} 
            & \includegraphics[width=\imgwidth]{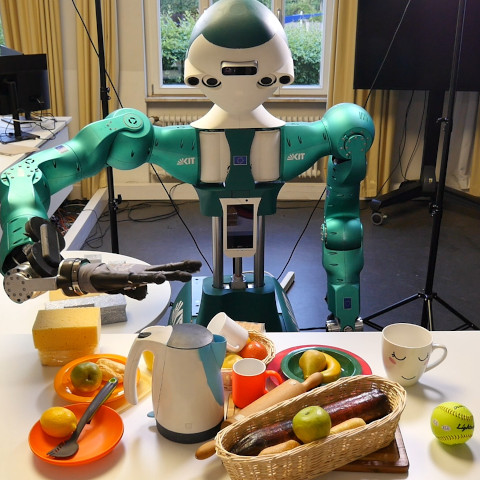} 
            & \includegraphics[width=\imgwidth]{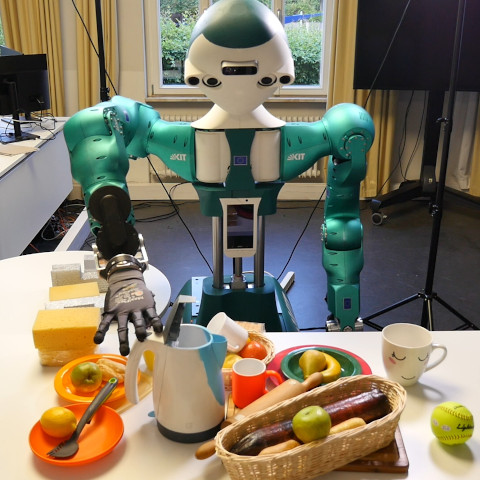} 
            & \includegraphics[width=\imgwidth]{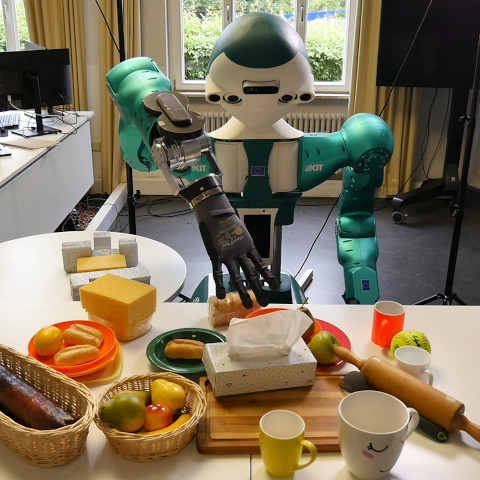}
            & \includegraphics[width=\imgwidth]{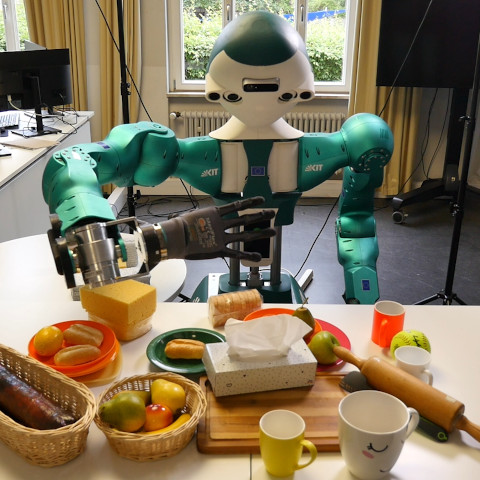} 
            & \includegraphics[width=\imgwidth]{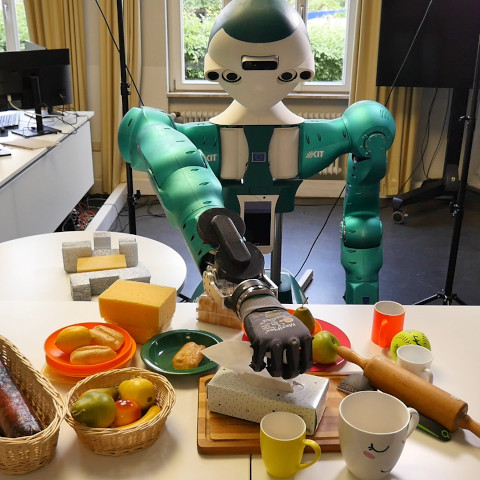}
        \\
        \bottomrule
    \end{tabular}
    \caption{
    Reproductions of tasks \taskabbr{PB} and \taskabbr{FT} learned from a third-person view without enough demonstrations lead to failure, due to unreachable geometric constraints, see tasks A, B, D, E. Generalizable task representations of \taskabbr{PB} and \taskabbr{FT} learned from enough demonstrations can be successfully executed in C and F.  Legend as in \cref{table:insert_priority}.
    }
    \label{table:failure_case_fetch_press}
    \vspace{-2ex}
\end{table*}

\subsubsection{Handling viewpoint mismatch}\label{sec:viewpoint_mismatch}

When demonstration videos are collected from different viewpoints, e.g., in the \task{CT}{3} task in \cref{fig:clean_table_task}, 
we first align the demonstrations into a common viewpoint by projecting the motions of the \slave{} objects (e.g., the brush) into each candidate local frame \smalleq{$\hat{\lf}_j$} on the \master{} object (e.g. the dustpan) (see also \cref{sec:criteria_dist}). 
In all \smalleq{$\vert \mathcal{P}_m \vert$} aligned common viewpoints, we apply PCE to extract the task representations. This solves the viewpoint mismatch problem of \cref{subfig:ct_mismatch}. The geometric constraints become obvious in the aligned viewpoint as shown in \cref{subfig:ct_constraints}.
K-VIL extracts two $\ptol$ constraints for the \task{CT}{3} task, the combination of which forms a parallel constraint. 
The estimated probability density functions of the two keypoints on the corresponding principal lines ensure their target positions to be above the edge of the dustpan.
It is important to note that, not only the demonstrations can be recorded from different viewpoints, but also the reproduction of the learned skill by the robot can be performed from a viewpoint that is significantly different from any demonstration, as shown in \cref{subfig:ct_deploy,subfig:ct_deploy_constraint}.
For the seek of clarity, we discuss the results of K-VIL in the aligned viewpoints in the remaining evaluations, although the demonstrations and reproductions happen in different viewpoints as discussed for the \task{CT}{3} task.

\begin{table*}[t]
    \centering
    \setlength\tabcolsep{0.5pt}
    \setlength\imgwidth{19mm}
    \setlength\imgheight{16mm}
    \begin{tabu}{cM{\imgwidth}M{\imgwidth}M{\imgwidth}M{\imgwidth}M{\imgwidth}M{\imgwidth}M{\imgwidth}M{\imgwidth}M{\imgwidth}}
       \toprule
            Tasks 
            & G: \task{PW}{1}
            & \multicolumn{2}{c}{H: \task{PW}{3}}  
            & \multicolumn{1}{c}{I: \task{PW}{4}} 
            & \multicolumn{5}{c}{J: \scaletask{PW}{11}{0.7}}
        \\
        \midrule
            \multirow{2}{*}[-3em]{\rotatebox[origin=c]{90}{TR}}
            & 3$\times \ptop$
            & \multicolumn{2}{|c}{$\ptop,\ptol$}
            & \multicolumn{1}{|c}{$\ptop,\ptoP$}
            & \multicolumn{5}{|c}{$\ptop,\ptoc$}
        \\
            & \begin{tikzpicture}
                \hspace{-2.5pt}
                \node (image) at (0, 0) {
                    \includegraphics[height=\imgheight, trim=50 0 30 0, clip]{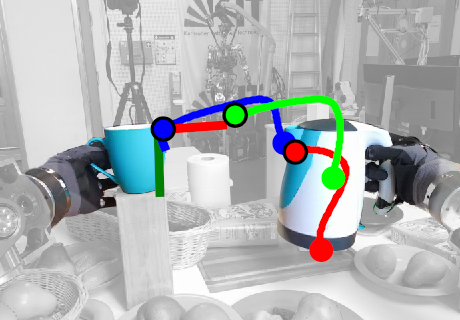} 
                };
                \node at (-0.2, -0.10) {$\lf$};
            \end{tikzpicture} 
            & \begin{tikzpicture}
                \hspace{-2.5pt}
                \node (image) at (0, 0) {
                    \includegraphics[height=\imgheight, trim=50 0 30 0, clip]{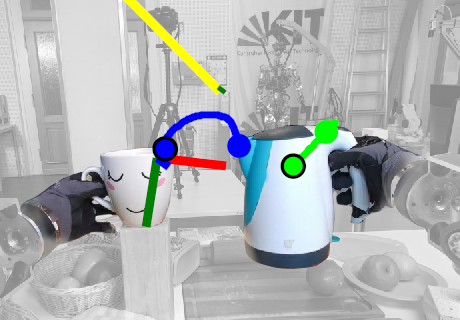}
                };
                \node at (-0.2, -0.20) {$\lf$};
                \draw[canaryyellow,line width=0.475mm] (-0.35, 0.6) -- (0.64, -0.4);
                \node at (0.23, -0.05) {\inlinekpt{black}{fill=green!60, line width=0.5pt,minimum size=0.08mm}};
                \node at (-0.16, 0.36) {\inlinekpt{black}{fill=yellow, line width=0.5pt,minimum size=0.08mm}};
            \end{tikzpicture} 
            & \begin{tikzpicture}
                \hspace{-2.5pt}
                \node (image) at (0, 0) {
                    \includegraphics[height=\imgheight, trim=60 0 20 0, clip]{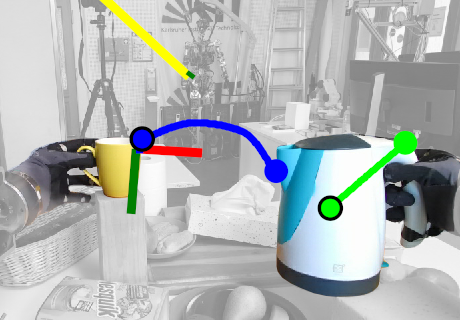}
                };
                \node at (-0.35, -0.15) {$\lf$};
                \draw[canaryyellow,line width=0.475mm] (-0.5, 0.6) -- (0.6, -0.4);
                \node at (0.37, -0.27) {\inlinekpt{black}{fill=green!40, line width=0.5pt,minimum size=0.08mm}};
                \node at (-0.35, 0.43) {\inlinekpt{black}{fill=yellow, line width=0.5pt,minimum size=0.08mm}};
            \end{tikzpicture}  
            & \begin{tikzpicture}
                \hspace{-2.5pt}
                \node (image) at (0, 0) {
                    \includegraphics[height=\imgheight, trim=40 0 40 0, clip]{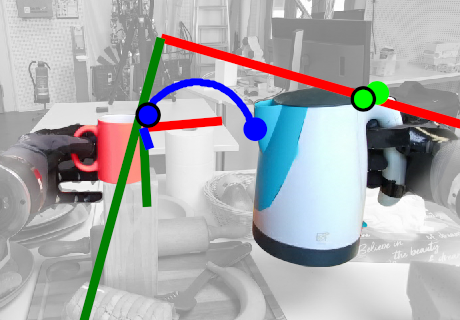}
                };
                \node at (-0.2, -0.05) {$\lf$};
                \node at (-0.41, 0.62) {\inlinekpt{black}{fill=yellow, line width=0.5pt,minimum size=0.08mm}};
            \end{tikzpicture}  
            & \begin{tikzpicture}
                \hspace{-2.5pt}
                \node (image) at (0, 0) {
                    \includegraphics[height=\imgheight, trim=0 50 15 50, clip]{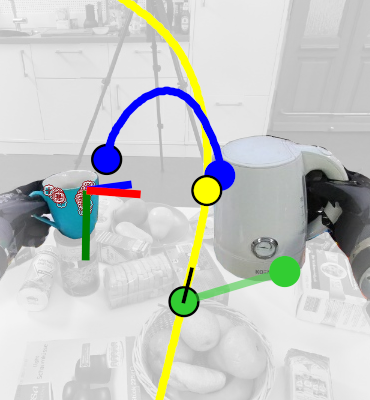}
                };
                \node at (-0.2, -0.15) {$\lf$};
            \end{tikzpicture}
            & \begin{tikzpicture}
                \hspace{-2.5pt}
                \node (image) at (0, 0) {
                    \includegraphics[height=\imgheight, trim=0 20 40 100, clip]{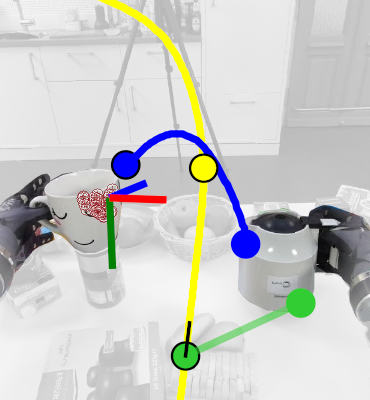}
                };
                \node at (-0.2, -0.15) {$\lf$};
            \end{tikzpicture} 
            & \begin{tikzpicture}
                \hspace{-2.5pt}
                \node (image) at (0, 0) {
                    \includegraphics[height=\imgheight, trim=15 70 40 60, clip]{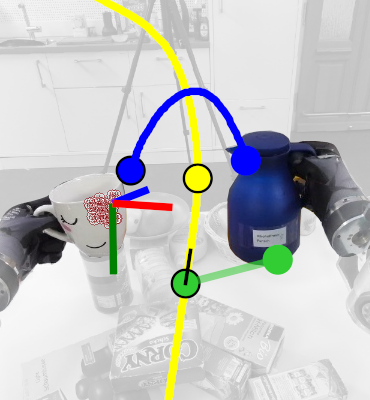}
                };
                \node at (-0.2, -0.15) {$\lf$};
            \end{tikzpicture} 
            & \begin{tikzpicture}
                \hspace{-2.5pt}
                \node (image) at (0, 0) {
                    \includegraphics[height=\imgheight, trim=0 50 20 50, clip]{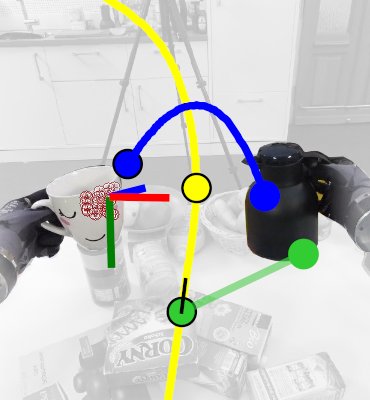}
                };
                \node at (-0.2, -0.15) {$\lf$};
            \end{tikzpicture} 
            & \begin{tikzpicture}
                \hspace{-2.5pt}
                \node (image) at (0, 0) {
                    \includegraphics[height=\imgheight, trim=55 20 60 10, clip]{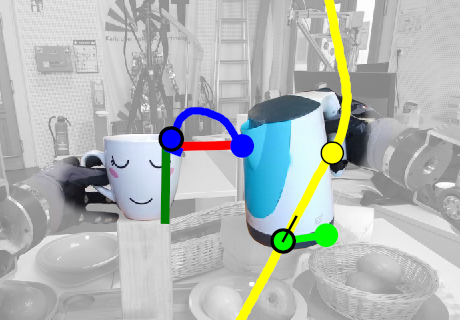}
                };
                \node at (-0.2, -0.15) {$\lf$};
            \end{tikzpicture} 
        \\
        \rotatebox[origin=c]{90}{Reproduction} 
            & \includegraphics[width=\imgwidth, trim=20 0 20 0, clip]{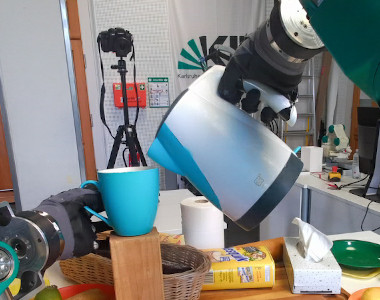} 
            & \includegraphics[width=\imgwidth, trim=20 0 20 0, clip]{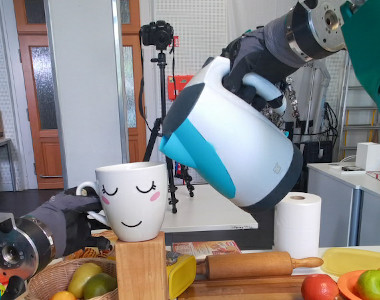} 
            & \includegraphics[width=\imgwidth, trim=20 0 20 0, clip]{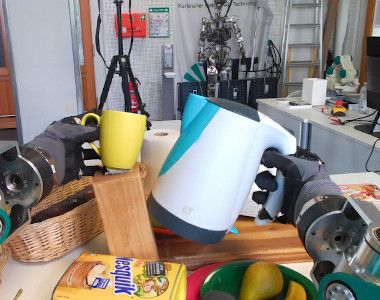} 
            & \includegraphics[width=\imgwidth, trim=20 0 20 0, clip]{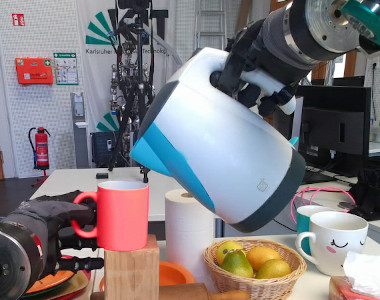}
            & \includegraphics[width=\imgwidth, trim=20 0 20 0, clip]{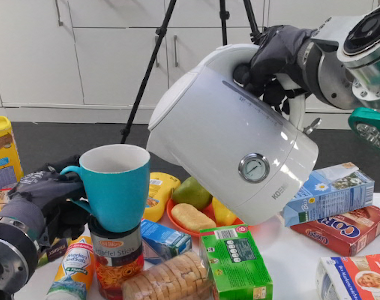}
            & \includegraphics[width=\imgwidth, trim=20 0 20 0, clip]{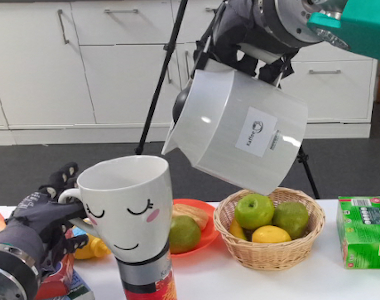}
            & \includegraphics[width=\imgwidth, trim=20 0 20 0, clip]{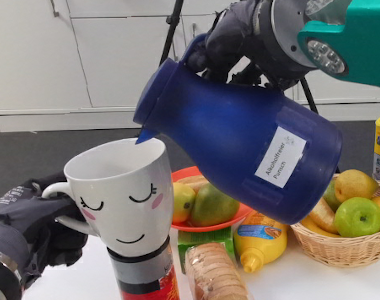}
            & \includegraphics[width=\imgwidth, trim=20 0 20 0, clip]{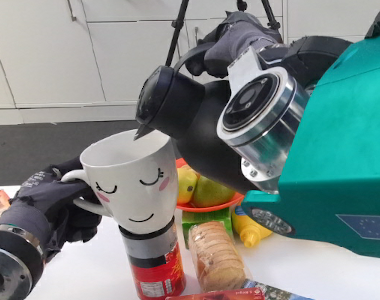}
            & \includegraphics[width=\imgwidth, trim=20 0 20 0, clip]{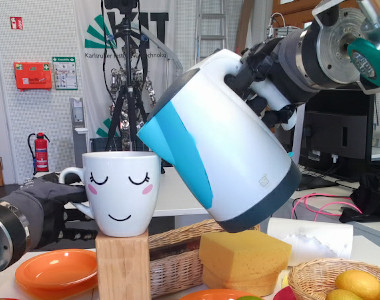}
        \\
        \bottomrule
    \end{tabu}
    \caption{
    Reproductions of tasks \taskabbr{PW} learned from different number of demonstrations. 
    For task H, I and J, we mark the learned principal manifold $\manifold{\text{line}}$ (\sampleline{yellow(process),line width=0.5mm}), $\manifold{\text{plane}}$ (\sampleline{red,line width=0.5mm}, \sampleline{ForestGreen,line width=0.5mm}) and $\manifold{\text{curve}}$ (\sampleline{yellow(process),line width=0.5mm}), respectively. 
    Additionally, the point (\inlinekpt{black}{fill=yellow(process), line width=0.5pt,minimum size=0.08mm}) indicates the mean of the demonstrated targets of keypoint $\bk_2$ on the corresponding principal manifolds.
    Other legends as in \cref{table:insert_priority}.
    }
    \label{table:pouring_repro}
    \vspace{-1ex}
\end{table*}
\begin{table*}
    \centering
    \setlength\tabcolsep{0.5pt}
    \setlength\imgwidth{24mm}
    \begin{tabular}{cM{\imgwidth}M{\imgwidth}M{\imgwidth}M{\imgwidth}M{\imgwidth}M{\imgwidth}}
       \toprule
            Tasks 
            & A: \task{HH}{1}
            & B: \task{HH}{1} 
            & C: \task{HH}{3} 
            & D: \task{HH}{3} 
            & E: \task{HH}{3}
            & F: \task{HH}{3}
        \\
        \midrule
            \multirow{2}{*}[-4em]{\rotatebox[origin=c]{90}{TR}}
            & \multicolumn{2}{c|}{3$\times \ptop$}
            & \multicolumn{4}{c}{$\ptop$}
        \\
            & \begin{tikzpicture}
                \hspace{-2.5pt}
                \node (image) at (0, 0) {
                    \includegraphics[width=\imgwidth]{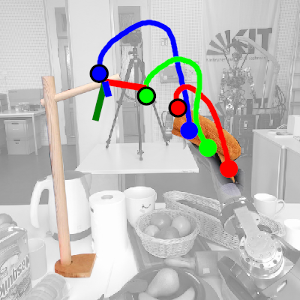} 
                };
                \node at (-0.35, 0.2) {$\lf$};
            \end{tikzpicture} 
            & \begin{tikzpicture}
                \hspace{-2.5pt}
                \node (image) at (0, 0) {
                    \includegraphics[width=\imgwidth]{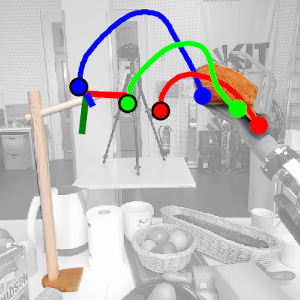}
                };
                \node at (-0.45, 0.15) {$\lf$};
            \end{tikzpicture} 
            & \begin{tikzpicture}
                \hspace{-2.5pt}
                \node (image) at (0, 0) {
                    \includegraphics[width=\imgwidth]{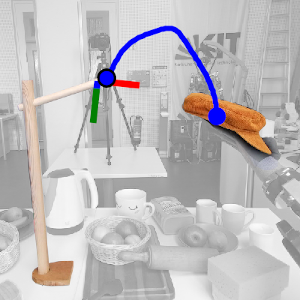}
                };
                \node at (-0.35, 0.25) {$\lf$};
            \end{tikzpicture}  
            & \begin{tikzpicture}
                \hspace{-2.5pt}
                \node (image) at (0, 0) {
                    \includegraphics[width=\imgwidth]{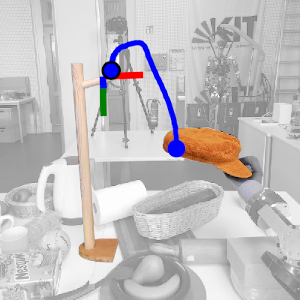}
                };
                \node at (-0.3, 0.25) {$\lf$};
            \end{tikzpicture}  
            & \begin{tikzpicture}
                \hspace{-2.5pt}
                \node (image) at (0, 0) {
                    \includegraphics[width=\imgwidth]{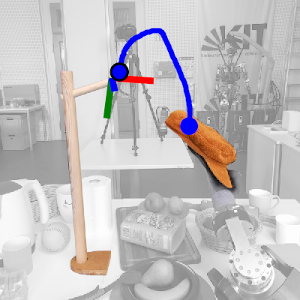}
                };
                \node at (-0.25, 0.25) {$\lf$};
            \end{tikzpicture} 
            & \begin{tikzpicture}
                \hspace{-2.5pt}
                \node (image) at (0, 0) {
                    \includegraphics[width=\imgwidth]{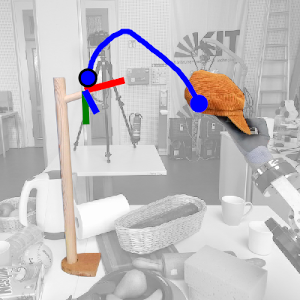}
                };
                \node at (-0.3, 0.25) {$\lf$};
            \end{tikzpicture} 
        \\
        \rotatebox[origin=c]{90}{Reproduction} 
            & \includegraphics[width=\imgwidth]{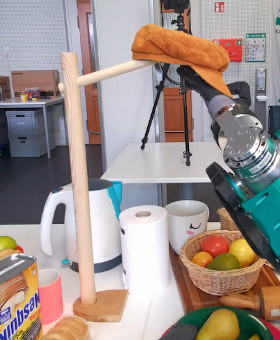} 
            & \includegraphics[width=\imgwidth]{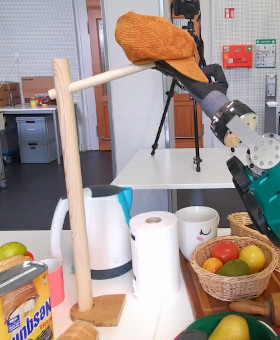} 
            & \includegraphics[width=\imgwidth]{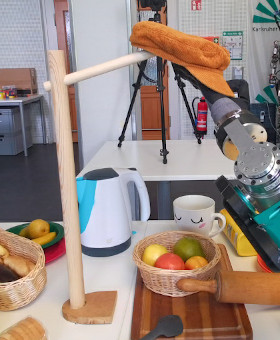} 
            & \includegraphics[width=\imgwidth]{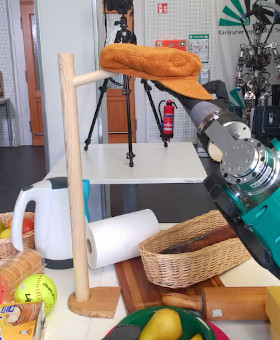}
            & \includegraphics[width=\imgwidth]{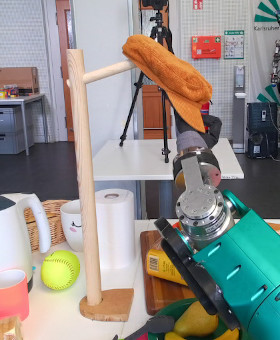} 
            & \includegraphics[width=\imgwidth]{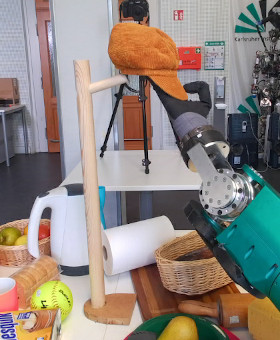}
        \\
        \bottomrule
    \end{tabular}
    \caption{
    Reproductions of tasks \taskabbr{HH}.  Legend as in \cref{table:insert_priority}.
    }
    \label{table:hang_exe}
    \vspace{-1.5ex}
\end{table*}

\subsubsection{Updating Constraints Incrementally}\label{sec:incremental}
Here, we apply \mbox{K-VIL} to few-shot imitation learning scenarios where additional demonstrations are incrementally provided for each task. In this case, K-VIL is trained based on the variance criteria to learn more generalizable task representations based on various linear and non-linear constraints (see Sections~\ref{sec:stage_1}-~\ref{sec:stage_2}).
\cref{fig:press_button,fig:fetch_tissue,fig:pouring,fig:hanging,fig:insertion} show the task representations of each task in $\mathcal{T}_E$ learned by K-VIL from several demonstrations.

\textbf{\mbox{Task extraction and reproduction of insertion tasks \taskabbr{IS}.}}
Providing several demonstrations allows us to consider variations in the demonstrated task, and thus to extract prioritized geometric constraints.
For example, the keypoint $\bk_1$ in \cref{subfig:insert_3} is the most invariant point on the stick across all demonstrations, while $\bk_2$ is subject to a $\ptol$ constraint. Note that this contrasts with \cref{subfig:insert_1}, where all keypoints were subject to $\ptop$ constraints.
With such task representation, KAC successfully handles all stick lengths by fulfilling the $\ptop$ and $\ptol$ constraints, as shown in \cref{table:insert_priority} for \task{IS}{3} (P). Despite a drop in accuracy and precision (see \cref{sec:eval_kac}, \cref{table:kac_eval}), the KAC without priority still leads to successful task completion in this case (see \task{IS_{np}}{3} (Q) in \cref{table:insert_priority}).
Overall, the extrapolation abilities of K-VIL are significantly increased by providing $3$ demonstrations instead of $1$.
Note that, due to the nature of the $\ptol$ constraint and the priority mechanism in KAC, sticks of arbitrary length can be handled. As the line manifold goes through both $\bk_1$ and $\bk_2$, K-VIL implicitly learns a colinear constraint for the two keypoints.

\textbf{Task extraction and reproduction of \taskabbr{PB} and \taskabbr{FT} tasks.}
For these two tasks, $3$ demonstrations are not sufficient to completely represent the task, and K-VIL may coincidentally extract a superfluous $\ptol$ constraint (see $\mathcal{M}_{\text{line}}$ in \cref{subfig:press_3} and \cref{subfig:tissue_3}). 
This forces the robot to place its hand similarly as demonstrated by the human. However, due to the third-person view adopted for the demonstration, this cannot be achieved by the robot, therefore resulting in failed executions of the tasks \task{PB}{3} and \task{FT}{3} as for \task{PB}{1} and \task{FT}{1} (see \cref{table:failure_case_fetch_press}).
The unnecessary $\mathsf{p2l}$ constraints are removed by providing \mbox{K-VIL} with an additional demonstration (see \cref{subfig:tissue_4,subfig:press_4}), allowing the tasks to be successfully reproduced (see \task{PB}{4} and \task{FT}{4} in \cref{table:failure_case_fetch_press}).

\textbf{Task extraction and reproduction of \taskabbr{PW} tasks.}
Similarly to \taskabbr{PB} and \taskabbr{FT} tasks, K-VIL's representation obtained from $3$ demonstrations for \taskabbr{PW} results in superfluous $\ptop$ and $\ptol$ constraints. 
Although the task may still be executed by the robot (see \task{PW}{3} in the second column of \cref{table:pouring_repro}), the superfluous constraints may lead to collisions between the kettle and the environment in other cases. 
For example, in the third column in \cref{table:pouring_repro}), with some specific initial poses of the kettle, the generated VMPs and the line constraints lead to rotation of the kettle in reversed direction during the execution, thus causing reproduction failures. 
These restrictive task representations are alleviated by providing K-VIL with an additional demonstration. By doing so, the problematic constraints are updated to different types, e.g., the $\ptol$ constraint in \cref{subfig:pour_3} becomes a $\ptoP$ constraint in \cref{subfig:pour_4}.
This significantly improves K-VIL's extrapolation abilities, as the $\ptoP$ constraint is necessary to constrain the kettle in a vertical plane while being less restrictive than the previous $\ptol$ constraint. 
Notice that the density force within the plane constraint ensures the tilting angle of the kettle is similar to the demonstrations.
A successful reproduction of such task representations is shown in \cref{table:pouring_repro} (\task{PW}{4}).
When enough demonstrations are provided ($N=11$ in \cref{subfig:pour_11}), K-VIL instead extracts a $\mathsf{p2c}$ constraint for the keypoint $\bk_2$ at the bottom of the kettle. 
The reproduction results with different types of kettles are shown in the last five columns of \cref{table:pouring_repro} (\scaletask{PW}{11}{0.7}). Intuitively speaking, the $\ptoc$ constraint aligns better with our understanding of a pouring task. Moreover, the last column in \cref{table:pouring_repro} shows that, given a tilted initial pose of the kettle, KAC is able to correct the pose of the kettle in the end.
The set of \taskabbr{PW} tasks in \cref{table:pouring_repro} also demonstrates that the proposed approach generalizes well to categorical objects with different colors, sizes and shapes, and is robust to change of background and viewpoints.

\textbf{Task extraction and reproduction of \taskabbr{HH} tasks.}
Unlike \taskabbr{PB}, \taskabbr{FT} and \taskabbr{PW} tasks, K-VIL's representation obtained from $3$ demonstrations for \taskabbr{HH} does not result in superfluous $\ptol$ constraints. 
Instead, due to the obvious pose variations in the hat, K-VIL consistently extracts a single keypoint $\bk_1$ on the backside of the hat with a $\ptop$ constraint, which encodes the demonstrated position invariances observed in the local frame near the end of the hanging stick.
As shown in \cref{table:hang_exe}, the target poses of the hat in \task{HH}{3} vary according to its different initial poses, while in \task{HH}{1} they are fully determined by the $3\ \ptop$ constraints. 
\cref{subfig:hat_4,subfig:hat_5,subfig:hat_shape} show the influence of the shape variations on the selection of local frames $\lf$.

\subsubsection{Evaluation summary}
\label{sec:eval_task_repr_summary}
As shown by our experiments, \mbox{K-VIL}'s task representations allow the successful learning of diverse tasks and their reproduction in new cluttered scenes with large shape and pose variations in categorical objects. 
As opposed to~\cite{Pari2022TheSE,yang2022viptl}, our approach is not constrained to conserving the same viewpoint between demonstrations and reproductions, and thus is more flexible.
Here, we further discuss the influence of the number and diversity of the demonstrations on such task representations.

\textbf{Limitations of scarce demonstrations.}
The task representations learned from scarce demonstrations hinder the performance and the extrapolation ability of K-VIL in three ways, namely,
\begin{enumerate*}[label=(\roman*)]
    \item they may be embodiment-dependent, and thus cannot be reproduced by the robot, e.g., in \cref{table:failure_case_fetch_press} for \task{PB}{1}, \task{PB}{3}, \task{FT}{1} and \task{FT}{3};
    \item they may lead to collisions during the execution, e.g., in \task{IS}{3}-\emph{ext. long} (see \cref{table:insert_priority}) and \task{PW}{3} for improper kettle start pose (see \cref{table:pouring_repro});
    \item when reproduced successfully, the control accuracy is reduced compared to the task representations learned from more demonstrations (see \cref{sec:eval_kac} for details).
\end{enumerate*}
These limitations motivate us to evaluate the number of demonstrations that are required to learn a generalizable task representation according to the criteria of \cref{sec:eval_protocal} for each considered task.

\begin{table}[t]
\centering
\setlength\tabcolsep{4pt}
\begin{tabular}{@{}cclllll@{}}
    \toprule
        & & \multicolumn{5}{c}{Number of demonstrations (N)} \\ \cmidrule{3-7}
        \multicolumn{2}{l}{$\mathcal{T}_{E}$}
        & 1 
        & 3 
        & 4 
        & 5 
        & 11
      \\ \midrule
        a
        & \taskabbr{PB} 
        & $3 \times \ptop$ 
        & $\ptop, \ptol$
        & \cellcolor{yellow!30}$\textcolor{blue}{\bm{\ptop}}$
        & \cellcolor{yellow!30}$\textcolor{blue}{\bm{\ptop}}$
        & \cellcolor{yellow!30}$\textcolor{blue}{\bm{\ptop}}$
      \\
        b 
        & \taskabbr{FT} 
        & $3 \times \ptop$ 
        & $\ptop, \ptol$
        & \cellcolor{yellow!30}$\textcolor{blue}{\bm{\ptop}}$
        & \cellcolor{yellow!30}$\textcolor{blue}{\bm{\ptop}}$
        & \cellcolor{yellow!30}$\textcolor{blue}{\bm{\ptop}}$
    \\
        c
        & \taskabbr{PW} 
        & $3 \times \ptop$ 
        & $\ptop, \ptol$
        & $\textcolor{blue}{\bm{\ptop}, \bm{\ptoP}}$
        & $\textcolor{blue}{\bm{\ptop}, \bm{\ptoP}}$
        & \cellcolor{yellow!30}$\textcolor{blue}{\bm{\ptop}, \bm{\ptoc}}$
    \\
        d
        & \taskabbr{HH} 
        & $3 \times \ptop$ 
        & \cellcolor{yellow!30}$\textcolor{blue}{\bm{\ptop}}$
        & \cellcolor{yellow!30}$\textcolor{blue}{\bm{\ptop}}$
        & \cellcolor{yellow!30}$\textcolor{blue}{\bm{\ptop}}$
        & \cellcolor{yellow!30}$\textcolor{blue}{\bm{\ptop}}$
    \\
        e
        & \taskabbr{IS} 
        & $3 \times \ptop$ 
        & \cellcolor{yellow!40}$\textcolor{blue}{\bm{\ptop}, \bm{\ptol}}$
        & \cellcolor{yellow!40}$\textcolor{blue}{\bm{\ptop}, \bm{\ptol}}$
        & \cellcolor{yellow!40}$\textcolor{blue}{\bm{\ptop}, \bm{\ptol}}$
        & \cellcolor{yellow!40}$\textcolor{blue}{\bm{\ptop}, \bm{\ptol}}$
    \\
        f
        & \taskabbr{CT}
        & $3 \times \ptop$ 
        & \cellcolor{yellow!40}$\textcolor{blue}{\bm{\ptol}, \bm{\ptol}}$
        & \cellcolor{yellow!40}$\textcolor{blue}{\bm{\ptol}, \bm{\ptol}}$
        & \cellcolor{yellow!40}$\textcolor{blue}{\bm{\ptol}, \bm{\ptol}}$
        & \cellcolor{yellow!40}$\textcolor{blue}{\bm{\ptol}, \bm{\ptol}}$
    \\ \bottomrule
\end{tabular}
\caption{Extraction tasks and the geometric constraints of each task learned from different number of demonstrations. We mark the cases (\raisebox{-1.3mm}{\sampleline{yellow!40,line width=3mm}}) where the learned task representations converge, and highlight the cases (in \textcolor{blue}{\textbf{blue}}) where the learned task representations are generalizable.}
\label{table:ext_tasks}
\vspace{-1.5ex}
\end{table}

\textbf{Adequate number of demonstrations.}
\cref{table:ext_tasks} summarizes the extracted geometric constraints of the five extraction tasks for different number $N$ of demonstrations. 
It is interesting to notice that the learned task representations do not change anymore, i.e., converge, after a certain number of demonstrations, e.g., $N=4$ for \taskabbr{PB} and \taskabbr{FT}, $N=11$ for \taskabbr{PW}, and $N=3$ for \taskabbr{HH} and \taskabbr{IS}.
Moreover, the task representations become generalizable almost at the same time as they converge. As an exception, the representations of \taskabbr{PW} already generalize with a $\ptop$ and a $\ptoP$ constraint learned from $4$ demonstrations.
Importantly, the learned geometric constraints do not have to include a $\ptop$ constraint, see e.g., the \taskabbr{CT} task that is represented by two $\ptol$ constraints.
Overall, this demonstrates that our approach efficiently extracts generalizable task representations from considerably less demonstrations than state-of-the-art approaches.

\textbf{Object pose and shape variations.}
Importantly, K-VIL's task representations rely on the variations observed in the demonstrations to extract appropriate constraints.
For example, in \taskabbr{PB}, the pose variations of the demonstrator's hand (\cref{subfig:press_4}) allow K-VIL to distinguish the tip of the middle finger (used to press the button) from the other candidate points on the hand. In other words, these variations enable the efficient extraction of the keypoint $\bk_1$ subject to a $\ptop$ constraint. 
The spatial distribution of keypoints across demonstrations is also obtained via pose variations.
For instance, this allows K-VIL to associate the keypoint $\bk_2$ with geometric constraints such as $\ptol$ (\cref{subfig:pour_3,subfig:ct_constraints}), $\ptoP$ (\cref{subfig:pour_4}) and $\ptoc$ (\cref{subfig:pour_11}). 

\begin{table}[t]
\centering
\begin{tabu}{@{}ccllccl@{}}
    \toprule
        index
        & $\mathcal{T}_{E}$  
        & role 
        & object
        & PV
        & SV
        & \multicolumn{1}{c}{TR}
    \\ \midrule
        \multirow{2}{*}{\hypertarget{tab:vari_1}{1}}
        & \multirow{2}{*}{\taskabbr{PB}}
        & \master{}
        & kettle
        & - 
        & \xmark
        & \multirow{2}{*}{\cref{subfig:press_3,subfig:press_4}}
    \\
        &
        & \slave{}
        & hand
        & \cmark
        & \xmark
        &
    \\ \tabucline[0.2pt gray!20 off 0pt]{-} 
        \multirow{2}{*}{2}
        & \multirow{2}{*}{\taskabbr{FT}}
        & \master{}
        & tissue box
        & - 
        & \cmark
        & \multirow{2}{*}{\cref{subfig:tissue_3,subfig:tissue_4}}
    \\
        &
        & \slave{}
        & hand
        & \cmark
        & \xmark
        &
    \\ \tabucline[0.5pt gray!20 off 0pt]{-} 
        \multirow{2}{*}{3}
        & \multirow{2}{*}{\taskabbr{PW}}
        & \master{}
        & teacups
        & - 
        & \cmark
        & \multirow{2}{*}{\cref{subfig:pour_3,subfig:pour_4,subfig:pour_11}}
    \\
        &
        & \slave{}
        & kettle
        & \cmark
        & \cmark
        &
    \\ \tabucline[0.5pt gray!20 off 0pt]{-} 
        \multirow{2}{*}{4}
        & \multirow{2}{*}{\taskabbr{HH}}
        & \master{}
        & rack
        & - 
        & \xmark
        & \multirow{2}{*}{\cref{subfig:hat_3}}
    \\
        &
        & \slave{}
        & hat
        & \cmark
        & \xmark
        &
    \\ \tabucline[0.5pt gray!20 off 0pt]{-} 
        \multirow{2}{*}{5}
        & \multirow{2}{*}{\taskabbr{HH}}
        & \master{}
        & rack
        & - 
        & \cmark
        & \multirow{2}{*}{\cref{subfig:hat_4}}
    \\
        &
        & \slave{}
        & hat
        & \cmark
        & \xmark
        &
    \\ \tabucline[0.5pt gray!20 off 0pt]{-} 
        \multirow{2}{*}{6}
        & \multirow{2}{*}{\taskabbr{HH}}
        & \master{}
        & rack
        & - 
        & \xmark
        & \multirow{2}{*}{\cref{subfig:hat_5,subfig:hat_shape}}
    \\
        &
        & \slave{}
        & hat
        & \cmark
        & \cmark
        &
    \\ \tabucline[0.5pt gray!20 off 0pt]{-} 
        \multirow{2}{*}{7}
        & \multirow{2}{*}{\taskabbr{IS}}
        & \master{}
        & paper roll
        & - 
        & \xmark
        & \multirow{2}{*}{\cref{subfig:insert_3,subfig:insert_5}}
    \\
        &
        & \slave{}
        & stick
        & \xmark
        & \cmark
        &
    \\ \tabucline[0.5pt gray!20 off 0pt]{-} 
        \multirow{2}{*}{8}
        & \multirow{2}{*}{\taskabbr{CT}}
        & \master{}
        & dustpan
        & - 
        & \cmark
        & \multirow{2}{*}{\cref{fig:clean_table_task}}
    \\
        &
        & \slave{}
        & brush
        & \cmark
        & \cmark
        &
    \\ \bottomrule
\end{tabu}
\caption{Pose variations (PV) and shape variations (SV) in the demonstrations along with the detected \master{} and \slave{} objects. The corresponding task representations (TR) are linked in the last column. Pose variations of the \master{} objects are not relevant (-) as the local frames representing the object pose are constructed on the \master{}s.}
\label{table:shape_pose_variation}
\vspace{-1.5ex}
\end{table}

Shape variations also facilitate the efficient extraction of keypoints and geometric constraints. 
For example, the stick length variations in the insertion task \taskabbr{IS} allow K-VIL to extract the keypoints $\bk_1$ and $\bk_2$ subject to a $\ptop$ and $\ptol$ constraint, respectively, 
as intuitively shown in \cref{subfig:insert_3}.
In other cases, shape variations in the \master{} objects help to remove redundancy in canonical local frames. 
For example, all the $J=300$ canonical local frames on the rack in task \taskabbr{HH} of \cref{subfig:hat_3} are equivalent. This is due to the absence of variations in the rack across the provided demonstrations.
In this case, K-VIL selects the local frame $\lf$ as the closest on average to the keypoint $\bk_1$ on the hat. 
This redundancy is removed by introducing shape variations in the \master{} objects. 
For example, by considering two variations of the rack in the task \taskabbr{HH} as in \cref{subfig:hat_4}, $\bk_1$ is position-invariant only if it is represented in the local frames near the contact point between the rack and the hat.
This allows K-VIL to focus on these local frames and to filter out the others.
We showed in \cref{subfig:hat_5} that K-VIL's representation converges and remains the same as \cref{subfig:hat_3,subfig:hat_4} even if more demonstrations with large shape and pose variations in the hats are available. 
Similarly, shape variations in the tissue boxes in \taskabbr{FT} allow the selection of local frames around the grasping point, while shape variations in the teacups (\taskabbr{PW}) ensure that local frames for the pouring task are around their rim.
A summary of the effect of pose and shape variations in the considered tasks is given by \cref{table:shape_pose_variation}.
Overall, pose or/and shape variations of \master{} and \slave{} objects in the demonstrations are not only handled by K-VIL, but also facilitate the efficient, joint extraction of local frames, keypoints, and geometric constraints.

\subsection{Evaluation of KAC}
\label{sec:eval_kac}
In \cref{sec:eval_task_repr}, we demonstrated K-VIL's ability to extract generalizable task representations from a small number of demonstrations. In particular, we showed that these task representations successfully adapt to new cluttered scenes with categorical objects, regardless of whether the task can be reproduced by the robot.
Therefore, in this section, we evaluate the proposed KAC in terms of control accuracy, control precision (i.e., repeatability) and success rate on the $18$ tasks described in \cref{sec:eval_protocal} (see also \cref{table:kac_eval}). 
Each task is reproduced $N_r = 20$ times according to the evaluation protocol in \cref{sec:eval_protocal}. 
For each task, we record the trajectories of all relevant keypoints during the execution period. 
Since we are particularly interested in the regulation behavior of KAC when the keypoints satisfy their corresponding geometric constraints, i.e., when the VMPs finish, we record the keypoint trajectories for an additional \SI{2}{\second} time window, denoted by $\mathscr{T}_\text{end}$.
These trajectories are then compared to the corresponding keypoints' target trajectories, which correspond to the VMP trajectories for keypoints subject to $\ptop$ constraint, and to the attractor trajectories for keypoints subject to other types of constraints.
Notice that, for the latter, the 3D position of the attractor is recovered from the corresponding 1-dimensional VMP in the orthogonal direction of the corresponding principal manifold (see also \cref{kac:attract}).

\begin{figure}[t]
    \centering
    \vspace{-0.5em}
    \begin{tikzpicture}
        \node (image) at (0.1, 0) {\includegraphics[width=0.95\columnwidth]{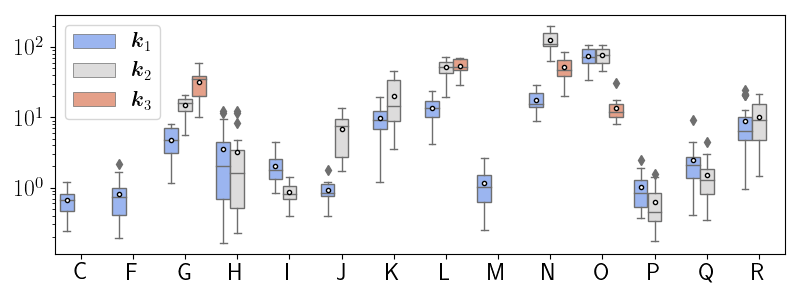}
        };
        \node at (-4.2, 0.1) {
            \scalebox{0.8}{
                \rotatebox{90}{Regulation Error (\unit{\milli\meter})}
            }
        };
    \end{tikzpicture}
    \vspace{-2em}
    \caption{
	Regulation errors between each keypoint and its target estimated in $\mathscr{T}_\text{end}$ over $N_r=20$ trials for tasks C to R in \cref{table:kac_eval}. The mean regulation error, i.e., control accuracy Acc., is depicted as \inlinecirc{}.
	The box shows the first and third quartiles of the regulation error of each keypoint, with the bar inside it indicating the median.
	}
    \label{fig:accuracy}
    \vspace{-1.5ex}
\end{figure}

First, we evaluate the ability of KAC to satisfy the learned keypoints' geometric constraints.
To do so, we compute the regulation error of each keypoint during $\mathscr{T}_\text{end}$ of each trial as 
\begin{equation*}
    \textstyle e_v = \frac{1}{T_r}\sum_{t=0}^{T_r}\Vert \bk_{l,v}^g(t) - \bk_{l,v}(t)\Vert_2, \quad v\in [1, N_r],
\end{equation*}
where $T_r$ is the total timesteps recorded in $\mathscr{T}_\text{end}$, and \smalleq{$\bk_{l,v}, \bk_{l,v}^g$} are the recorded and target positions of the considered keypoint in the $v\nth$ trial, respectively.
\cref{fig:accuracy} displays the distribution of the keypoint's regulation errors for $20$ trials of each task, where the mean values (\inlinecirc{}) correspond to the control accuracy 
\begin{equation*}
    \textstyle \text{Acc.} = \frac{1}{N_r}\sum_{v=0}^{N_r} e_v.
\end{equation*}
Second, we evaluate the control precision (i.e., repeatability) of KAC for all keypoints and all tasks. It is computed as
\begin{equation}
    \text{Prec.} = \sqrt{\frac{1}{N_r}\frac{1}{T_r} \sum_{v=1}^{N_r} \sum_{t=1}^{T_r} \Vert \bk_{l,v}(t) - \bm{\mu}_{\bk_{l, v}} \Vert_2^2},
\end{equation}
where we defined $\mathbf{\bm{\mu}}_{\bk_{l,v}} = \frac{1}{T_r} \sum_{t=1}^{T_r} \bk_{l,v}(t)$.
Finally, we also report the success rate obtained for each task according to the evaluation protocols described in \cref{sec:eval_protocal}.

\cref{table:kac_eval} presents the evaluation results of KAC in terms of the three aforementioned metrics.
As suggested by the qualitative evaluations in \cref{table:failure_case_fetch_press}, when learned from fewer than $4$ demonstrations, the tasks \taskabbr{PB} and \taskabbr{FT} result in a $0\%$ success rate (see A-B, D-E in \cref{table:kac_eval}). As discussed in \cref{sec:incremental}, this is due to geometric constraints that are unreachable for the robot.
In contrast, the task representations learned from $4$ demonstrations are generalizable (see \cref{table:ext_tasks} in \cref{sec:eval_task_repr_summary}), and thus KAC reaches sub-millimeter control accuracy and precision, as well as $\geq 90\%$ success rates (see C and F in \cref{table:kac_eval}).

As shown in \cref{fig:accuracy} and \cref{table:kac_eval} (G-J), \task{PW}{4} and \scalebox{1.0}{\scaletask{PW}{11}{0.8}} outperform \task{PW}{3} in terms of control accuracy, precision, and success rate. This is due to the fact that, in contrast to \task{PW}{3}, \task{PW}{4} and \scaletask{PW}{11}{0.7} are generalizable (see \cref{table:ext_tasks} and \cref{table:pouring_repro}). Although \task{PW}{1} displays a relatively high control precision and success rate, its control accuracy remains low and the pose of the kettle is fully constrained by $3$ $\ptop$ constraints, which hinders K-VIL's extrapolation abilities.
Interestingly, while $\bk_1$'s control accuracy in the task \taskabbr{PW} increases with the number of demonstrations, $\bk_2$'s highest control accuracy is obtained in \task{PW}{4} (I). This is due to $\bk_2$'s $\ptoP$ constraint in \task{PW}{4}, which is easier to fulfill than the $\ptoc$ constraint in \scaletask{PW}{11}{0.7}.
We observe a lower precision in \task{PW}{4} than in \scaletask{PW}{11}{0.7} for the same reason.

Similarly as in \taskabbr{PW}, \task{HH}{3} (M) and \task{IS}{3} (P) are reproduced with higher control accuracy and precision than \task{HH}{1} (L) and  \task{IS}{1} (N), respectively.
Moreover, thanks to the priority introduced in KAC, \task{HH}{1} (L) and \task{IS}{1} (N) achieve $95\%$ and $62\%$ success rate, respectively, despite the single available demonstration. The lower performance of \task{IS}{1} is explained by the fact that we consider an extremely long stick, which requires high generalization capabilities, in the reproduction (see \cref{table:insert_priority}). In contrast, the hat in \taskabbr{HH} is only slightly deformable and thus results in low shape variations.
Notice that, for \taskabbr{HH} and \taskabbr{IS}, we do not compare the control accuracy between keypoints subject to different geometric constraints (in gray in \cref{table:kac_eval}). As these keypoints are controlled according to different priorities within KAC, their reported accuracy highly depends on the shape variations occurring in these two tasks.
For example, when using sticks of various lengths in the insertion task \task{IS}{1} (N), KAC assigns the highest priority to the keypoint $\bk_1$, so that $\bk_1$ is obviously controlled with higher accuracy than $\bk_2$ and $\bk_3$. Interestingly, $\bk_2$'s highest regulation error in this task is $\sim$\SI{150}{\milli\meter}, which corresponds to the maximum length difference between the sticks used in the demonstration and in the reproduction.
Therefore, in \taskabbr{HH} and \taskabbr{IS}, the control accuracy of $\bk_2$ and $\bk_3$ is rather affected by the experiment setups (e.g., stick lengths) than by KAC.
Since the two $\ptol$ constraints in \task{CT}{3} share the same priority, the two keypoints are equally controlled towards the region with high likelihood on the corresponding principal lines. The two spring-damper systems for the two keypoints result in equilibrium. Therefore, the control accuracy and the precision of the two keypoints are comparable.

\begin{table}[t]
    \centering
    \setlength\tabcolsep{3pt}
    \tabulinesep=1.5pt
    \begin{tabu}{@{}crcccccccc@{}}
        \toprule
          &  & & \multicolumn{3}{c}{Acc. (\unit{\milli\meter})} & \multicolumn{3}{c}{Prec. (\unit{\milli\meter})} & 
        \\ \cmidrule{4-6} \cmidrule{7-9}
          \multicolumn{2}{c}{$\mathcal{T}_R$} & TR & $\bk_1$ & $\bk_2$ & $\bk_3$  & $\bk_1$ & $\bk_2$ & $\bk_3$ & R (\%) 
        \\ \midrule
        A & \task{PB}{1} & \cref{subfig:press_1}                & \nokpt & \nokpt & \nokpt      & \nokpt & \nokpt & \nokpt & 0 \\
        B & \task{PB}{3} & \cref{subfig:press_3}                & \nokpt & \nokpt & \nokpt      & \nokpt & \nokpt & \nokpt & 0 \\
        C & \task{PB}{4} & \cref{subfig:press_4}                & $0.67$ & - & -                  & $0.34$ & - & - & $90$ \\ 
        \tabucline[0.5pt gray!20 off 0pt]{-}
        D & \task{FT}{1} & \cref{subfig:tissue_1}               & \nokpt & \nokpt & \nokpt      & \nokpt & \nokpt  & \nokpt & $0$ \\
        E & \task{FT}{3} & \cref{subfig:tissue_3}               & \nokpt & \nokpt & \nokpt      & \nokpt & \nokpt  & \nokpt & $0$ \\
        F & \task{FT}{4} & \cref{subfig:tissue_4}               & $0.82$ & - & -                  & $0.65$ & -  & - & 100 \\
        \tabucline[0.5pt gray!20 off 0pt]{-}
        G & \task{PW}{1} & \cref{subfig:pour_1}                 & $4.81$ & $15.01$ & $32.09$          & $0.75$ & $0.78$ & $0.72$ &  $95$ \\
        H & \task{PW}{3} & \cref{subfig:pour_3}                 & $3.56$ & $3.21$ & -               & $1.20$ & $2.12$ & - & $87$ \\
        I & \task{PW}{4} & \cref{subfig:pour_4}                 & $2.01$ & $0.88$ & -               & $1.04$ & $5.67$ & - & $94$ \\
        J & \scaletask{PW}{11}{0.7} & \cref{subfig:pour_11}     & $0.93$ & $6.80$ & -               & $0.42$ & $0.56$ & - & $100$ \\
        K & \scaletask{PW_{np}}{11}{0.7} & \cref{subfig:pour_11} & $9.71$ & $20.08$ & -             & $0.90$ & $1.16$ & - & $100$ \\
        \tabucline[0.5pt gray!20 off 0pt]{-}
        L & \task{HH}{1} & \cref{subfig:hat_1}                  & $13.53$ & \textcolor{gray!80}{$51.33$} & \textcolor{gray!80}{$54.28$}                 & $0.67$ & $0.67$ & $0.67$ & $95$ \\
        M & \task{HH}{3} & \cref{subfig:hat_3}                  & $1.48$ & - & -                 & $0.47$ & - & - & $95$ \\
        \tabucline[0.5pt gray!20 off 0pt]{-}
        N & \task{IS}{1}  & \cref{subfig:insert_1}              & $17.77$ & \textcolor{gray!80}{$125.40$} & \textcolor{gray!80}{$51.98$}              & $1.55$ & $1.30$ & $1.34$ & $62$ \\
        O & \task{IS_{np}}{1}  & \cref{subfig:insert_1}         & $74.55$ & \textcolor{gray!80}{$77.05$} & \textcolor{gray!80}{$13.40$}         & $1.03$ & $0.97$ & $1.02$ & $25$ \\
        P & \task{IS}{3} & \cref{subfig:insert_3}               & $1.01$ & $0.63$ & -               & $0.36$ & $0.36$ & - & $95$ \\
        Q & \task{IS_{np}}{3}  & \cref{subfig:insert_3}         & $2.46$ & $1.50$ & -               & $2.32$ & $1.39$ & - & $90$ \\ 
        \tabucline[0.5pt gray!20 off 0pt]{-}
        R & \task{CT}{3} & \cref{subfig:ct_constraints} & $8.89$ & $10.22$ & -               & $2.04$ & $2.03$ & - & $95$ \\
        \bottomrule
    \end{tabu}
    \caption{Evaluation of KAC in terms of control accuracy (Acc.), precision (Prec.), and success rate (R) for each category of tasks with task representations (TR) learned from different numbers \circled{N} of demonstrations. 
    Ablation studies (denoted by $\mathsf{{\cdot}_{np}}$) are also conducted in tasks K, O, and Q by removing the priority (see \cref{kac:priority}) from KAC. 
    The cases where data is not available due to failure execution and where a specific keypoint is not required are denoted as \nokpt{} and -, respectively. Gray numbers in N and O indicate inconsequential values.
    }
    \label{table:kac_eval}
    \vspace{-1.5ex}
\end{table}

Compared to \task{IS}{1}, the ablation study \task{IS_{np}}{1} (O), conducted without KAC's priorities, shows a drop in success rate ($25\%$), as $\bk_3$ achieves the highest accuracy at the expense of $\bk_1$ and $\bk_2$. 
This is expected as the three identical virtual spring-damper systems for $\bk_1,\bk_2$ and $\bk_3$ are in equilibrium, and $\bk_3$ usually locates in the middle of $\bk_1$ and $\bk_2$.
As the ablation tasks \scaletask{PW_{np}}{11}{0.8} (K) and \task{IS_{np}}{3} (Q) are reproduced from generalizable task representations, removing KAC's priorities does not affect their success rate. However, the control accuracy and precision drop compared to \scaletask{PW}{11}{0.8} (J) and \task{IS}{3} (P).

\section{Discussion}
\label{sec:conclusion}

In this paper, we proposed the novel keypoints-based visual imitation learning (K-VIL) approach that learns sparse, object-centric, and embodiment-independent task representations from a small set of demonstration videos.
K-VIL's task representations are based on the extraction of geometric constraints by a PCE, which covers a wide range of constraints.
The proposed PCE enables one-shot and few-shot VIL and updates the learned task representations when additional demonstrations are incrementally provided, thus endowing them with enhanced extrapolation capabilities.
K-VIL's task representations also include task-specific keypoint control policies encoded as VMPs, which are leveraged for task execution by a prioritized keypoint-based admittance controller (KAC). Compared to control policies based on RL or on visual servoing, VMPs allow a flexible temporal scaling and support via-points (including start and target position) adaptation. Therefore, they crucially contribute to K-VIL's generalization capabilities by extrapolating the keypoint target positions on the learned principal manifold.

As highlighted in our evaluation, K-VIL consistently learned generalizable task representations for six daily manipulation tasks, which involved highly cluttered scenes, new instances of categorical objects, and large variations in object poses and shapes. Importantly, we showed that the learned task representations converges and becomes generalizable with significantly fewer demonstrations than state-of-the-art approaches such as \cite{sermanet_timecontrastive_2018,Sharma2019ThirdPersonVI,pathakICLR18zeroshot,jin_generalizable_2022}.
Interestingly, the sparse keypoint-based geometric constraints extracted by \mbox{K-VIL} mostly aligned with human intuition. 
This includes the extraction of a single $\ptop$ constraint for pressing a button, of a pair of $\ptop$ and $\ptol$ constraints for the insertion task, 
and of a $\ptop$ coupled with a $\ptoc$ constraint for the pouring task, among others.

It is important to emphasize that the decomposed control and priority mechanism of the KAC allowed us to endow K-VIL with reliable extrapolation capabilities. Indeed, our quantitative evaluations demonstrated K-VIL's ability to reproduce the learned task representations with high control accuracy, control precision, and success rate. 
Particularly, \mbox{K-VIL} accurately handled very large shape variations in the considered insertion task.
In contrast, previous works did not or only briefly discuss the extrapolation capabilities of their approaches~\cite{sieb2020graph,yang2022viptl,jin_generalizable_2022,Adversarial2022}. For instance, Jin \& Jagersand~\cite{jin_generalizable_2022} only showed extrapolation to another instance of the hammer category with very small shape variation without providing any quantitative evaluations. 

It is important to note that the variations in object poses and shapes play an essential role in learning generalizable task representations. This is even more relevant when only a small number of demonstrations are provided.
Without such variations, K-VIL can still generalize to categorical objects thanks to the dense visual descriptors, but achieve lower control accuracy, precision, and success rate, and may fail in some extreme cases, e.g., in the one-shot VIL setup.

\subsection{Limitations and Future Work}

K-VIL imposes limitations in terms of visual perception models and task representations. On the one hand, we assume that the keypoints are on the surface of objects and omit transparent, reflective, and thin objects (note that this is also discussed in \cite{simeonov_neural_2021,yen2022nerfsupervision}). This hinders K-VIL from being used in many real-world tasks. 
Furthermore, all keypoints must be visible in the demonstrations, which may not always be enforced in reality. 
In other words, K-VIL learns from demonstrations with and without viewpoint mismatch, as long as the keypoints of interest are not occluded.
In the long run, we believe that the dense correspondence models should be combined with state-of-the-art scene representation models (e.g., \cite{yen2022nerfsupervision}) or with point generative models (e.g., \cite{Lei2022CaDeXLC}) for better correspondence detection and for tackling the occlusion problems.
This would allow the imitator to observe objects that are visually more challenging and to learn the task from demonstrations with (self-)occlusions.

It is worth noticing that K-VIL's keypoints correspond to the sub-symbolic parameters of a motion. Therefore, they do not necessarily have a clear semantic interpretation, which is also important for learning comprehensive task models.
Bridging the gap between the symbolic and sub-symbolic levels remains an important challenge in (visual) IL. 
Importantly, the symbolic representation of a task~\cite{Dreher2022,Hassanin2018VisualAA} also has limitations, which can be alleviated by integrating sub-symbolic information. 
For example, a $\mathsf{contain}$ affordance in a pouring task implies that the opening of the spout of the kettle should be placed above the $\mathsf{contain}$ affordance region~\cite{hadjivelichkov_oneshot_2022}. 
However, this semantic representation alone cannot describe different types of pouring: For example, pouring beer requires tilting the glass and aligning the beer with the side of a glass.
Instead, additional sub-symbolic parameters would allow realizing specific styles of task execution.

In this sense, K-VIL deals with the sub-symbolic part of the task. Namely, its ability to update the geometric constraints allows us
\begin{enumerate*}[label=(\roman*)]
    \item to reproduce a task with a specific style, and
    \item to eliminate unnecessary keypoints and geometric constraints and to update the distribution of the keypoints on the extracted constraints when more demonstration styles are available.
\end{enumerate*}
\mbox{K-VIL} may then be augmented with an extraction method~\cite{jiang_synergies_2021} to estimate the links between the extracted keypoints and the symbolic task representation. For instance, the probability distribution of the keypoints on their principal manifolds may be used to determine the affordance regions~\cite{hadjivelichkov_oneshot_2022,do_affordancenet_2018}, the spatial relations~\cite{Kartmann2021}, and the grasping or effect points~\cite{qin_keto_2020,jiang_synergies_2021}.
We will investigate these aspects in our future work.

In this paper, we only considered uni-manual manipulation tasks that can be modeled as the combination of five basic geometric constraints in \cref{fig:pca_pme} in a single layer of \master{}-\slave{} relationship.
As future work, we plan to extend \mbox{K-VIL} to include other types geometric constraint and to bimanual manipulation tasks by considering bimanual coordination strategies~\cite{Krebs2022} and a hierarchy of \master{}-\slave{} relationships. 
Moreover, we will extend K-VIL for periodic motions such as stirring or wiping motions~\cite{yang2022viptl}, as well as for handling articulated objects~\cite{xu_universal_2022,wu_vatmart_2022}.

\def\url#1{}
\bibliographystyle{IEEEtran} 
\bibliography{jianfeng_library_simplified,HumanoidsGroup}

\vfill

\end{document}